\documentclass[11pt,letterpaper]{article}

\usepackage[title]{appendix}

\usepackage{caption}
\usepackage{subcaption}
\usepackage{fullpage}
\usepackage{times}
\usepackage{array}
\usepackage{ragged2e}
\usepackage{multirow}
\usepackage{fancyhdr, amssymb, mathtools}
\usepackage[ruled,vlined]{algorithm2e}
\usepackage[dvipsnames,table]{xcolor}
\usepackage[margin = 1 in]{geometry}
\usepackage{graphicx}
\usepackage{outlines}
\usepackage{amsmath}

\usepackage{graphicx}
\usepackage{dirtytalk}

\usepackage{multirow}
\usepackage{arydshln}

\usepackage[english]{babel} 
\usepackage{blindtext}
\usepackage{amsfonts}
\usepackage{amssymb}
\usepackage{bbm}
\usepackage{mathtools}

\usepackage{hhline}
\usepackage{makecell}

\usepackage{xltabular}
\usepackage{arydshln} 
\usepackage{wrapfig}
\usepackage{enumitem}
\usepackage{floatrow}
\usepackage{xfrac}
\usepackage[font={footnotesize}, labelfont={bf}, labelsep=space, justification=justified, skip=0pt]{caption}
\usepackage[hang,flushmargin]{footmisc} 
\usepackage[colorlinks=true, allcolors=blue]{hyperref}

\definecolor{headcolor}{RGB}{255, 255, 255}
\definecolor{columncolor}{RGB}{255, 255, 255}
\definecolor{modelcolor}{RGB}{255, 255, 255}
\definecolor{optioncolor}{RGB}{255, 255, 255}

\usepackage[
backend=biber,
style=numeric-comp,
sorting=none,
maxnames = 10
]{biblatex}
\usepackage{cleveref} 
\AtBeginEnvironment{appendices}{\crefalias{section}{appendix}}

\newcommand\brackets[1]
{\mathopen{}\left[#1\right]\mathclose{}}
\newcommand\parens[1]{\mathopen{}\left(#1\right)\mathclose{}}
\newcommand\braces[1]{\mathopen{}\left\{#1\right\}\mathclose{}}

\newcommand{\cmt}[1]{} 
\newcommand{\fbest}{$y^*$}

\DeclareMathOperator*{\E}{\mathbb{E}}
\newcommand\Yb{\mathord{\mathrm{\textbf{Y}}}}

\newcommand{\xsspace}{\mathbb{X}}
\newcommand{\rsspace}{\mathbb{R}}

\newcommand{\mub}{\boldsymbol{\mu}}
\newcommand{\epsb}{\boldsymbol{\epsilon}}
\newcommand{\Sigmab}{\boldsymbol{\Sigma}}
\newcommand{\betab}{\boldsymbol{\beta}}

\newcommand{\thetab}{\boldsymbol{\theta}}

\newcommand{\omegab}{\boldsymbol{\omega}}

\newcommand{\yb}{\boldsymbol{y}}

\newcommand{\ub}{\boldsymbol{u}}
\newcommand{\tb}{\boldsymbol{t}}
\newcommand{\zetab}{\boldsymbol{\zeta}}
\newcommand{\xb}{\boldsymbol{x}}
\newcommand{\Xb}{\boldsymbol{X}}
\newcommand{\Ib}{\boldsymbol{I}}
\newcommand{\zb}{\boldsymbol{z}}
\newcommand{\hb}{\boldsymbol{h}}
\newcommand{\mb}{\boldsymbol{m}}
\newcommand{\Rb}{\boldsymbol{R}}
\newcommand{\Cb}{\boldsymbol{C}}

\newcommand{\Lb}{\boldsymbol{L}}

\newcommand{\pib}{\boldsymbol{\pi}}
\newcommand{\Ab}{\boldsymbol{A}}
\newcommand{\zetai}{\zeta i}

\newcommand{\gtorch}{\texttt{GPyTorch}}
\newcommand{\gp}{\texttt{GP+}}
\newcommand{\DNS}{\texttt{DNS-ROM}}
\newcommand{\sine}{\texttt{Sinusoidal}}
\newcommand{\borehole}{\texttt{Borehole}}
\newcommand{\boreholeM}{\texttt{\borehole-Mixed}}
\newcommand{\HEA}{\texttt{HEA}}
\newcommand{\wing}{\texttt{Wing}}
\newcommand{\hoip}{\texttt{HOIP}}
\newcommand{\beam}{\texttt{beam deflection}}
\newcommand{\MEI}{\texttt{SFBO\textsubscript{EI}}}
\newcommand{\LMGP}{\texttt{MFBO}}
\newcommand{\BOT}{\texttt{BoTorch}}
\newcommand{\nta}{\texttt{NTA}}
\newcommand{\uql}{\texttt{UQLab}}

\usepackage{titling}
\thanksheadextra{}{}
\thanksheadextra{}{} 
\usepackage{lipsum} 

\title{{GP+}: A Python Library for Kernel-based learning via Gaussian Processes}
\date{\vspace{-5ex}}
\usepackage{authblk}
\author[1]{Amin Yousefpour}
\author[1]{Zahra Zanjani Foumani}
\author[1]{Mehdi Shishehbor}
\author[1]{Carlos Mora}
\author[1]{Ramin Bostanabad \thanks{Corresponding Author: Raminb@uci.edu \\\href{https://github.com/Bostanabad-Research-Group/GP-Plus}{GitHub Repository: https://github.com/Bostanabad-Research-Group/GP-Plus}}}
\affil[1]{Department of Mechanical and Aerospace Engineering, University of California, Irvine}

\begin{document}
    \pagenumbering{arabic}
    \sloppy
    \maketitle
    
    \noindent \textbf{Abstract}\\
In this paper we introduce \gp, an open-source library for kernel-based learning via Gaussian processes (GPs) which are powerful statistical models that are completely characterized by their parametric covariance and mean functions. \gp~is built on PyTorch and provides a user-friendly and object-oriented tool for probabilistic learning and inference. As we demonstrate with a host of examples, \gp~has a few unique advantages over other GP modeling libraries. We achieve these advantages primarily by integrating nonlinear manifold learning techniques with GPs' covariance and mean functions. As part of introducing \gp, in this paper we also make methodological contributions that $(1)$ enable probabilistic data fusion and inverse parameter estimation, and $(2)$ equip GPs with parsimonious parametric mean functions which span mixed feature spaces that have both categorical and quantitative variables. We demonstrate the impact of these contributions in the context of Bayesian optimization, multi-fidelity modeling, sensitivity analysis, and calibration of computer models. 

    \section{Introduction} \label{sec: intro}
Gaussian processes (GPs) are indispensable building blocks of many powerful probabilistic frameworks such as Bayesian optimization (BO) \cite{RN578, RN1242, RN1375, RN1392, RN1576, RN1577, RN1632, RN1761, RN1789, RN1794, RN1845}, function and operator learning \cite{RN332, RN1881, RN1886, RN1890, RN1559, RN1558}, data-driven calibration of expensive simulations \cite{RN271, RN272, RN334, RN337, RN705, RN841}, and multi-fidelity (MF) modeling \cite{RN1266, RN1275, RN1838, RN1786, RN791}. The recent software and hardware developments, combined with the new means of data collection and the society's drive to tackle ever-challenging goals, have sparked significant innovations in the broad field of machine learning (ML). GPs have also substantially benefited from these advancements and recent works have enabled them to leverage GPUs \cite{RN1894, RN1270} and to accommodate high dimensions \cite{RN321, RN530, RN713, RN1228, RN1412}, large datasets \cite{RN718, RN893, RN503, RN698, RN818, RN1574, RN1479, RN1573, RN347}, or disjoint feature spaces that have both qualitative and quantitative variables \cite{RN1075, RN1215, RN1214, RN1492, RN1559}.
In line with these advancements, in this paper we introduce \gp~which is a modular and user-friendly library that aims to empower researchers and practitioners in harnessing the full potential of GPs for a wide range of applications such as {\color{black}emulation (i.e., probabilistic metamodeling)}, single- and multi-fidelity Bayesian optimization (SFBO and MFBO), kernel-based generalized MF modeling, inverse parameter estimation, anomaly detection, and sensitivity analyses. 

As reviewed in \Cref{sec: related-work}, there are quite a few existing libraries for GP modeling and in fact we leverage one of them (i.e., \gtorch~\cite{RN1270}) in developing ours. While these libraries have been successfully used in many applications across sciences and engineering, we believe a few distinct features set \gp~aside. 
{\color{black}First}, we design parametric covariance and prior mean functions that automatically enable GPs to fuse and emulate multi-source datasets, detect anomalies, find calibration parameters of computer models (i.e., inverse parameter estimation), and handle categorical variables. These functions leverage kernel-based nonlinear manifold learning, provide interpretable solutions (see \Cref{sec: functionalities} for multiple examples), and their parameters are all jointly learnt via the maximum a posteriori (MAP) method. 
{\color{black}Second}, we provide a unified platform to use GPs for many tasks such as SFBO and MFBO, probabilistic regression, and sensitivity analysis. As shown in \Cref{sec: functionalities}, all of these functionalities are achieved via a few lines of codes. As an example, we develop a hyperparameter estimation routine based on the method of continuation \cite{RN1062} which, at the expense of slightly higher computational costs, provides more numerical stability and accuracy. This method is accessible by merely changing the default optimization settings in \gp. 
Lastly, \gp~is accompanied with a rich set of datasets (from engineering applications) and benchmark analytic examples that can be used by researchers in evaluating the performance of emulation techniques beyond GPs.

The rest of this paper is organized as follows. In \Cref{sec: gp-background} we provide a brief background on GPs and then review the relevant literature in \Cref{sec: related-work}. 
We introduce the primary components of \gp~(i.e., its covariance and prior mean functions) in \Cref{sec: GP_Plus} where we also introduce the three new methodological innovations of this paper. 
The first contribution is focused on the kernel and endows GPs with probabilistic embeddings that benefit both MF modeling (in terms of quantifying model-form errors) and mixed data emulation (in terms of learning the relations among the categorical variables and their levels). 
The second contribution is on designing parametric mean functions such that they naturally handle multi-source or mixed data that have categorical features. 
The third contribution is on inverse parameter estimation where the goal is to probabilistically calibrate computer models using limited high-fidelity (HF) data such as observations or experiments. 
We provide some details and examples on some of the most important functionalities of \gp~ in \Cref{sec: functionalities} where we also conduct comparative studies against existing methods. 
Concluding remarks and future research directions are provided in \Cref{sec: conclusion}.

\subsection{Nomenclature}
Unless otherwise stated, throughout the paper we denote scalars, vectors, and matrices with regular, bold lower-case, and bold upper-case letters, respectively (e.g., $x, \xb,$ and $\Xb$). Vectors are by default column vectors and subscript or superscript numbers enclosed in parenthesis indicate sample numbers. For instance, $x^{(i)}$ or $\xb^{(i)}$ denote the $i^{th}$ sample in a training dataset while $x_i$ indicates the $i^{th}$ component of the column vector $\xb = \brackets{x_1, \cdots, x_{dx}}^T$. For clarity, we sometimes indicate the size of vectors and matrices via subscripts, e.g., $\xb_n$ and $\Xb_{nq}$. Specifying the size is useful in cases where we do not follow our notational convention that distinguishes vectors and matrices (e.g., $\Yb_q$ is a vector while $\Yb$ is a matrix). 

Lastly, we distinguish between a function and samples taken from that function by specifying the functional dependence. As an example, $y(x)$ and $y(\xb)$ are functions while $y$ and $\yb$ are a scalar and a vector of values, respectively. 
We also assume functions accommodate batch computations. That is, a single-response function returns a column vector of $n$ values if $n$ inputs are simultaneously fed into it, i.e., $\yb = y(\Xb)$.
    \section{Background on Gaussian Processes} \label{sec: gp-background}
To explain the working principles of GPs, we consider $\Yb_q$ and $\Yb_{n}$ which are two jointly normal random vectors of sizes $q$ and $n$, respectively\footnote{For this part of the description, we distinguish between the random vector $\Yb_q$ and the specific realization $\yb_{q}$ that it takes.}. We write this joint distribution as:
\begin{equation} 
    \begin{split}
        p\left( \left[\begin{array}{l}
        \Yb_q \\
        \Yb_n
        \end{array}\right] \right)
        = \mathcal{N}_{q+n}\left(\left[\begin{array}{c}
        \mub_{q} \\
        \mub_{n}
        \end{array}\right],\left[\begin{array}{cc}
        {\Sigmab}_{qq} & \Sigmab_{qn} \\
        \Sigmab_{nq} & \Sigmab_{nn}
        \end{array}\right]\right)
    \end{split}
    \label{eq: mvn}
\end{equation}
where the subscripts indicate the array sizes, $\mub_q = \E[\Yb_q]$, $\Sigmab_{qq} = cov(\Yb_q) = \E[(\Yb_q - \mub_q)(\Yb_q - \mub_q)^T]$, and $\Sigmab_{qn} = \Sigmab_{nq}^T = \E[(\Yb_q - \mub_q)(\Yb_n - \mub_n)^T]$. 
If $\yb_n$ is observed for $\Yb_n$, we can update our knowledge on $\Yb_q$ based on its conditional distribution:
\begin{equation} 
    \begin{split}
    p\left( \Yb_q \mid \Yb_n = \yb_n \right) = \mathcal{N}_{q}\left({\mub^*_q}, \Sigmab^*_{qq}\right)
    \end{split}
    \label{eq: conditional-mvn}
\end{equation}
where $\mub^*_q =\mub_q + \Sigmab_{qn} \Sigmab^{-1}_{nn} (\yb_n - \mub_n)$ is the conditional mean vector and $\Sigmab^*_{qq} = \Sigmab_{qq} - \Sigmab_{qn} \Sigmab^{-1}_{nn} \Sigmab_{nq}$ is the conditional covariance matrix. 
Similarly, in interpolation with GPs one first assumes that the given data $\yb_n$ and the to-be-pblackicted values $\yb_q$ are jointly Gaussian and then infers the latter via \Cref{eq: conditional-mvn}. However, the mean vectors and covariance matrices in \Cref{eq: mvn} are unknown and thus \Cref{eq: conditional-mvn} can be used for pblackiction only after we $(1)$ endow the underlying GP with a parametric mean function and a parametric covariance function (or kernel), and $(2)$ estimate the parameters of these two sets of functions\footnote{In a fully Bayesian setting, instead of estimating the parameters, their posterior distributions are obtained and pblackictions on $\yb_q$ require marginalization with respect to these distributions. Due to the significantly higher computational costs of fully Bayesian techniques and their marginal accuracy improvements in the case of GPs, we recommend and use MAP.}.


More formally, assume the training dataset $\braces{\xb^{(i)}, y^{(i)}}_{i=1}^n$ is given where $\xb = [x_1, ..., x_{dx}]^T \in \xsspace \subset \rsspace^{dx}$ and $y^{(i)} = y(\xb^{(i)}) \in \rsspace$ denote the inputs and response\footnote{We focus on regression problems whose output dimensionality is one but note that GPs can handle multi-response or multi-task problems as well \cite{RN1902, RN328, RN287}.}, respectively. Given $\yb = [y^{(1)}, \cdots, y^{(n)}]^T$ and $\Xb$ whose $i^{th}$ row is $(\xb^{(i)})^T$, our goal is to pblackict $y(\xb^*)$ at the arbitrary point $\xb^* \in \xsspace$.
Following the above description, we assume $\yb = [y^{(1)}, \cdots, y^{(n)}]^T$ is a realization of a GP with the following parametric mean and covariance functions:
\begin{subequations} 
    \begin{equation} 
        \begin{split}
            \mathbb{E}[y(\xb)] = m(\xb; \betab),
        \end{split}
        \label{eq: gp-mean}
    \end{equation}
    \begin{equation} 
        \begin{split}
            \text{cov}\left(y(\xb), y(\xb')\right) = c(\xb, \xb'; \sigma^2, \thetab) = \sigma^2 r(\xb, \xb'; \thetab)
        \end{split}
        \label{eq: gp-cov}
    \end{equation}
    \label{eq: gp-mean and cov}
\end{subequations} 
where $\betab$ and $\thetab$ are the parameters of the mean and covariance functions, respectively. 
The mean function in \Cref{eq: gp-mean} can take on many forms such as polynomials or even a feedforward neural network (FFNN). In many applications of GP modeling, a constant value is used as the mean function (i.e., $m(\xb; \betab) = \beta$) in which case the performance of the GP depends entirely on its kernel.
In \Cref{eq: gp-cov}, $\sigma^2$ is the process variance (or inverse precision) and $r(\cdot, \cdot)$ is the correlation function whose parameters are collectively denoted via $\thetab$.
Common choices for $r(\cdot, \cdot)$ are the Gaussian, power exponential, and Mat\'ern correlation functions defined as:
\begin{subequations} 
    \begin{equation} 
        r(\boldsymbol{x}, \boldsymbol{x}^{\prime}; \omegab)=\exp \left\{-\sum_{i=1}^{dx} 10^{\omega_{i}}(x_{i}-x_{i}^{\prime})^{2}\right\}
        \label{eq: gaussian-kernel}
    \end{equation}
    \begin{equation} 
        r(\boldsymbol{x}, \boldsymbol{x}^{\prime}; \omegab, p)=\exp \left\{-\sum_{i=1}^{dx} 10^{\omega_{i}}|x_{i}-x_{i}^{\prime}|^p\right\}
        \label{eq: power-kernel}
    \end{equation}
    \begin{equation} 
        r(\boldsymbol{x}, \boldsymbol{x}^{\prime}; \omegab) = \frac{2^{1-\nu}}{\Gamma(\nu)}K_{\nu}\left(\sqrt{2\nu}\times \sum_{i=1}^{dx} 10^{\omega_{i}}(x_{i}-x_{i}^{\prime})^{2}\right)^{1+\nu} 
        \label{eq: matern-kernel}
    \end{equation}
    \label{eq: all-kernels}
\end{subequations} 
where $\omega_i \in \color{black}{\rsspace}$\footnote{To ensure numerical stability, $\omega_i$ is typically bounded to a subset of $\color{black}{\rsspace}$, e.g., $[-10, 4]$.}, $p \in [1, 2]$, $\nu \in \{\frac{1}{2},\frac{3}{2}, \frac{5}{2} \}$, $K_{\nu}$ is the modified Bessel function of the second kind, and $\Gamma$ is the gamma function. 
The inductive bias that the kernels in \Cref{eq: all-kernels} encode into the learning process is that close-by input vectors $\xb$ and $\xb^{\prime}$ have similar (i.e., correlated) output values. The degree of this correlation is quantified by the interpretable length-scale (aka roughness) parameters where the magnitude of $10^{\omega_i}$ is directly related to the response fluctuations along $x_i$.

Having defined these kernels we can now write the likelihood function of the observation vector $\yb$ as:
\begin{equation} 
    p(\yb; \betab, \sigma^2, \thetab) = 
    (2 \pi )^{-\frac{n}{2}}|\Cb|^{-\frac{1}{2}} 
    \times \exp \left\{\frac{-1}{2}(\yb-\mb)^T\Cb^{-1}(\yb-\mb)\right\} 
    \label{eq: likelihood}
\end{equation}
where $|\cdot|$ denotes the determinant operator, $\Cb_{nn}:=c(\Xb, \Xb; \sigma^2, \thetab)$ is the covariance matrix whose $(i, j)^{t h}$ element is $C_{i j} = c(\xb^{(i)}, \xb^{(j)}; \sigma^2, \thetab) = \sigma^2r(\xb^{(i)}, \xb^{(j)}; \thetab)$, and $\mb$ is an $n \times 1$ vector whose $i^{th}$ element is $m_i=m(\xb^{(i)}; \betab)$. 
The point estimates for $\betab, \sigma^2,$ and $ \thetab$ can now be found by maximizing the likelihood function in \Cref{eq: likelihood}.
Alternatively, Bayes' rule can be used to leverage prior knowledge in estimating these parameters. Specifically, the joint posterior distribution of the parameters is:
\begin{equation} 
    p(\betab, \sigma^2, \thetab \mid \yb) = 
    \frac{p(\yb \mid  \betab, \sigma^2, \thetab) p(\betab, \sigma^2, \thetab)}{p(\yb)},
    \label{eq: bayes_rule}
\end{equation}
where $p(\yb)$ is the evidence. Since $p(\yb)$ is a normalizing constant, we can find the MAP estimates of $\betab, \sigma^2,$ and $ \thetab$ by maximizing the right-hand-side of \Cref{eq: bayes_rule}. That is: 
\begin{equation} 
    \begin{split}
        [\widehat{\betab}, \widehat{\sigma^2}, \widehat{\thetab}] = 
        \underset{\betab, \sigma^2, \thetab}{\operatorname{argmax}}\left|2 \pi \Cb\right|^{-\frac{1}{2}} \times 
        \exp \left\{\frac{-1}{2}(\yb-\mb)^T \Cb^{-1}(\yb-\mb)\right\}
        \times p(\betab, \sigma^2, \thetab)
    \end{split}
    \label{eq: map-max}
\end{equation}
or equivalently:
\begin{equation} 
    \begin{split}
        [\widehat{\betab}, \widehat{\sigma}^2, \widehat{\thetab}] = \underset{\betab, \sigma^2, \thetab}{\operatorname{argmin}} \hspace{2mm} {L_{MAP}}=
        \underset{\betab, \sigma^2, \thetab}{\operatorname{argmin}} \hspace{2mm} \frac{1}{2} \log (|\Cb|)+\frac{1}{2}(\yb-\mb)^T \Cb^{-1}(\yb-\mb)- \log\left(p(\betab, \sigma^2, \thetab)\right)
    \end{split}
    \label{eq: map-gp}
\end{equation}
where $\log (\cdot)$ denotes the natural logarithm.
We can now efficiently estimate all the model parameters by minimizing \Cref{eq: map-gp} via a gradient-based optimization algorithm\footnote{Since the profile of the objective function in \Cref{eq: map-gp} has many local minima, it is important to start the gradient-based optimization via multiple initial guesses. We control this setting in \gp~via the \texttt{num\_restarts} parameter whose default value is $32$.} and then adopt \Cref{eq: conditional-mvn} to obtain the mean and variance of the response distribution at the arbitrary point $\xb^*$:
\begin{subequations} 
    \begin{equation} 
        \E[y(\xb^*)] = \mu(\xb^*) = 
        m(\xb^*; \widehat{\betab}) + c(\xb^*, \Xb; \widehat{\thetab}, \widehat{\sigma}^2) \boldsymbol{C}^{-1}(\yb-\mb)
        \label{eq: gp-mean-scalar}
    \end{equation}
    \begin{equation} 
        \text{cov}(y(\xb^*), y(\xb^*)) = \tau^2(\xb^*)=
        c(\xb^*,\xb^*; \widehat{\thetab}, \widehat{\sigma}^2) - c(\xb^*, \Xb; \widehat{\thetab}, \widehat{\sigma}^2) \Cb^{-1} c(\Xb, \xb^*; \widehat{\thetab}, \widehat{\sigma}^2))
        \label{eq: gp-var-scalar}
    \end{equation}
    \label{eq: gp-mean-var-scalar}
\end{subequations}    
where $c(\xb^*, \Xb; \widehat{\thetab},\widehat{\sigma}^2)$ is a $1 \times n$ row vector with entries $c_i = c(\xb^*, \xb^{(i)}; \widehat{\thetab},\widehat{\sigma}^2)$ and its transpose is $c(\Xb, \xb^*; \widehat{\thetab},\widehat{\sigma}^2)$.
\Cref{eq: gp-mean-scalar,eq: gp-var-scalar} can be straightforwardly extended to pblackict the response distribution for  a batch of samples denoted by $\Xb^*$:
\begin{subequations} 
    \begin{equation} 
        \E[y(\Xb^*)] = 
        m(\Xb^*; \widehat{\betab}) + c(\Xb^*, \Xb; \widehat{\thetab}, \widehat{\sigma}^2) \Cb^{-1}(\yb-\mb)
        \label{eq: gp-mean-vec-noise}
    \end{equation}
    \begin{equation} 
        \text{cov} \textcolor{black}{\parens{y\parens{\Xb^*}, y(\Xb^*)}}=
        c(\Xb^*, \Xb^*; \widehat{\thetab}, \widehat{\sigma}^2) -
        c(\Xb^*, \Xb; \widehat{\thetab}, \widehat{\sigma}^2) \Cb^{-1} c(\Xb, \Xb^*; \widehat{\thetab}, \widehat{\sigma}^2)
        \label{eq: gp-var-vec}
    \end{equation}
    \label{eq: gp-mean-var-vec}
\end{subequations}    
The above formulations build interpolating GPs. To handle datasets with noisy observations, the nugget or jitter parameter, denoted by $\delta$ \cite{RN1917, RN332, RN1908}, is used where $\Cb$ is replaced by $\Cb_{\delta}= \Cb + \delta \Ib_{nn}$ where $\Ib_{n×n}$ is the $n \times n$ identity matrix (with this adjustment, the stationary noise variance estimated by the GP is $\widehat\delta$)\footnote{Some recent works \cite{RN292, bostanabad2018leveraging, macdonald2015gpfit} apply the nugget directly to $\Rb$ but herein we adhere to \cite{RN1917} and add $\delta$ to $\Cb$.}.
In addition to modeling stationary noise, the nugget parameter is also used to mitigate the numerical issues associated with $\Cb$. That is, even with noise-free $\yb$, $\Cb_{\delta}$ is used while minimizing \Cref{eq: map-gp} to ensure the correlation matrix is always invertible. 
When the nugget parameter is used for fitting a GP to noisy observations, \Cref{eq: mvn} takes on the following form:
\begin{equation} 
    \begin{split}
        p\left( \left[\begin{array}{l}
        \Yb_q \\
        \Yb_n
        \end{array}\right] \right)
        = \mathcal{N}_{q+n}\left(\left[\begin{array}{c}
        \mub_{q} \\
        \mub_{n}
        \end{array}\right],\left[\begin{array}{cc}
        {\Sigmab}_{qq}+\delta \Ib_{qq} & \Sigmab_{qn} \\
        \Sigmab_{nq} & \Sigmab_{nn} + \delta \Ib_{nn}
        \end{array}\right]\right)
    \end{split}
    \label{eq: mvn-noise}
\end{equation}
which means that \Cref{eq: gp-mean-var-vec} should be updated to:
\begin{subequations} 
    \begin{equation} 
        \E[y(\Xb^*)] = 
        m(\Xb^*; \widehat{\betab}) + c(\Xb^*, \Xb; \widehat{\thetab}, \widehat{\sigma}^2) \Cb^{-1}_{\delta}(\yb-\mb)
        \label{eq: gp-mean-vec-noise}
    \end{equation}
    \begin{equation} 
        \text{cov} \parens{y\parens{\Xb^*}, y(\Xb^*)} =
        c(\Xb^*, \Xb^*; \widehat{\thetab}, \widehat{\sigma}^2) -
        c(\Xb^*, \Xb; \widehat{\thetab}, \widehat{\sigma}^2) \Cb^{-1}_{\delta} 
        c(\Xb, \Xb^*; \widehat{\thetab}, \widehat{\sigma}^2) + \widehat\delta\Ib.
        \label{eq: gp-var-vec-noise}
    \end{equation}
    \label{eq: gp-mean-var-vec-noise}
\end{subequations}    

We highlight that \Cref{eq: gp-var-vec-noise} does not consider the additional uncertainties incurblack by estimating the parameters of the mean and covariance functions (note that \Cref{eq: conditional-mvn} assumed the mean vector and covariance matrices are known). These additional uncertainties can be quantified by building and using the GP within a Bayesian framework where sampling methods (e.g., Markov Chain Monte Carlo or MCMC) are requiblack for marginalization as closed-form expressions are only available for specific cases (see \cite{RN1896} for an example). Since such sampling methods are typically expensive and the provided benefits are marginal, MAP is frequently employed in GP modeling {\color{black}\cite{murphy2012machine}}.
    \section{Related Works} \label{sec: related-work}

Many open-source GP libraries have been recently developed and in this section we review some of the most well-known ones. 
One of the earliest open-source GP packages is TreedGP \cite{gramacy2007tgp} which is developed in R and primarily aims to address the stationarity and scalability issues of GPs. In particular, TreedGP recursively partitions the input space via parallel and axes-aligned boundaries \cite{chipman1998bayesian} and then endows each partition with a GP whose covariance function is stationary. TreedGP uses Bayesian averaging to combine these GPs which is particularly important for obtaining smooth predictions on the partition boundaries. The major limitations of TreedGP are its non-differentiability on the boundaries, high computational costs (as the Bayesian analyses rely on MCMC \cite{harkonen2022mixtures}), reliance on trees which can only partition the input space with axis-aligned boundaries \cite{candelieri2021treed}, and inability to efficiently handle categorical features in small-data applications. 

\begin{wrapfigure}{R}{0.4\textwidth}
    \includegraphics[width=1\textwidth]{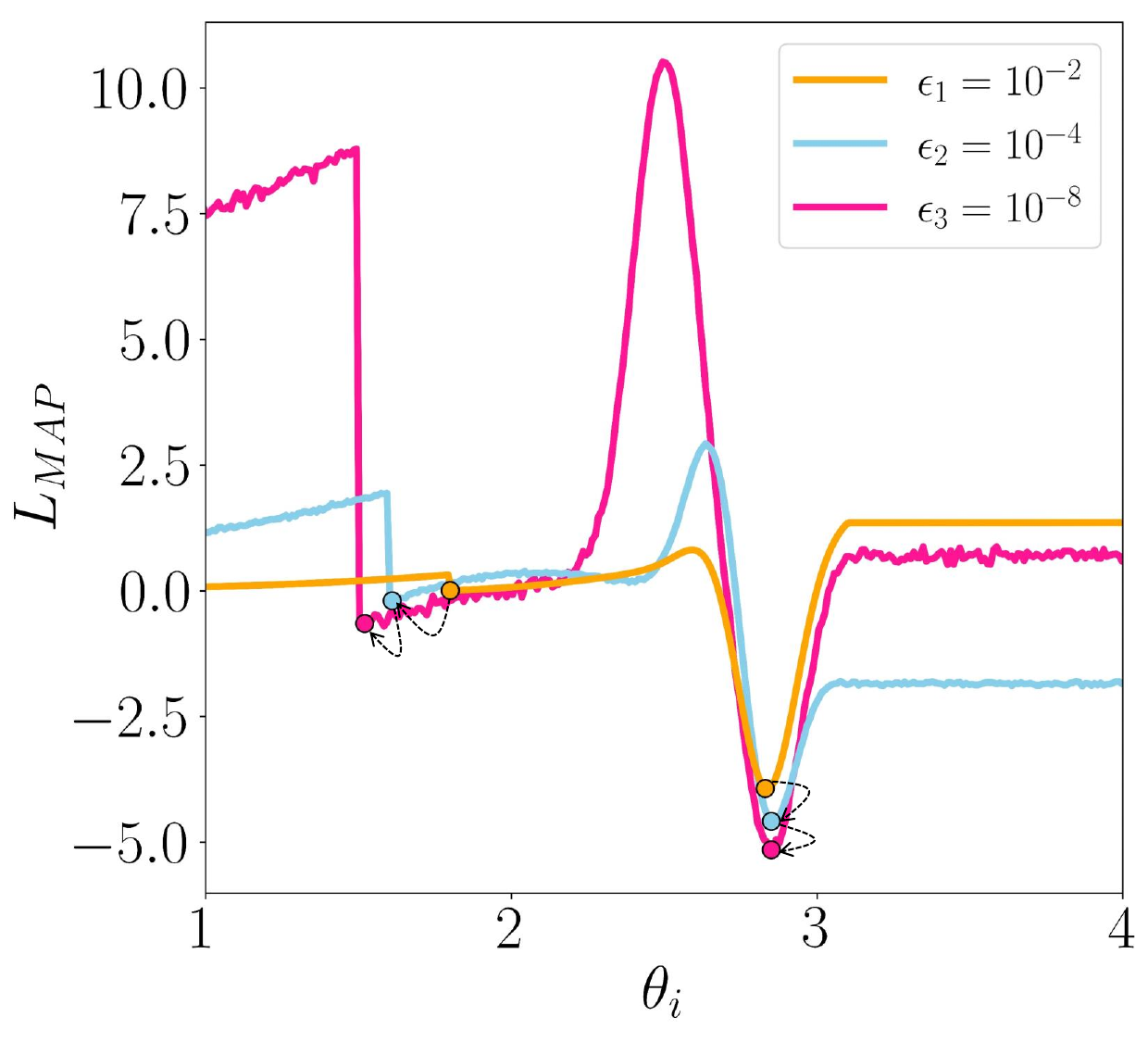}
    \caption{\textbf{Schematic illustration of continuation-based optimization:} The profile of $L_{MAP}$ in \Cref{eq: map-gp} is smoothed with a larger $\epsilon$ (or, equivalently, a larger nugget).}
    \label{fig: continuation}
\end{wrapfigure}

GPfit \cite{macdonald2015gpfit} and GPM \cite{bostanabad2018leveraging} are also R packages and they are primarily designed to improve the hyper-parameter optimization process at the expense of increased computational costs. GPfit has a multi-step pre-processing stage that aims to improve the quality of the initial points that are used via L-BFGS\footnote{Limited-memory Broyden–Fletcher–Goldfarb–Shanno} in minimizing \Cref{eq: map-gp}. 
Unlike GPfit, GPM develops a multi-step continuation-based strategy to increase both the robustness and accuracy of the optimization process. In particular, GPM indirectly controls $\delta$ via the auxiliary parameter $\epsilon$ that puts a lower bound on the smallest eigenvalue of the correlation matrix\footnote{The rationale behind this choice is that the smallest eigenvalue of $\Rb$ can sometimes be negative due to numerical issues.}. GPM first uses a large value for $\epsilon$ (e.g., $10^{-2}$) and minimizes \Cref{eq: map-gp} while requiring the smallest eigenvalue of $\Rb$ to always be larger than the imposed $\epsilon$. In addition to guaranteeing numerical robustness, this requirement dramatically smooths the profile of the objective function and hence most (if not all) optimizations quickly converge to the same solution. Then, GPM relaxes the constraint on $\Rb$ (e.g., $\epsilon = 10^{-3}$) and repeats the optimization while using the solution(s) of the previous step as the initial guess(es) in the current step, see \Cref{fig: continuation}. These steps are continued until the minimum value of $\epsilon$ is reached and then the parameters of the final GP are chosen by identifying the step (or $\epsilon$) at which the leave-one-out cross-validation (LOO-CV) error of the model is minimized. 
GPy \Cite{gpy2014} is a popular object-oriented library that is implemented based on numeric Python (NumPy) by the Sheffield machine learning group. GPy provides a number of basic and advanced functionalities for GP regression that include multi-output learning and non-Gaussian likelihood functions which are accompanied with Laplace approximation \cite{RN509} and expectation propagation since exact inference with non-Gaussian likelihoods is not tractable. 
However, GPy does not fully leverage modern hardware capabilities (e.g., GPU acceleration) and integration with deep neural networks (NNs) which are increasingly crucial in contemporary GP applications. It also lacks some of the most recent advancements that enable GPs to accommodate high dimensions or categorical features. 

One of the first open-source GP libraries that supports GPU acceleration and leverages automatic differentiation is GPflow \Cite{RN1894} which is based on Tensorflow and has an object-oriented Python front-end. GPflow supports regression and classification problems, uses variationally sparse methods for scalability to large data, and provides both Bayesian and point-estimate-based inference classes for Gaussian and non-Gaussian likelihoods. While GPflow has significant capabilities, it lacks some of the key recent advancements in GPs such as natural integration with SFBO or MFBO frameworks, fusing multi-source data, calibrating unknown parameters, or directly supporting categorical variables. 

Perhaps the most widely used open-source package for GP modeling is GPytorch \Cite{RN1270} which accommodates a wealth of functionalities such as deep kernel \cite{RN1901} and multi-task/multi-output learning \cite{RN1902, RN328, RN287}, dimensionality reduction via latent variable GPs \cite{RN1903, RN1904}, and variational and approximate modeling for handling non-Gaussian likelihoods (e.g., in classification) or large datasets. A particular feature of GPytorch is its use of preconditioned conjugate gradients in accelerating the expensive computations (i.e., matrix inversion and determinant calculation) associated with exact GPs in \Cref{eq: map-gp,eq: gp-mean-var-vec}. This feature is coined blackbox matrix-matrix multiplication (BBMM) and is uniquely suited for GPU-based computations. GPytorch relies on Pyro \cite{bingham2018pyro} for probabilistic programming, forms the backbone of BoTorch which is an open-source Python library for BO, and has specialized kernels for MF modeling or handling categorical variables. 
Given the versatility and robustness of GPytorch, we use some of its core functionalities to build \gp~while addressing the limitations of GPytorch in handling categorical features, directly solving inverse problems, learning from MF datasets, or interfacing with SFBO/MFBO engines.

Other notable open-source GP libraries that we mention in passing include those of Ambikasaran et al. \cite{RN1898} which hierarchically factors the covariance matrix\footnote{As long as it is built with specific covariance functions such as the Gaussian or Mat\'ern.} into a product of block low-rank updates of the identity matrix to accelerate matrix inversion and determinant calculation (their method loses accuracy for $dx>3$), Vanhatalo et al. \cite{RN1900} which is a Matlab library that integrates various elementary computational tools (e.g., sparse approximation) for GP-based regression, and GPML \cite{RN332} which is also a Matlab library and has been widely used in a wide range of applications.

    \section{Kernel-based Learning} \label{sec: GP_Plus}
The \textit{vanilla} GP formulations reviewed in \Cref{sec: gp-background} break down in high dimensions or with large samples \textcolor{black}{\cite{hensman2013gaussian,bengio2005curse}}, do not directly accommodate MF modeling or MFBO \textcolor{black}{\cite{eweis2022data,foumani2023multi}}, and cannot handle categorical features \textcolor{black}{\cite{oune2021latent}}. Since the scalability issue of GPs is rigorously studied in many recent works, in \gp~we focus on holistically addressing the latter two limitations based on the ideas that were first introduced in \cite{oune2021latent}. In particular, we generalize the concept of kernel-based learning for GPs by introducing new bases and kernels with customized parametric functions that directly enable probabilistic learning from multi-source data and handling qualitative features. Compablack to existing works that also develop new kernels for GPs (see for example \cite{tao2021latent,bonilla2007multi,poloczek2017multi} for handling categorical inputs, building multi-response emulators, and MF modeling), our functions are quite versatile and produce nonlinearly learned embeddings that, while being low-dimensional and highly interpretable, enable GPs to model more complex relations.

\begin{figure}[!h]
    \centering
        \includegraphics[width=1\linewidth]{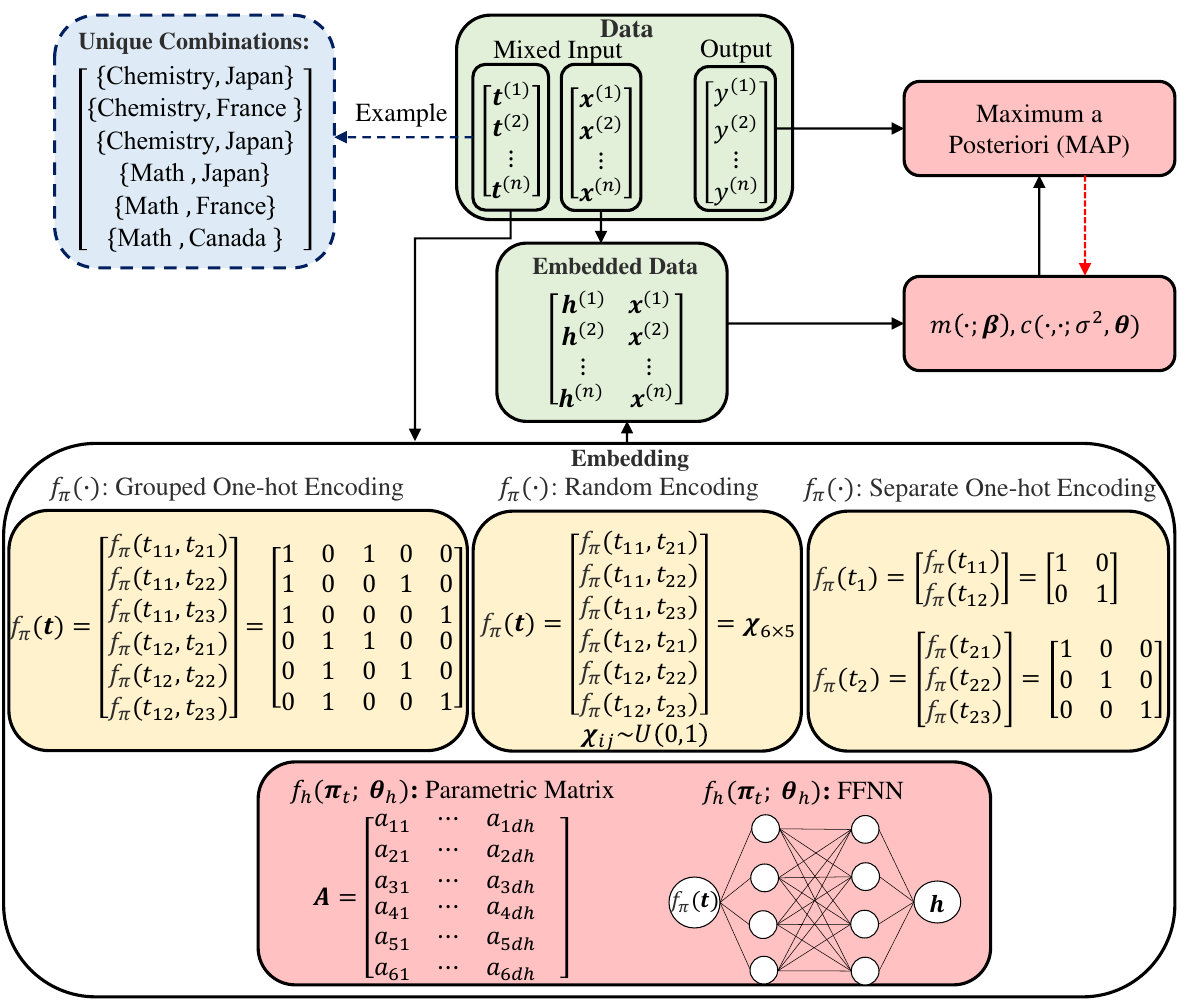}
    \vspace{-.7cm}
    \caption{\textbf{Emulation via \gp~ in mixed input spaces:} We first endow the categorical variables $\tb$ with some quantitative prior representations which are then mapped to a low-dimensional embedding with a parametric function. The embedded variables $\hb$ are then concatenated with $\xb$ and fed into the mean and covariance functions. All the model parameters are jointly learnt via MAP. }
    \label{fig: emulation-flowchart}
\end{figure}

To explain our kernel-based approach, we consider an emulation scenario where the input space includes two qualitative features $ t_1=\{Math, Chemistry\} $ and $ t_2=\{Japan, France, Canada\}$ which have $l_1=2$ and $l_2=3$ levels, respectively. Vanilla GPs cannot directly work with $\tb = [t_1, t_2]^T$ since typical kernels such as those in \Cref{eq: all-kernels} require each feature to be associated with a distance metric while categorical variables naturally lack such measures. 
As schematically illustrated in \Cref{fig: emulation-flowchart}, we address this limitation by first endowing the categorical variables $\tb = [t_1, \cdots, t_{dt}]^T$ with the quantitative prior representations $\pib_t = f_{\pi}(\tb)$ where $f_{\pi}(\cdot)$ is a deterministic user-specified function. These priors are typically high dimensional (i.e., $d\pi>dt$) and can be designed in many ways (we describe some of these below, see {\color{black}supplementary comments on our GitHub
page} for more options). To blackuce the dimensionality of these representations while learning the effects of $\tb$ on the response, we then pass $\pib_t$ through the parametric embedding function $f_h(\pib_t; \thetab_h)$ to obtain $\hb$ which is a $dh$ dimensional \textit{latent} representation of $\tb$ where $d\pi \gg dh$. Since $\hb = f_h(f_\pi(\tb); \thetab_h)$ are quantitative, they can be easily used to develop new kernels. For instance, we can extend the Gaussian and Mat\'ern correlation functions as:
\begin{subequations} 
    \begin{equation} 
        \begin{split}
            r\left(\ub, \ub^{\prime}; \omegab, \thetab_h\right) = 
            \exp \left\{-\sum_{i=1}^{dx} 10^{\omega_i}(x_i-x_i^{\prime})^2 -
            \sum_{i=1}^{dh}(h_i - h^{\prime}_i)^2 \right \} 
        \end{split}
        \label{eq: gaussian-kernel-GP_Plus}
    \end{equation}
    \begin{equation} 
        \begin{split}
            r\left(\ub, \ub^{\prime}; \omegab, \thetab_h\right) = 
            \frac{2^{1-\nu}}{\Gamma(\nu)}K_{\nu}
            \left(\sqrt{2\nu}\times \sum_{i=1}^{dx} 10^{\omega_{i}}(x_{i}-x_{i}^{\prime})^{2} + \sum_{i=1}^{dh}(h_i - h^{\prime}_i)^2 \right)^{1+\nu} 
        \end{split}
        \label{eq: matern-kernel-GP_Plus}
    \end{equation}
    \label{eq: all-kernels-GP_Plus}
\end{subequations}
where $\ub = \left[ \begin{array}{l} \xb \\ \tb \end{array} \right]$. We note that $(1)$ no scale parameters are associated with $\hb$ in \Cref{eq: gaussian-kernel-GP_Plus,eq: matern-kernel-GP_Plus} since, as opposed to $\xb$, $\hb$ are learnt, and $(2)$ $\thetab_h$ are estimated jointly with the other parameters of the GP via MAP where the covariance matrix in \Cref{eq: map-gp} is now built via one of the correlation functions in \Cref{eq: all-kernels-GP_Plus}.

As we explain in the proceeding subsections, the above kernel reformulations not only allow GPs to operate in feature spaces with categorical variables, but they also enable GPs to directly fuse MF datasets (from an arbitrary number of sources) or inversely estimate calibration parameters. Given this general applicability of our approach, we have equipped \gp~with various mechanisms to design the priors and parameterize the embeddings. We believe these options increase the interpretability of the model (in particular, the learnt embeddings) as well as computational efficiency.

In \Cref{fig: emulation-flowchart} we schematically demonstrate a few options for designing $\pib_t$ and the embedding functions for an emulation example where the feature space has quantitative variables $\xb$ and the two categorical variables $\boldsymbol{t}=[t_1,t_2]$ mentioned above. 
As shown in the top row of the embedding block, $f_{\pi}(\cdot)$ can simply be a deterministic bijective\footnote{Surjective and injective functions may also be used especially if some prior knowledge encourages such choices. We focus on bijective functions in this paper and leave other choices for future studies.} function that $(1)$ groups the one-hot-encoded representations of $\tb$ into a single matrix (this option is the default in \gp), $(2)$ builds a random matrix whose unique rows correspond to the unique combinations of $\tb$, or $(3)$ constructs multiple matrices where each one corresponds to the one-hot-encoding of one of the categorical variables. 
The second row in the embedding block of \Cref{fig: emulation-flowchart} illustrates two options for $f_h(\pib_t; \thetab_h)$ that consist of parametric matrices (denoted by $\Ab$ in \Cref{fig: emulation-flowchart}) and FFNNs. The construction of $f_h(\pib_t; \thetab_h)$ is affected by $f_{\pi}(\cdot)$. For instance, the row size of $\Ab$ depends on $\pib_{t}$\footnote{Hence, the random prior encoding can work with a smaller $\Ab$ compablack to the grouped one-hot-encoding.} while its column size is chosen by the user and determines the dimensionality of the to-be-learnt embedding. The number of $\Ab$ matrices also depends on the priors since it should match with the number of matrices that $f_{\pi}(\cdot)$ generates. We note that these dependencies are automatically enforced in \gp~and the available options for $f_{\pi}(\cdot)$ or $f_h(\pib_t; \thetab_h)$ in \gp~can be easily accessed by changing its default settings. 

Once the parametrized embedding is constructed with any of the procedures described above (or other settings available in \gp), they are concatenated with the numerical features and used in our reformulated mean and covariance functions to build the likelihood function. Then, all the model parameters are estimated via MAP. 
In the following subsections, we elaborate on how these embeddings as well as reformulated mean and covariance functions benefit MF modeling, inverse parameter estimation, and MFBO.

\subsection{Multi-fidelity Modeling via Deterministic Embedding} \label{subsec: det-mfmodeling}
The premise of MF modeling is to leverage low-fidelity (LF) data to blackuce the reliance on expensive high-fidelity (HF) samples in many-query applications such as design optimization \cite{chakraborty2017surrogate,korondi2021multi}, uncertainty propagation and variance blackuction \cite{dixon2023covariance}, BO, calibration of computer models \cite{absi2016multi}, and sensitivity analysis \cite{RN452}. With the exception of a few recent works such as \cite{RN1770, RN1876, RN1845}, most existing MF techniques fuse only two data sources while imposing a specific functional relation between them. For instance, the method of Kennedy and O’Hagan (KOH) \cite{RN705} and its various extensions \cite{RN648, RN272, RN271, RN337, RN344, RN335, RN336, RN334, RN270} fuse the HF and LF data based on the following generic relation:
\begin{equation} 
    \begin{split}
        y_h(\xb) = \rho \times y_l(\xb, \zetab) + y_b(\xb, \zetab) + \varepsilon
    \end{split}
    \label{eq: koh}
\end{equation}
where $y_h(\xb)$ and $y_l(\xb,\zetab)$ denote the HF and LF sources, respectively, $y_b(\xb, \zetab)$ is the bias function that aims to quantify the systematic bias of the LF data source, $\zetab$ are the calibration parameters whose values must be inversely estimated during the fusion process (see \Cref{subsec: prob-calibration}), and $\varepsilon$ denotes normal noise whose variance may be known or not. \Cref{eq: koh} is based on {\color{black}some strong assumptions} that do not always hold in practice (e.g., existence of only one LF data source whose bias is additive). 
To dispense with such inflexible assumptions, the MF modeling capabilities of \gp~are based on converting the fusion process into a nonlinear latent variable learning problem \cite{RN1838}. 

\begin{figure}[!t]
    \includegraphics[width=1\linewidth]{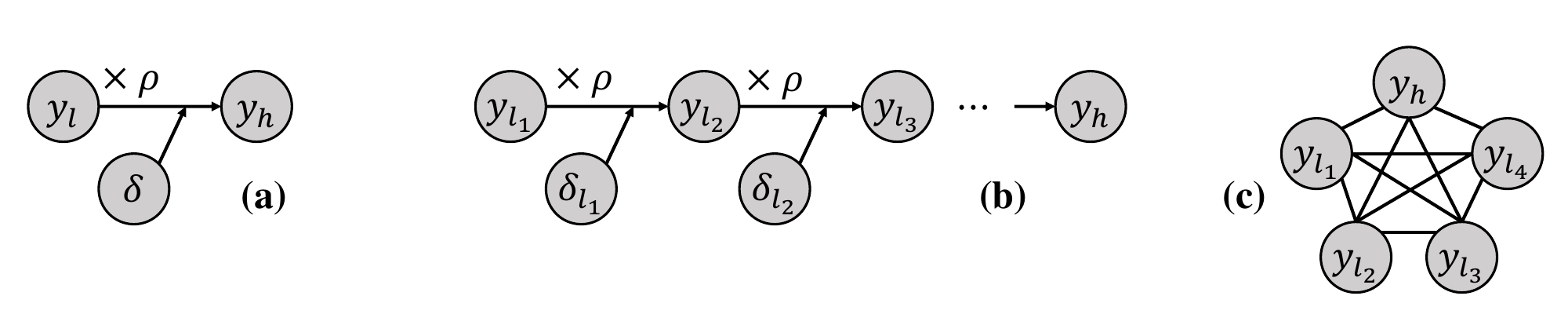}
    \caption{\textbf{Graphical representation of multi-fidelity modeling techniques:} The method of KOH \cite{RN705} \textbf{(a)} and its extension to hierarchical techniques \textbf{(b)} impose specific relations between the data sources. However, \gp~\textbf{(c)} does not impose any prior relation among the data sources and its structure resembles an undirected graph.}
    \label{fig: fusion-graph}
\end{figure}

Suppose we have $ds$ data sources of varying accuracy levels and aim to emulate all sources while dealing with $(1)$ scarce data (especially from accurate sources), $(2)$ unknown and source-dependent noise variances, and $(3)$ nontrivial biases of LF sources with respect to the HF source, i.e., we do not rely on any knowledge on the relative accuracy of the LF sources and their bias form (e.g., additive, multiplicative, etc.), see \Cref{fig: fusion-graph}. To this end \gp~first augments the input space with the additional \textit{categorical} variable $s=\{'1', \cdots, 'ds'\}$ whose $j^{th}$ element corresponds to data source $j$ for $j = 1, \cdots, ds$. Upon this augmentation, the $ds$ datasets are concatenated as:
\begin{equation} 
    \begin{split}
        \boldsymbol{U}=\left[\begin{array}{cc}
        \boldsymbol{U}_1 & '\mathbf{1}'_{n_{1} \times 1} \\
        \boldsymbol{U}_2 & '\mathbf{2}'_{n_{2} \times 1} \\
        \vdots & \vdots \\
        \boldsymbol{U}_{ds} & '\mathbf{ds}'_{n_{ds} \times 1}
        \end{array}\right] 
        \text { and }
        \boldsymbol{y}=\left[\begin{array}{c}
        \boldsymbol{y}_1 \\
        \boldsymbol{y}_2 \\
        \vdots \\
        \boldsymbol{y}_{ds}
        \end{array}\right]          
    \end{split}
    \label{eq: GP_Plus-fidelity-append}
\end{equation}
\noindent where the subscripts $1, 2, ..., ds$ correspond to the data sources, $n_j$ is the number of samples obtained from source $j$, $\boldsymbol{U}_j$ and $\boldsymbol{y}_j$ are, respectively, the $n_j \times (dx + dt)$ feature matrix and the ${n_j \times 1}$ vector of responses obtained from $s(j)$, and $'\boldsymbol{j}'$ is a categorical vector of size ${n_j \times 1}$ whose elements are all set to $'j'$. Once the unified $\{\boldsymbol{U}, \boldsymbol{y}\}$ dataset is built, \gp~fits an emulator to it following a process similar to \Cref{fig: emulation-flowchart}.
Given the importance of identifying the relative discrepancies among data sources, \gp~slightly changes the correlation functions in \Cref{eq: all-kernels-GP_Plus} to learn two embeddings where the first one encodes the categorical variables in the input space (denoted by $\boldsymbol{t}$ in \Cref{fig: emulation-flowchart}) while the second one encodes the data source identifier ($s$). Following this modeling assumption, the correlation functions in \Cref{eq: all-kernels-GP_Plus} are updated as:
\begin{subequations} 
    \begin{equation} 
        \begin{split}
            r(\ub, \ub^{\prime}; \omegab, \thetab_h, \thetab_z)= 
            \exp \left\{-\sum_{i=1}^{dx} 10^{\omega_i}(x_i-x_i^{\prime})^2\ -
            \sum_{i=1}^{dh}(h_i-h_i^{\prime})^2 -
            \sum_{i=1}^{dz}(z_i-z_i^\prime)^2\right\}
        \end{split}
        \label{eq: gaussian-kernel-GP_Plus-mf}
    \end{equation}
    \begin{equation} 
        \begin{split}
            r(\ub, \ub^{\prime}; \omegab, \thetab_h, \thetab_z)= 
            \frac{2^{1-\nu}}{\Gamma(\nu)}K_{\nu}
            \left(\sqrt{2\nu}\times \sum_{i=1}^{dx} 10^{\omega_{i}}(x_{i}-x_{i}^{\prime})^{2} + 
            \sum_{i=1}^{dh}(h_i - h^{\prime}_i)^2 + \sum_{i=1}^{dz}(z_i-z_i^\prime)^2 \right)^{1+\nu} 
        \end{split}
        \label{eq: matern-kernel-GP_Plus-mf}
    \end{equation}
    \label{eq: all-kernels-GP_Plus-mf}
\end{subequations}
\noindent where $\ub= \left[ \xb, \tb, s \right]^T$ and $\boldsymbol{z} = f_z(\boldsymbol{\pi}_s; \thetab_z)$ is the latent representation of data source $s$ and is obtained similar to $\hb$. 
Looking at \Cref{eq: all-kernels-GP_Plus-mf} we observe that the correlation between the estimated outputs of sources $s$ and $s^\prime$ at the same inputs is:
\begin{align}
    0 \leq 
    r(\begin{bmatrix}
        \boldsymbol{x}\\
        \boldsymbol{t}\\
        s
    \end{bmatrix} , \begin{bmatrix}
        \boldsymbol{x}\\
        \boldsymbol{t}\\
        s^\prime
    \end{bmatrix}) = 
    \exp \left\{0 -
    0 -
    \sum_{i=1}^{d_z}(z_i-z_i^\prime)^2\right\}
    \leq 1
\end{align}
\label{eq: GP_Plus-Corelation-extended3}
\noindent which illustrates that highly correlated data sources must have similar latent representations, (i.e., they must be encoded with close-by points in the $z-$space), see \Cref{sec: functionalities} for multiple examples. 

{\color{black}We highlight that these learned latent distances provide an average measure of correlation among the sources and cannot identify local discrepancies since the encodings in the $z-$space are not functions of $\xb$ or $\tb$}. {\color{black} Thus, if some LF sources are only locally correlated with the HF source, they will be encoded relatively far from the HF source in the learned embedding. This implies that those sources provide valuable insights in certain areas of the domain and keeping or dropping them depends on the specific applications. For instance, if the application is emulation, techniques such as cross-validation or train-test splits can assist in determining which sources to keep or drop. However if the goal is multi-fidelity BO (which typically starts with very small initial data, especially from the HF source), we recommend keeping all the sources during the optimization process (see \cite{foumani2023effects} for more details).}

MF modeling in \gp~differs significantly from most existing methods in that its structure does not prioritize learning any source (e.g., the HF source) over the others, i.e., \gp~aims to integrate all the data sets together to improve its accuracy in emulating all the sources. {\color{black}For example, multilevel best linear unbiased estimators (MBLUE) and approximate control variate (ACV) are two variance blackuction-based techniques that leverage MF data to more accurately learn the HF source \cite{MLBLUE,ACV}. Since these methods prioritize surrogating the HF source and do not build surrogates for the LF sources, they cannot be used in applications such as MFBO where one has to emulate all sources.}
In this paper, we do not explore the possibility of prioritizing emulation of a particular source (e.g., the HF source) but note that this direction can be pursued in a number of ways such as constraining the embeddings, penalizing the objective function in \Cref{eq: map-gp}, or designing specific priors.

\subsubsection{Source-Dependent Noise Modeling} \label{sec: multi_noise}
Noise inevitably arises in most applications and incorrectly modeling it blackuces the performance of any emulator. As mentioned in \Cref{sec: gp-background}, GPs model noise via the nugget or jitter parameter, $\delta$, which changes the covariance matrix from $\boldsymbol{C}$ to $\Cb_{\delta}= \Cb + \delta \Ib_{nn}$. 
Although this approach works quite well in SF problems, it does not yield the same benefits in MF emulation due to the dissimilar nature of the data sources and their corresponding noises. Consider a bi-fidelity scenario where the HF data comes from an experimental setup and is subject to measurement noise, while the LF data is generated by a deterministic computer code that has a systematic bias due to missing physics. In this case, using only one nugget parameter for MF emulation is obviously not an optimum choice. 

To address this issue effectively, we follow \cite{foumani2023effects} and use a nugget vector $\boldsymbol{\delta}=[\delta_1,\delta_2,\dots,\delta_{ds}]$ to modify the covariance matrix:
\begin{equation} 
    \begin{split}
        \boldsymbol{C}_\delta=\boldsymbol{C}+\boldsymbol{N}_\delta
    \end{split}
    \label{eq: sep_noise_corr}
\end{equation}
\noindent where $\boldsymbol{N}_\delta$ denotes an $n \times n$ diagonal matrix whose $(i,i)^{th}$ element is the nugget element corresponding to the data source of the $i^{th}$  sample. For instance, suppose the $i^{th}$  sample $\boldsymbol{u}^{(i)}$ is generated by source $ds$. Then, $(i,i)^{th}$ element of $\boldsymbol{N}_\delta$ is $\delta_{ds}$. With this modification, the estimated stationary noise variance for the $i^{th}$ data source is $\widehat{\delta}_i$. 
We highlight that \gp~uses \Cref{eq: sep_noise_corr} by default when learning from multi-source data and updates the training and inference formula accordingly. For instance, all model parameters in this case are obtained as:
\begin{equation} 
    \begin{split}
        [\widehat{\betab}, \widehat{\sigma}^2, \widehat{\thetab},\widehat{\boldsymbol{\delta}}] = 
        \underset{\betab, \sigma^2, \thetab, \boldsymbol{\delta}}{\operatorname{argmin}} \hspace{2mm} \frac{1}{2} \log (|\Cb_{\delta}|)+\frac{1}{2}(\yb-\mb)^T \Cb_{\delta}^{-1}(\yb-\mb)- \log\left(p(\betab, \sigma^2, \thetab, \boldsymbol{\delta})\right)
    \end{split}
    \label{eq: map-gp-multi-noise}
\end{equation}

{\color{black}We highlight that the above formulations model stationary noise for each data source. We make this choice due to the fact that modeling an input-dependent noise increases the number of hyperparameters by at least $ds\times dx$ which can result in overfitting \cite{RN340,RN717}. Therefore, to balance the risk of overfitting with the uncertainty quantification capacity of our emulator, we assume the noise variance is not a function of the input variables and only depends on the data source.}

\subsection{Multi-fidelity Modeling via Probabilistic Embedding} \label{subsec: prob-mfmodeling}
The MF modeling approach described in \Cref{subsec: det-mfmodeling} is deterministic in that the learnt embedding encodes a data source with a single point in the $z-$space. To more accurately quantify the epistemic uncertainties and model form errors, in this paper we develop a computationally efficient technique to learn probabilistic embeddings. While we can naturally obtain such embeddings within a Bayesian setting, we opt for a variational approach as it is computationally much more efficient. 

\begin{figure}[!h]
    \centering
        \includegraphics[width=0.6\linewidth]{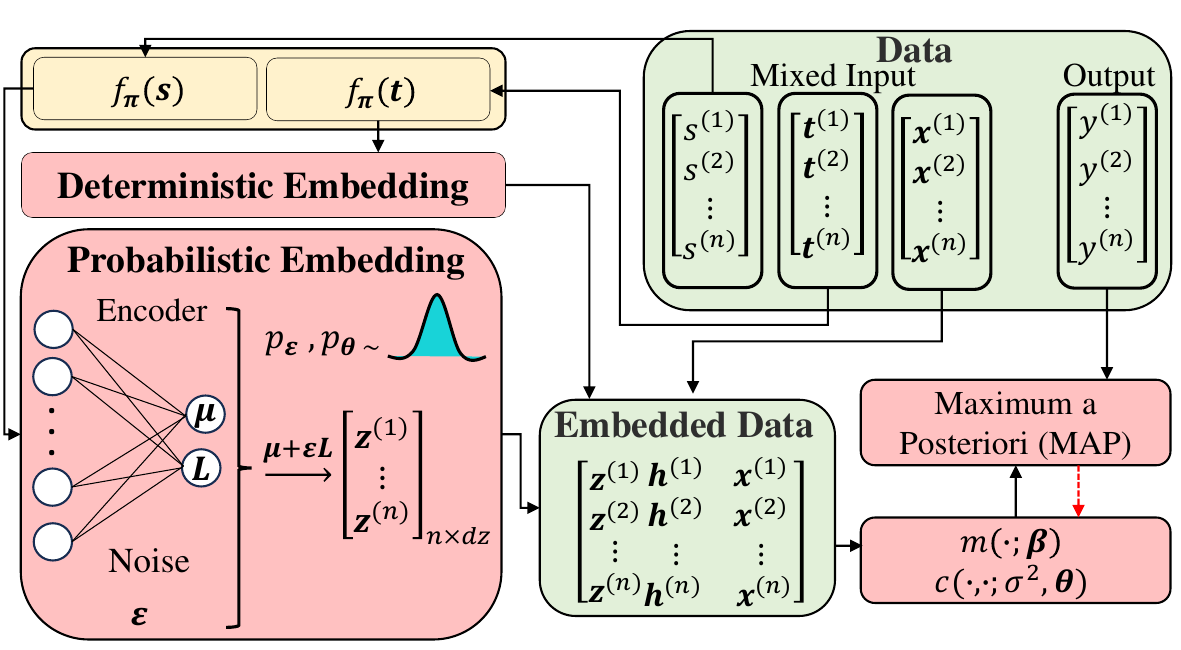}
    \vspace{-.1cm}
    \caption{\textbf{Probabilistic multi-fidelity modeling via \gp:} Categorical inputs $\tb$ are mapped to latent points in the $h-$space while the source indicator variable $s$ is mapped to a conditional distribution in $z-$space. Both mappings are achieved via deterministic and differentiable functions. Due to the probabilistic nature of $\zb$, multiple forward passes are requiblack to obtain the final outputs of the model.}
    \label{fig: mf-prob-encoder}
\end{figure}

To obtain the probabilistic latent representation of the categorical source indicator variable $s$, we reformulate $f_z(\boldsymbol{\pi}_s; \thetab_z)$ to obtain the conditional distribution $q(\zb\mid s)$.
To this end we use the reparameterization trick \cite{kingma2013auto} and design $f_z(\pib_s; \thetab_z)$ accordingly. Specifically, we model $q(\boldsymbol{z}\mid s)$ via a multi-variate normal distribution\footnote{Other distributions can also be used but we have had great success with simple ones such as the bivariate normal distribution when $dz=2$.} that is fully characterized via the mean vector $\mub_z$ and covariance matrix $\Sigmab_z=\Lb_z\Lb_z^T$ where $\Lb_z$ denotes the lower Cholesky decomposition of $\Sigmab_z$. $\zb$ is then obtained by:
\begin{equation} 
    \zb=\mub_z + \Lb_z\boldsymbol{\varepsilon}
    \label{eq: z-reparameterization}
\end{equation}
where $\boldsymbol{\varepsilon}$ is a $dz$ dimensional vector whose elements are independent standard normal variables. $\mub_z$ and $\Lb_z$ in \Cref{eq: z-reparameterization} are obtained via the differentiable and deterministic function $f_z(\pib_s; \thetab_z)$ which we choose to be a fully connected FFNN, see \Cref{fig: mf-prob-encoder}. 

We note that \gp~by default only builds a probabilistic encoding for $s$ and not $\tb$ to avoid overfitting: since the number of categorical variables and their levels is typically much larger than the number of levels of $s$ (which indicates the number of sources), probabilistically encoding $\tb$ requires an FFNN with a large number of parameters and hence may result into overfitting especially if the training data is small. An alternative approach ({\color{black}which can be achieved} in \gp~by changing its default parameters) is to encode a subset of $\tb$ in a probabilistic latent space. We leave pursuing this direction to our future studies as it is application specific.

With a probabilistic fidelity embedding, we must sample from $q(\zb\mid s)$ multiple times during both training and testing since even for fixed $s, \tb,$ and $\xb$ the pblackicted covariance from the GP model varies due to $\epsb_z$ (the effect on the mean function depends on its formulation, see \Cref{sec: gp_mixed_basis_functions}). Hence, for any fixed values of $s, \tb, \xb,$ and model parameters (i.e., $\betab, \thetab_h, \sigma^2,$ and $\thetab_z$), we generate $M$ samples from $q(\zb\mid s)$\footnote{These samples are generated by drawing $M$ random $\epsb_z$ vectors of size $dz$ from a standard multi-variate normal distribution.} to build an ensemble or mixture of $M$ GPs. Since $\epsb_z$ are independent and identically distributed (IID), each member of the GP ensemble is equally probable and we can extend \Cref{eq: gp-mean-var-vec} via the laws of total expectation and covariance \Cite{wolpert2010conditional,rudary2009pblackictive}. 
To this end, we must obtain expressions for $\mb$ and $\Cb$ in \Cref{eq: map-gp} during training. We distinguish among the GPs in the mixture model via the random variable $I$ (whose probability mass function or PMF is $p(I=k) = \frac{1}{M}$ for $k=1, \cdots, M$) and we calculate the ensemble mean as:
\begin{equation}
    \bar{m}(\ub) = \E[m_k(\ub)] = \frac{1}{M} \sum_{k=1}^{M} m_k(\ub)
    \label{eq: ens-mean}
\end{equation}
where $m_k(\ub)$ is the expected value of the $k^{th}$ GP in the ensemble and correspondingly we denote $\bar{\mb} = \frac{1}{M} \sum_{k=1}^{M} \mb_k$ as the ensemble mean over the training data. 

To obtain the ensemble expression for $\Cb$, we start by writing the covariance between the two random variables $y(\ub)$ and $y(\ub^{\prime})$ as:
\begin{equation}
    \text{cov}\left(y(\ub), y(\ub')\right) = 
    \E\left[\text{cov}\left(y(\ub), y(\ub')\right)|I\right] + \text{cov}\left(\E[y(\ub)|I], \E[y(\ub')|I]\right)
    \label{eq: cov-law-of-total}
\end{equation}
Given the PMF of $I$ and \Cref{eq: gp-cov}, we can calculate the first term on the right hand side of \Cref{eq: cov-law-of-total} as:
\begin{equation}
    \E\left[\text{cov}\left(y(\ub), y(\ub')\right) \right |I ] = 
    \frac{1}{M} \sum_{k=1}^{M} c_k(\ub, \ub'; \sigma^2, \thetab) = 
    \bar{c}(\ub, \ub'; \sigma^2, \thetab)
    \label{eq: ens-cov1}
\end{equation}
where the subscript $k$ only affects the $z-$components in the kernel. For instance, $c_k(\ub, \ub'; \sigma^2, \omegab, \thetab_{h}, \thetab_{z}) = \sigma^2 \exp \left\{-\sum_{i=1}^{dx} 10^{\omega_i}(x_i-x_i')^2\ - ||\hb-\hb'||^2_2 - ||\zb_k-\zb_k'||^2_2\right\}$ for a Gaussian kernel. 

We now turn to the second term on the right hand side of \Cref{eq: cov-law-of-total} and represent it as:
\begin{equation}
    \text{cov}\left(\E[y(\ub)|I], \E[y(\ub')|I]\right)= 
    \frac{1}{M} \sum_{k=1}^{M} \left(m_k(\ub) - \bar{m}(\ub) \right) \left(m_k(\ub') - \bar{m}(\ub)\right).
    \label{eq: ens-cov2}
\end{equation}
Inserting \Cref{eq: ens-cov1,eq: ens-cov2} into \Cref{eq: cov-law-of-total} we obtain:
\begin{equation}
    \text{cov}\left(y(\ub), y(\ub')\right) = 
    \bar{c}(\ub, \ub'; \sigma^2, \thetab)+ 
    \frac{1}{M} \sum_{k=1}^{M} \left(m_k(\ub) - \bar{m}(\ub) \right) \left(m_k(\ub') - \bar{m}(\ub)\right)
    \label{eq: ens-cov}
\end{equation}
which allows us to calculate the ensemble $\Cb$ for \Cref{eq: map-gp} as:
\begin{equation}
    \bar{\Cb} = \frac{1}{M} \sum_{k=1}^{M} \Cb_k + (\mb_k - \bar{\mb})(\mb_k - \bar{\mb})^T
    \label{eq: ens-cor}
\end{equation}
where $\boldsymbol{C}_k$ denotes the covariance matrix of the $k^{th}$ ensemble member whose $(i, j)^{t h}$ element is given by $ c_k(\ub^{(i)}, \ub^{(j)}; \sigma^2, \thetab)$ for $i, j=1, \ldots, n$. We use $\bar{\Cb}$ and $\bar{\mb}$ while solving the optimization problem in \Cref{eq: map-gp}.

For pblackiction, we take $Q$ samples from the probabilistic fidelity embedding to determine the ensemble mean and variance:
\begin{subequations} 
    \begin{equation} 
        \E[y(\ub^*)] =\bar{\mu}(\ub^*)= \frac{1}{Q} \sum_{k=1}^{Q} \mu_k(\ub^*)
        \label{eq: gp-ens-mean-scalar}
    \end{equation}
    \begin{equation} 
        \text{cov}(y(\ub^*), y(\ub^*))=\bar{\tau}^2(\ub^*) =
        \frac{1}{Q}\sum_{k=1}^{Q} \parens{ \tau_{k}^2(\ub^*) + \mu_k^2(\ub^*)} - \bar{\mu}^2(\ub^*)
        \label{eq: gp-ens-var-scalar}
    \end{equation}
    \label{eq: gp-ens-mean-and-var} 
\end{subequations}
where $\mu_k(\ub^*)$ and $\tau_{k}^2(\ub^*)$ are the mean and variance of the $k^{th}$ GP and obtained via \Cref{eq: gp-mean-var-scalar}. We note that the ensemble variance formula is a specific instance of the ensemble covariance given in \Cref{eq: ens-cov}. Additionally, we typically use $Q>M$ to blackuce the training costs.

\subsection{Gaussian Processes with Mixed Basis Functions} \label{sec: gp_mixed_basis_functions}
The parametric mean and covariance functions in \Cref{eq: gp-mean and cov} can be formulated in many ways. In this regard, most advancements have focused on designing the kernel (e.g., the ones we develop in \Cref{eq: all-kernels-GP_Plus,eq: ens-cov} or deep kernels \cite{lee2017deep,al2017learning}) since it significantly affects the performance of the resulting GP model. However, the mean function in \Cref{eq: gp-mean} plays an important role in many applications that involve, e.g., extrapolation \cite{planas2021evolutionary}, fusing multi-source data, or identifying model form errors. 

\begin{figure}[!h] 
    \centering
    \includegraphics[width=1\linewidth]{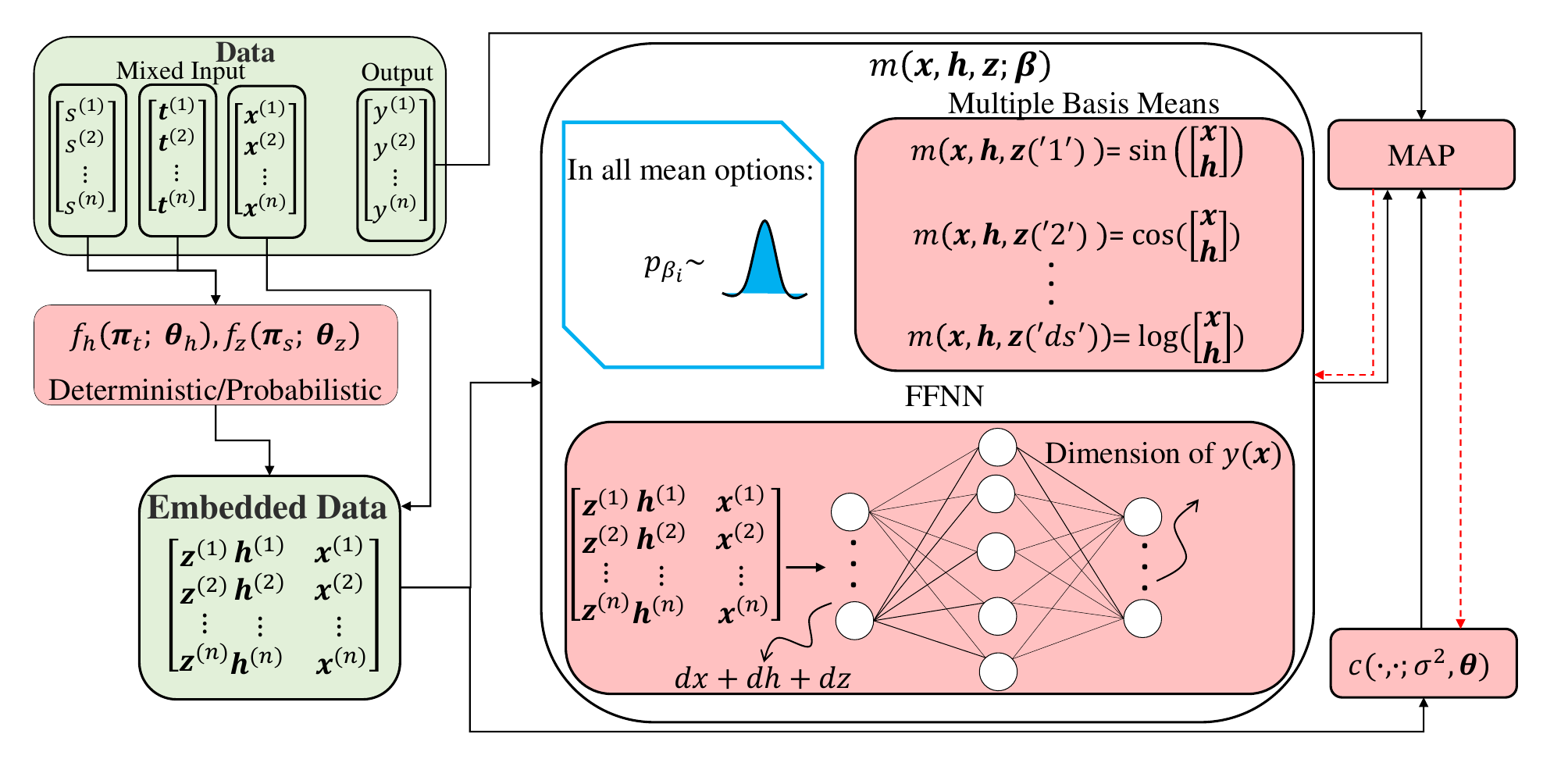}
    \vspace{-.7cm}
    \caption{\textbf{Multi-fidelity modeling via mixed basis functions:} Two generic options are defined in \gp~for building the mixed bases: $(1)$ pblacketermined bases where multiple bases like polynomial, $sin(\cdot)$ and $cos(\cdot)$ can be defined for each data source,
    $(2)$ FFNNs with user-defined architectures. All the parameters of the mean functions have a normal prior and are jointly learned through MAP.}
    \label{fig: Mixed_Basis_Functions}
\end{figure}

Existing techniques typically leverage polynomials (in combination with other analytic functions such as $\sin(\cdot), \log(\cdot), \cdots$) or NNs in designing $m(\xb; \betab)$. In \gp, we extend these methods to seamlessly include the categorical variables in the mean function. Specifically, our idea is to feed the learnt representations of $\tb$ and $s$ into the mean function instead of the original categorical variables, i.e., we reparameterize $m(\xb, \tb, s; \betab)$ to $m(\xb, \hb, \zb; \betab)$, see \Cref{fig: Mixed_Basis_Functions}. A major difference between our reparameterization and other alternatives (such as an NN whose inputs are one-hot encoded representation of $\tb$ and $s$) is that our mean and covariance functions are directly coupled since the latent variables used in $m(\xb, \hb, \zb; \betab)$ are parameterized in the kernel, i.e., $\hb=f_h(\pib_t; \thetab_h)$ and $\zb=f_z(\pib_s; \thetab_z)$. {\color{black} Based on this idea, in \gp~we provide the following two options for modeling the mean function: (1) Having a global function that is shablack among \textit{all} combinations of $\tb$ and $s$, and (2) Having mixed basis functions where a unique mean function is learnt for specific combinations of the categorical variables (e.g., in MF modeling, we can learn a unique mean function for each of the $s$ data sources)}.

 {\color{black}\Cref{fig: Mixed_Basis_Functions} illustrates two generic options that we define in \gp~for building mixed bases.
The first option builds the mean function based on pre-determined bases that can include polynomials, $sin(\cdot), log(\cdot)$ or any other analytic functions. The second option is based on a fully connected FFNN whose architecture (e.g., number of hidden layers and their sizes) should be designed by the user. The size of the input layer of the NN depends on the dimensionality of $\xb, \hb,$ and $\zb$ while its output layer size depends on the response dimensionality (hence the output size is $1$ for a single-response dataset).}

Mixed bases are useful in applications where the input space has categorical features. The MF modeling approach described in \Cref{subsec: det-mfmodeling,subsec: prob-mfmodeling} is one such application as it requires adding the categorical variable $s$ to the original input space. As shown in \Cref{sec: functionalities}, using mixed bases improves MF modeling by allowing the fused GP model to emulate data source $i$ with a unique mean function that better captures the global and local features of source $i$. 
To visualize this benefit, we consider the simple \sine~example described in \Cref{sec: analytic_example} where $4$ and $20$ noisy data points from the HF and LF sources, respectively, are provided and the goal is to emulate both the HF and LF sources while inversely learning the model form error of the LF source. To investigate the effects of mixed bases, we build two GPs where the first one learns a single constant mean function for the fused data while the second one considers different mean functions for the two sources, namely, a zero mean for the HF source and a second-degree polynomial for the LF source. 

\begin{figure}[!t]
  \centering
    \begin{subfigure}{1\textwidth}
        \includegraphics[width=\textwidth]{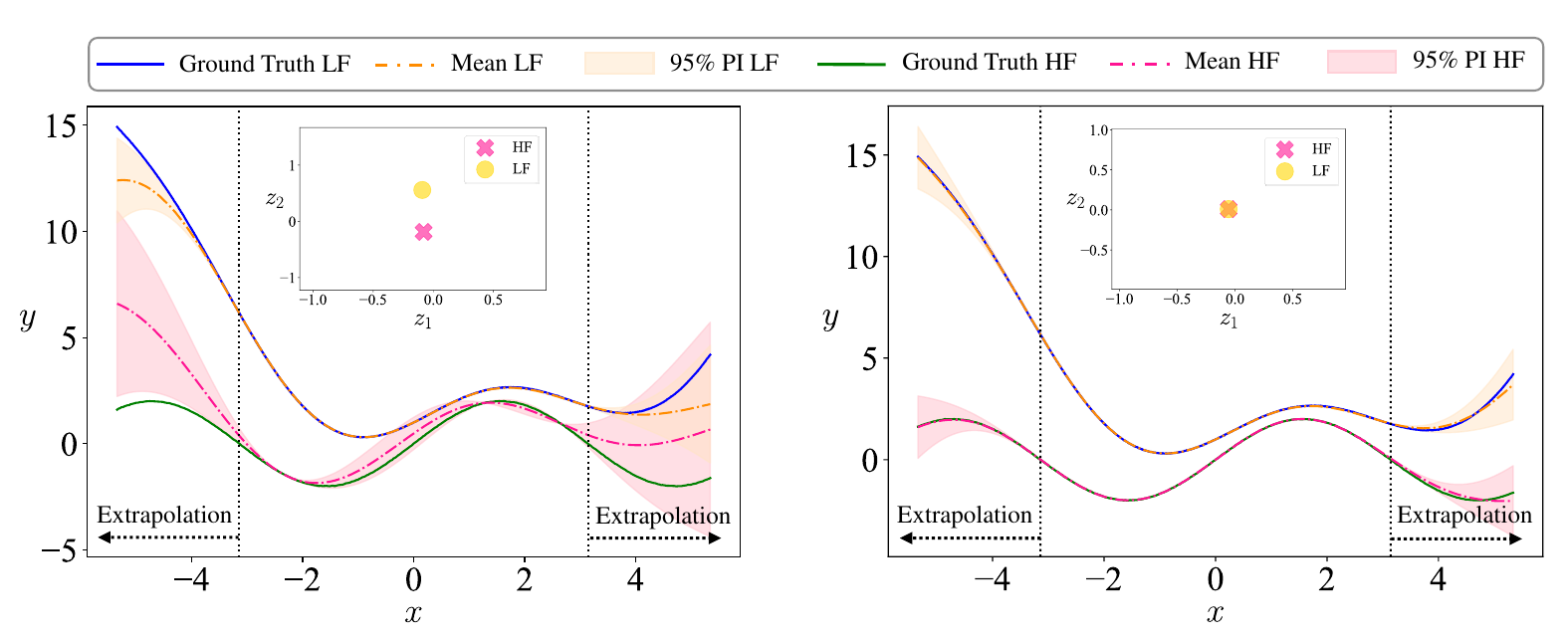}
        \vspace{-7mm}
        \vspace{-2mm}
        \end{subfigure}
        \begin{subfigure}{0.49\textwidth}
        \caption{\textbf{Single base}}
        \label{fig: Single_base}
    \end{subfigure}
    \begin{subfigure}{0.49\textwidth}
        \caption{\textbf{Mixed bases}}
        \label{fig: Mixed_bases}
    \end{subfigure}
    \vspace{-2mm}
    \caption{\textbf{Effect of mixed bases in MF emulation:} The comparison is conducted on the $1$-dimensional bi-fidelity problem where the LF source has a polynomial bias relative to the HF source, see \sine~example in \Cref{table: analytic-formulation}. Mixed bases \textbf{(b)} enables the emulator to better capture the local and global biases which, in turn, increases its interpolation and extrapolation powers. The learnt fidelity manifolds are shown as insets in each figure.}
    \label{fig: multi_basis_1d}
\end{figure}

The results are illustrated in \Cref{fig: multi_basis_1d} and indicate that the second GP emulates both sources better than the first GP in both interpolation and extrapolation. As it can be seen in \Cref{table: analytic-formulation} the true model form error is $0.3x^2-0.7x+1$ while the discoveblack one with the second GP is $0.2981x^2-0.7059x+0.9939$. We attribute the small differences between these two functions primarily to the fact that the training data is very small and noisy. As also shown in \Cref{fig: multi_basis_1d} we observe that the inclusion of the mixed bases affects the learnt encoding for $s$ whose two levels are mapped to distant latent points in the first GP but close-by points in the second GP. This behavior indicates that the entire model form error in the second GP can be obtained by comparing the mean functions associated with the two sources. 

In the above example, the true model form error is a $2^{nd}$ degree polynomial and so we choose polynomial bases (of degree zero and two for the HF and LF sources, respectively) as the mean functions for the second GP. In practice, however, identification of the true model form error in realistic applications is much more challenging due to its unknown form, high dimensionality of the problem, lack of data, or noise. In these scenarios, we recommend using mean functions such as FFNNs that can adapt to the data and better model the global and local trends. 

\subsection{Inverse Parameter Learning for Computer Models} \label{subsec: prob-calibration}
Most computer models are built to be applicable to a broad range of applications. Using these models in a specific context typically relies on adjusting their context-specific parameters using our domain knowledge or some measurements/observations. For instance, finite element (FE) models can be used to simulate the behavior of many materials under a wide range of loading conditions \cite{RN311}. 
However, for a specific application such as modeling the thermoforming process of woven carbon fiber reinforced plastics \cite{10.1063/1.4963592, BOTELHO20031843, RN791}, a specific FE model is needed. The material parameters such as directional Young's moduli and yarn-yarn friction coefficients in this FE model should be \textit{calibrated} such that it can reproduce force-displacement curves obtained via experiments such as tension or three-point bending tests. Since such experimental data does not include the material parameters themselves, one must solve an inverse problem where the FE model's calibration parameters are estimated such that the model fits the experimental data. During this process, it is implicitly presumed that the response (e.g., force-displacement curve) is sufficiently sensitive to the calibration parameters (e.g., Young's moduli and yarn-yarn friction coefficients) as otherwise they cannot be accurately estimated.

Calibration of a computer model is tightly connected to that model's bias with respect to an HF data source (e.g., experiments). To explain this connection, we note that most computer models suffer from systematic errors that arise from, e.g., their missing physics, the simplifying assumptions made during their development, or numerical errors. To mitigate the effect of these errors, computer models sometimes include a few additional calibration or tuning parameters that may not even correspond to any physical properties of the system.
One example is the artificial viscosity parameter that is used to stabilize explicit solution methods that are needed when modeling dynamic processes such as fracture via the FE method.
A related example is the calibration of physics-based blackuced-order models (ROMs) \cite{RN1182, RN1655, RN1854, RN1786, dvorak1992transformation, roussette2009nonuniform} that simplify expensive computer models (such as direct numerical simulations or DNS) to gain computational speedups. Such simplifications introduce some bias into the ROMs whose effects are typically mitigated by calibrating material parameters such that a ROM can reproduce small HF data obtained from DNS. That is, even if the material parameters are known, one may have to adjust them for ROMs.

\begin{figure}[!b]
    \centering
    \includegraphics[width=1\linewidth]{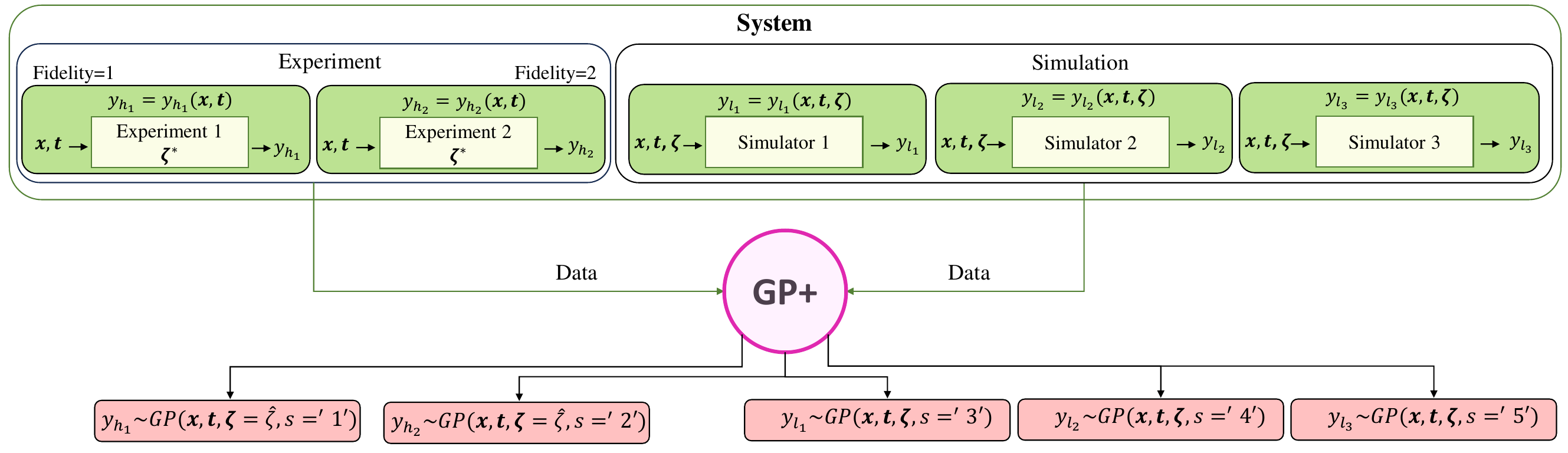}
    \caption{\textbf{Inverse parameter calibration via data fusion:} \gp~builds a fused model that jointly emulates all the sources while inversely estimating the calibration parameters of the HF sources. It is assumed that {\color{black}$(1)$ all the data sources share the same inputs, $(2)$ the HF data sets correspond to the same underlying system, and $(3)$ the calibration parameters represent some properties of the HF sources.}}
    \label{fig: calibration-flowchart}
\end{figure}

Calibration of computer models is closely related to MF modeling since it requires fusing multiple datasets that typically have different levels of fidelity (e.g., fusing simulations with experiments or observations). 
Hence, we extend the capabilities introduced in \Cref{subsec: det-mfmodeling,subsec: prob-mfmodeling} to accommodate inverse parameter learning for computer models. For this extension, we consider two application scenarios:
\begin{itemize}
    \item Simultaneous calibration of multiple ($>1$) computer models: We presume that the calibration parameters of these models correspond to some unobserved characteristics of a system. We make this assumption since it is not optimal to jointly calibrate the \textit{tuning} parameters of different models that are added to them for reasons besides characterizing unobserved features of a system (note also that the number of these tuning parameters generally varies across different models that simulate the same system). 
    \item Calibration of a single computer model: We do not distinguish between the calibration parameters regardless of whether they are merely tuning knobs or they correspond to some unobservable features. 
\end{itemize}
We highlight that in both scenarios we can use multiple HF data sets in \gp~as long as they correspond to the same physical system (e.g., obtaining force-displacement curves via different universal testing machines that have different levels of fidelity), see \Cref{fig: calibration-flowchart}. 

Following the notation of previous sections, we denote the quantitative inputs by $\xb$ and the latent representations of qualitative inputs and the categorical source indicator variable by $\hb$ and $\zb$, respectively. These inputs are shablack across all the data sources but, as described above, the LF sources have additional quantitative inputs that correspond to the calibration parameters and are denoted by $\zetab = [\zeta_1, \zeta_2, \ldots, \zeta_{d_\zeta}]^T$. The “best” calibration parameters ($\zetab^*$) should be estimated using the HF data to {\color{black} accurately characterize the physical system}.
We denote these estimates by $\widehat{\zetab}$ and modify the correlation function to obtain them. For instance, the Gaussian and Mat\'ern correlation functions are reformulated as follows:
\begin{subequations} 
    \begin{equation} 
        \begin{split}
        \begin{aligned}
            r(\ub, \ub^{\prime}; \omegab, \thetab_h, \thetab_z)= 
            \exp \Bigg\{-\sum_{i=1}^{dx} 10^{\omega_i}(x_i-x_i^{\prime})^2\ -
            \sum_{i=1}^{dh}(h_i-h_i^{\prime})^2 -\\
            \sum_{i=1}^{dz}(z_i-z_i^\prime)^2
            & - \sum_{i=1}^{d\zeta} 10^{\omega_{i+dx}}(\zeta_{i}-\zeta_{i}^{\prime})^2 \Bigg \}
            \end{aligned}
        \end{split}
        \label{eq: gaussian-kernel-GP_Plus-cal}
    \end{equation}
    \begin{equation} 
        \begin{split}
            r(\ub, \ub^{\prime}; \omegab, \thetab_h, \thetab_z)= 
            \frac{2^{1-\nu}}{\Gamma(\nu)}K_{\nu}
            \Bigg(\sqrt{2\nu}\times \sum_{i=1}^{dx} 10^{\omega_{i}}(x_{i}-x_{i}^{\prime})^{2} + 
            \sum_{i=1}^{dh}(h_i - h^{\prime}_i)^2 +\\ \sum_{i=1}^{dz}(z_i-z_i^\prime)^2 + 
            \sum_{i=1}^{d\zeta} 10^{\omega_{i+dx}}(\zeta_{i}-\zeta_{i}^{\prime})^2
            \Bigg)^{1+\nu} 
        \end{split}
        \label{eq: matern-kernel-GP_Plus-cal}
    \end{equation}
    \label{eq: all-kernels-GP_Plus-cal}
\end{subequations}
where $\ub = \brackets{\xb, \tb, s, \zetab}^T$ and $\omegab$, $\thetab_h$ and $\thetab_z$ are defined as before. 
While training the model, the correlation between LF samples can be readily calculated via \Cref{eq: all-kernels-GP_Plus-cal}. However, if at least one of the samples is an HF one, in the last term of \Cref{eq: gaussian-kernel-GP_Plus-cal,eq: matern-kernel-GP_Plus-cal} we use $\widehat{\zeta_i}$ which are estimated jointly with all the other parameters of the model via MAP.

Similar to \Cref{subsec: prob-mfmodeling}, we can inversely learn the calibration parameters within a probabilistic setting to more accurately quantify the uncertainties compablack to the deterministic counterpart based on the MAP. 
Following \Cref{subsec: prob-mfmodeling}, we learn the calibration parameters within a variational framework by formulating their joint posterior with a multivariate normal distribution that is fully characterized by its mean vector and covariance matrix. 
We learn the parameters of this joint distribution via the reparametrization trick:
\begin{equation}
    \zeta_i=\mu_{\zetai} +\tau_{\zetai} \boldsymbol{\varepsilon}_{\zetai},
    \label{eq: EnsMean-cal}
\end{equation}
where $\boldsymbol{\varepsilon}_{\zetai} \sim \mathcal{N}(0,1)$ is an auxiliary noise variable while $\mu_{\zetai}$ and $\tau_{\zetai}$ parameterize the posterior distribution of $\zetai$. 
During both the training and pblackiction phases, we draw samples for calibration parameters. As a result, the values of $\zetab$ fluctuate in each optimization iteration which consequently changes the covariance and mean functions of the emulator. To efficiently consider these variations, we follow our method used for probabilistic manifold modeling (see \Cref{subsec: prob-mfmodeling}) and employ ensembling to calculate both the mean vector and covariance matrix. 

{\color{black} We highlight that in both deterministic and probabilistic calibration cases, the estimated calibration parameters and the learnt bias are tightly connected in that the former depends on what bias form has been chosen. Most existing methods \cite{RN377, RN375, RN271, RN272, RN335, RN336, RN645, RN607, RN295, RN705} first assume a specific functional form (e.g., a GP or a polynomial) for the bias and  the relation between the LF and HF sources (see \Cref{eq: koh} and \Cref{fig: fusion-graph} for one example). Then, given data from both LF and HF sources, they estimate the calibration parameters and the parameters of the bias function. For these approaches, the estimated calibration parameters are strongly dependent on the \textit{assumed} form of the bias function and how it relates the HF and LF sources. If these assumptions are incorrect, the calibration results will be misleading. 
In \gp~we significantly relax these assumptions. Specifically, we $(1)$ do not assume the bias term is additive, and $(2)$ simultaneously calibrate multiple sources (rather than just calibrating one source at a time). Therefore, we do not eliminate the so-called identifiability issue but provide the means that analysts can use to address it depending on the application. For instance, an effective way to blackuce identifiability issues is using multiple-response data during calibration \cite{RN607, RN272, RN271, RN277}. Similarly, the multi-source calibration mechanism in \gp~provides the calibration process with more information and hence has the potential to blackuce non-identifiability. Additionally, in \gp~we can use mixed basis functions which can help analysts in choosing appropriate mean functions for each source and study the effects of this choice on the estimated calibration parameters, accuracy on unseen data, and learnt bias functions (note that the difference between two mean functions essentially gives the global bias between the corresponding sources, see \Cref{fig: multi_basis_1d}).}
    \section{Functionalities of \gp~and Comparative Studies} \label{sec: functionalities}

In this section, we demonstrate the core functionalities of \gp~and compare them against some of the widely used methods or open-source GP modeling packages. 
We start with emulation and MF modeling in \Cref{sec: emulation_functionality,sec: mfmodeling_functionality}, respectively, where we also study the potential benefits of using a probabilistic embedding instead of a deterministic one in \gp~in \Cref{sec: Pro_vs_Det_compare_functionality}. 
Then, in \Cref{sec: Calibration_Functionality} we conduct a few carefully designed studies to evaluate the capabilities of \gp~in inverse parameter estimation. Finally, in \Cref{sec: BO} we assess the performance of \gp~in BO which is a many-query outer-loop application where GPs are dominantly used for emulation. 

Throughout this section, we use normalized root mean squablack error (NRMSE) and normalized interval score (NIS) for assessing the accuracy of, respectively, the mean values and pblackiction intervals provided by a GP:

\begin{equation} 
    \begin{split}
        NRMSE=\frac{1}{std(\boldsymbol{y})}\sqrt{\frac{1}{n_{\text{test}}} \sum_{i=1}^{n_{\text{test}}}({y}^{(i)}-{\mu}^{(i)})^2}
    \end{split}
    \label{eq: nrmse}
\end{equation}
\begin{equation}
\begin{aligned}
    NIS= \frac{1}{std(\boldsymbol{y})} (& \frac{1}{n_{\text{test}}} \sum_{i=1}^{n_{\text{test}}}(\mathcal{U}{}^{(i)} - \mathcal{L}{}^{(i)}) + \frac{2}{v}(\mathcal{L}{}^{(i)} - {y}^{(i)}) \mathbbm{1}\{{y}^{(i)} < \mathcal{L}{}^{(i)}\} \\
           & + \frac{2}{v}({y}^{(i)} - \mathcal{U}{}^{(i)}) \mathbbm{1}\{{y}^{(i)} > \mathcal{U}{}^{(i)}\})
\end{aligned}
\label{eq: nis}
\end{equation}
\noindent where ${y}^{(i)}=y({\boldsymbol{u}}^{(i)})$ and ${\mu}^{(i)}=\mu{({\boldsymbol{u}}^{(i)})}$ denote the output and pblackicted mean of test sample ${\boldsymbol{u}}^{(i)}$, respectively, and $std(\boldsymbol{y})=std(y(\boldsymbol{U}))$ shows the standard deviation of the test samples obtained at $\boldsymbol{U}$. $\mathcal{U}^{(i)}$ and $\mathcal{L}^{(i)}$ are upper and lower endpoints of the pblackiction interval for the $i^{th}$ test sample. These endpoints are the pblackictive quantiles at levels $v/2$ and $1-v/2$, respectively. We use $95\%$ pblackiction interval ($v=0.05$) and hence these endpoints are defined as $\mathcal{U}^{(i)}= {\mu}^{(i)}+1.96 {\tau}^{(i)}$ and $\mathcal{L}^{(i)}= {\mu}^{(i)}-1.96{\tau}^{(i)}$. $\mathbbm{1}\{\cdot\}$ is an indicator function which is $1$ if its condition holds and zero otherwise. 
For both metrics in \Cref{eq: nrmse,eq: nis} lower values indicate more accuracy. 

{\color{black}Unless otherwise stated, we use the default settings of \gp~which are detailed in \Cref{table: model_options,table: optim_options}. For instance, the prior distributions of the parameters in all the examples are $\omega_i\sim N(-3,3)$, $\beta \sim N(0,1)$, $A_{ij} \sim N(0,1)$, $\sigma^2 \sim LN(0,1)$\footnote{Log-Normal}, and $\delta_i \sim LHS(0,0.01)$\footnote{Log-Half-Horseshoe with zero lower bound and scale parameter $0.01$.}  \cite{carvalho2010horseshoe}.}

\subsection{Emulation} \label{sec: emulation_functionality}
We use three analytic and two engineering problems to compare the performance of \gp~against the GP emulation capabilities of GPytorch\footnote{GPyTorch uses the Adam optimizer by default and to improve its performance we use a learning rate scheduler during the training process. To avoid convergence to local optima, we repeat the optimization process $64$ times, each with a different initialization for the model's parameters.}, MATLAB\footnote{We use automatic relevance determination (ARD) squablack exponential kernel for GP emulation in MATLAB.}, BoTorch, and SMT2 \cite{Matlab,BoTorch,RN1270,saves2024smt}. These problems are briefly described below (see details in \Cref{sec: appendix-Data}) and they cover a range of input dimensionality and characteristics (e.g., \hoip~only has categorical inputs). {\color{black}To provide a comprehensive analysis, we illustrate the emulation comparisons in this section and discuss the computational costs of these baselines in \Cref{sec: Computational Costs of Emulation}.}

\begin{figure}[!t]
    \centering
        \includegraphics[width=1\linewidth]{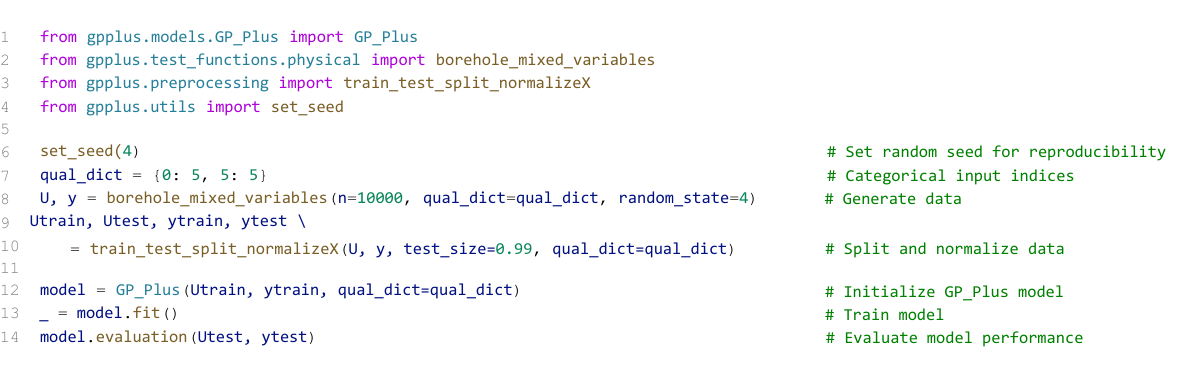}
    \vspace{-.7cm}
    \caption{\textbf{Emulation via \gp:} We emulate the \boreholeM~ function by importing the necessary modules, identifying the indices of the categorical variables, generating data, initializing the \gp~model, and finally training and evaluating the model.}
    \label{fig: code_screenshot_boreholem}
\end{figure}

As shown in \Cref{fig: code_screenshot_boreholem}, emulation via \gp~is achieved via a few lines of code regardless of whether the problem has categorical variables or not. As opposed to \gp, MATLAB handles categorical variables by first one-hot encoding them and then treating the resultant variables as numerical. 
BoTorch leverages mixed single-task GP (MST-GP) which defines two distinct kernels for numerical and categorical features. Specifically, MST-GP uses the Mat\`ern kernel for the numerical features while for the categorical features it calculates the exponential of their normalized binary distance which is $0$ when the two categorical variables are the same. The final kernel of MST-GP is the combination of the categorical and numerical parts (see \Cref{sec: MST_GP} for details). 
Since MST-GP is specifically developed to handle mixed input spaces that have both categorical and numerical features, we use GPytorch in problems with only quantitative features (note that Gpytorch cannot handle categorical inputs). SMT2 uses a Gaussian Kernel for problems involving solely numerical inputs which is multiplied by a categorical kernel in case the input space has qualitative features. The options for categorical kernel provided in SMT2 are homoscedastic hypersphere ($\text{SMT2}_{\text{HH}}$) \cite{zhou2011simple}, exponential homoscedastic hypersphere ($\text{SMT2}_{\text{EHH}}$) \cite{saves2023mixed}, and Gower distance-based correlation kernels ($\text{SMT2}_{\text{Gower}}$) \cite{halstrup2016black}. 

\wing~and \borehole~are two single-response analytic examples whose input space only has quantitative features. The dimensionality of the input space for \wing~and \borehole~is $10$ and $8$, respectively. To compare the performance of the three methods in mixed input spaces, we convert the first and sixth features of \borehole~to categorical variables with $5$ distinct levels. This analytic example is referblack to as \boreholeM~(see \Cref{sec: analytic_example} for further details). We generate $10000$ samples from the HF source in \Cref{table: analytic-formulation} and use $1\%$ of the data for training and the rest for testing. \hoip~and \nta~are both $3$-dimensional problems and only have categorical inputs (see \Cref{sec: engineering_example} for more details). We use $150$ and $100$ HF samples from \hoip~and \nta, respectively, for emulation and the rest of the HF data for testing. 
For all problems, we repeat the emulation process $10$ times and report the average values for each metric to ensure the metrics are robust to random initialization.


Emulation results are summarized in \Cref{table: results_emulation} which demonstrate that \gp~more accurately pblackicts the responses in all examples and provides more reliable pblackiction intervals. In the case of \wing~or \borehole~which only have quantitative features, we attribute the superiority of \gp~primarily to the parameter optimization process. Specifically, we use MAP (as opposed to maximum likelihood estimation or MAE), search for the length-scale parameters (i.e., $\omegab$ in the correlation function) in the log scale, and leverage L-BFGS-B\footnote{Limited-memory Broyden–Fletcher–Goldfarb–Shanno that considers simple bounds on the variables.}. for optimization. These choices smooth the profile of the objective function and accelerate the convergence.
{\color{black}The comparable performance of SMT2 and \gp~in terms of NRMSE is attributed to SMT2's usage of the so-called profiling technique \cite{RN783} for parameter estimation.} 

\begin{table}[!t]
    \centering
    \small
    \setlength{\tabcolsep}{3pt} 
    \begin{tabular}{c|cc|cc|cc|cc|cc} 
        \cline{2-11} 
        &\multicolumn{2}{>{\columncolor{headcolor}}c|}{\wing} & \multicolumn{2}{>{\columncolor{headcolor}}c|}{\borehole} & \multicolumn{2}{>{\columncolor{headcolor}}c|}{\boreholeM} & \multicolumn{2}{>{\columncolor{headcolor}}c|}{\hoip} & \multicolumn{2}{>{\columncolor{headcolor}}c}{\nta} \\ \hline
        Model & NRMSE & NIS & NRMSE & NIS & NRMSE & NIS & NRMSE & NIS & NRMSE & NIS \\ \hline
        \gp   & $\cellcolor{columncolor}0.0010$ & $\cellcolor{columncolor}\textbf{0.0049}$ & $\cellcolor{columncolor}\textbf{0.0008}$ & $\cellcolor{columncolor}\textbf{0.0045}$ & $\cellcolor{columncolor}\textbf{0.0023}$ & $\cellcolor{columncolor}\textbf{0.0139}$ & $\cellcolor{columncolor}\textbf{0.490}$ & $\cellcolor{columncolor}\textbf{2.9563}$ & $\cellcolor{columncolor}\textbf{0.2950}$ & $\cellcolor{columncolor}\textbf{2.2128}$ \\
        MATLAB & $\cellcolor{columncolor}0.0045$ & $\cellcolor{columncolor}0.0447$ & $\cellcolor{columncolor}0.0033$ & $\cellcolor{columncolor}0.0430$ & $\cellcolor{columncolor}0.0042$ & $\cellcolor{columncolor}0.0464$ & $\cellcolor{columncolor}0.5404$ & $\cellcolor{columncolor}4.198$ & $\cellcolor{columncolor}0.4043$ & $\cellcolor{columncolor}2.6360$ \\
        MST-GP  & $\cellcolor{columncolor}-$ & $\cellcolor{columncolor}-$ & $\cellcolor{columncolor}-$ & $\cellcolor{columncolor}-$ & $\cellcolor{columncolor}0.0062$ & $\cellcolor{columncolor}0.0338$ & $\cellcolor{columncolor}0.515$ & $\cellcolor{columncolor}3.940$ & $\cellcolor{columncolor}1.38$ & $\cellcolor{columncolor}2.499$ \\
        GPytorch & $\cellcolor{columncolor}0.0081$ & $\cellcolor{columncolor}0.0715$ & $\cellcolor{columncolor}0.0078$ & $\cellcolor{columncolor}0.0641$ & $\cellcolor{columncolor}-$ & $\cellcolor{columncolor}-$ & $\cellcolor{columncolor}-$ & $\cellcolor{columncolor}-$ & $\cellcolor{columncolor}-$ & $\cellcolor{columncolor}-$ \\ 
        \text{SMT2} & $\textbf{0.0009}$ & $0.0052$ & $\textbf{0.0008}$ & $0.0054$ & $-$ & $-$ & $-$ & $-$ & $-$ & $-$ \\
        $\text{SMT2}_{\text{Gower}}$ & $-$ & $-$ & $-$ & $-$ & $0.0034$ & $0.0331$ & $0.526$ & $3.5468$ & $0.3939$ & $2.5104$ \\
        $\text{SMT2}_{\text{HH}}$ & $-$ & $-$ & $-$ & $-$ & $0.0159$ & $0.2899$ & $0.8912$ & $10.8922$ & $1.4979$ & $9.6483$ \\
        $\text{SMT2}_{\text{EHH}}$ & $-$ & $-$ & $-$ & $-$ & $0.0126$ & $0.2203$ & $3.2678$ & $21.9938$ & $1.8296$ & $8.9458$ \\\hline
    \end{tabular}
    \caption{\textbf{Comparison of Emulation Accuracy:} We test the performance of \gp~in emulation against GPyTorch, Matlab, and MST-GP on five examples. The reported NRMSE and NIS are for unseen data and averaged across $10$ repetitions.}
    \label{table: results_emulation}
\end{table}

\gp~outperforms all baselines in \boreholeM, \hoip, and \nta~ which have categorical variables. In these problems, \gp~explicitly learns the relations between different categorical variables and their levels which not only improve the emulation performance, but also provide visually interpretable embeddings (see \Cref{fig: fidelity_borehole,fig: fidelity_alloy,fig: fidelity_HOIP} in \Cref{sec: BO} for an example). 
{\color{black}We note \gp~estimate more parameters than other methods in \Cref{table: results_emulation} since it directly learns the correlations among categorical variables. This approach results in a more expensive and challenging optimization process which can converge to suboptimal solutions if the training data is very small and the categorical variables have many levels. 
To mitigate potential overfitting issues in such cases, we recommend using tighter priors in $f_h(\boldsymbol{\pi_{\boldsymbol{t}}}, \boldsymbol{\theta}_h)$. Correspondingly, in \nta~and \hoip~with $240$ and $480$ distinct categorical combinations, respectively, we use $\mathcal{N}(0,0.1)$ and $\mathcal{N}(0,0.01)$ priors for $f_h(\boldsymbol{\pi_{\boldsymbol{t}}}, \boldsymbol{\theta}_h)$.}

\subsection{Multi-fidelity Modeling} \label{sec: mfmodeling_functionality}

In this section, we assess the performance of \gp~in MF emulation by comparing it against widely used emulators. Our baselines include vanilla GPs trained only on the HF data (V-GP), FFNNs, and single-task multi-fidelity GPs (STMF-GPs) introduced by BoTorch (detailed in \Cref{sec: STMF_GP}). Furthermore, we examine different versions of \gp~with distinct basis functions explained in \Cref{sec: gp_mixed_basis_functions}. The mean function in these versions are formulated as a single constant, multiple constants (the number of constants is $ds-1$, as we consider zero mean for HF source), and finally an FFNN.

Since the performance of FFNNs is sensitive to their architecture, we design small, medium, and large networks and for each network size test many different scenarios and report the results of the most accurate ones (see \Cref{sec: appendix_MF_details} for details). In the case of STMF-GP, as detailed in \Cref{sec: STMF_GP} the fidelity indices are numerical and must reflect the relative accuracy of the data sources. STMF-GP lacks a built-in metric for determining these indices and relies on the user to provide these values. To address this issue, we first leverage the learnt embedding (i.e., the $z-$space) of \gp~to find the order of these indices and then assign two different sets of values to them to assess this method's sensitivity to the assigned values. These values are outlined in \Cref{table: mfst_gp_fidelities} and we denote the corresponding models by $STMF-GP_1$ and $STMF-GP_2$. 

We use two analytic (\sine~and \wing) and one engineering (\DNS) examples for the comparison (see \Cref{table: analytic-formulation} and \cite{deng2023data} for details on these examples). \sine~is a $1$-dimensional, bi-fidelity example for which we generate a dataset consisting of $400$ HF and $2000$ LF samples. \wing~has $4$ fidelity sources ($1$ HF and $3$ LFs) and we produce $1500$ samples from the HF source and $4000$ samples from each of the LF sources. In both analytic examples, $1\%$ of data is used for training and the rest of the HF data for testing. \DNS~is a $5$-dimensional problem on fracture modeling of metallic alloys where the data are generated via four different simulators with $70, 110, 170, 250$ samples. In this example, we use $20\%$ of the samples for training and the rest of the HF data for testing.

The results for each approach on each problem are summarized in \Cref{table: results_mfmodeling} and demonstrate that \gp~significantly outperforms the other baselines in all problems. More specifically, while V-GP is limited to the small HF data, \gp~effectively leverages the information provided by the LF data to learn the HF source. 
The poor performance of STMF-GP is due to the fact that from a methodological standpoint it models the inter-relations between the data sources incorrectly. In addition to providing low accuracy, the pblackictions of STMF-GP are sensitive to the values assigned to its fidelity indices. This is evident in \Cref{table: results_mfmodeling} where the pblackiction errors for two different yet close sets of random indices ($STMF-GP_1$ and $STMF-GP_2$, see also \Cref{table: mfst_gp_fidelities}) are very different. 
Regarding FFNNs, we attribute their poor performance in all problems to their architecture and, in particular, {\color{black}their simple mechanism for handling fidelity levels.} These FFNNs simply one-hot encode the fidelity indices and ignore the intricate correlations among the corresponding data sources. 
Compablack to other methods, the reported NRMSEs for FFNNs are more sensitive to the model architecture and notably change as the network size varies. This sensitivity is partly due to the small size of the MF data and can perhaps be improved by iteratively refining the architecture or the optimization parameters (e.g., learning rate schedule or regularization weights). However, we avoid such detailed refinements since none of the other methods are fine-tuned. 


Comparing the results of different versions of \gp~reveals that in all cases using mixed bases improves MF modeling by better capturing the global and local features of each source (compare the first row to other mean functions). This choice benefits NIS slightly more than NRMSE since the former metric relies on both the expected value and estimated variance, i.e., $\tau^2({\boldsymbol{u}^*}^{(i)})$. For instance, \gp~with multiple constants as $m(\ub; \betab)$ and medium FFNN $m(\ub; \betab)$ achieve similar NRMSEs ($0.2201$ vs $0.2062$), but their NIS significantly differs ($0.9274$ vs $0.5475$) in \sine.

\wing~and \DNS~are relatively complex problems with small amounts of data and different types of noise (e.g., in\DNS~the noise variance depends on the source while in \wing~it does not). \gp~is very well suited to tackle these types of problems because the number of its hyperparameters scales much better than FFNNs, is not limited to the small HF data, and better estimates noise as explained in \Cref{sec: multi_noise}. Accordingly, we observe lower pblackiction errors for \gp~in these examples.

\begin{table}[]
\centering
\small
\begin{tabular}{|c|c|cc|cc|cc|}
\hline
\multirow{2}{*}{Model}  & \multirow{2}{*}{Option} & \multicolumn{2}{c|}{\sine} & \multicolumn{2}{c|}{\wing} & \multicolumn{2}{c|}{\DNS} \\ \cline{3-8} 
                         &           & NRMSE  & NIS    & NRMSE  & NIS    & NRMSE  & NIS    \\ \hline
\multirow{4}{*}{\gp} & $\text{Single constant as } m(\ub; \betab)$     & $0.2501$ & $1.2070$ & $0.0743$ & $0.4294$ & $0.1572$ & $0.9051$ \\ \cline{2-2}
                         & $\text{Multiple constants as } m(\ub; \betab)$     & $0.2201$ & $0.9274$ & $\textbf{0.0729}$ & $0.4419$ & $0.1560$ & $0.8935$ \\ \cline{2-2}
                         & $\text{Small FFNN as } m(\ub; \betab)$     & $\textbf{0.1999}$ & $0.8261$ & $0.0751$ & $\textbf{0.3884}$ & $0.1535$ & $0.8561$ \\ \cline{2-2}
                         & \text{Medium FFNN as } $m(\ub; \betab)$     & $0.2062$ & $\textbf{0.5475}$ & $0.0751$ & $0.4072$ & $\textbf{0.1528}$ & $\textbf{0.8477}$ \\ \hline
V-GP                     & $-$         & $0.4156$ & $1.9842$ & $0.1794$ & $0.9152$ & $0.2101$ & $1.0856 $\\ \hline
\multirow{3}{*}{FFNN}    & $\text{Small}$     & $0.8076$ & $-$      & $0.6295$ & $-$      & $0.2693$ & $-$      \\ \cline{2-2}
                         & $\text{Medium}$    & $0.6238$ & $-$      & $0.4320$ & $-$      & $0.2297$ & $-$      \\ \cline{2-2}
                         & $\text{Large}$     & $0.5244$ & $-$      & $0.3543$ & $-$      & $0.2221$ & $-$      \\ \hline
\multirow{2}{*}{STMF-GP} & $STMF-GP_1$ & $0.4835$ & $6.7312$ & $0.1219$ & $1.0125$ & $0.1618$ & $1.0625$ \\ \cline{2-2}
                         & $STMF-GP_2$ & $0.5362$ & $8.5698$ & $0.2001$ & $1.1661$ & $0.1707$ & $0.9651$ \\ \hline
\end{tabular}
\caption{\textbf{Multi-fidelity emulation:} We test the performance of \gp\ in various settings against V-GP, STMF-GP, and FFNN across three examples and report NRMSE and NIS on unseen HF data.}
    \label{table: results_mfmodeling}
\end{table}


\subsubsection{Deterministic and Probabilistic Embedding} \label{sec: Pro_vs_Det_compare_functionality}

As explained in \Cref{subsec: det-mfmodeling,subsec: prob-mfmodeling}, one of the distinctive features of \gp~is its ability to learn both probabilistic and deterministic embeddings for MF modeling. We revisit the \wing~example with smaller datasets to evaluate \gp's efficacy in data-scarce scenarios. Specifically, we generate $1000$ samples from the HF source and $2000$ samples from each LF source and use $1\%$ for training and the rest for testing. Throughout this section, $f_z(\pib_s; \thetab_z)$) is an FFNN with a single five-neuron hidden layer and we use the multiple constants option of \gp~to model the mean function of the GPs\footnote{With this option, a constant is learnt for each of the LF sources since the data is generated by four sources in \wing~and $0$ is used for the HF source}. 
Similar to the previous sections, we repeat both deterministic and probabilistic simulations $10$ times.

\begin{figure}[!t]
  \centering
    \begin{subfigure}{0.49\textwidth}
    \includegraphics[width=\textwidth]{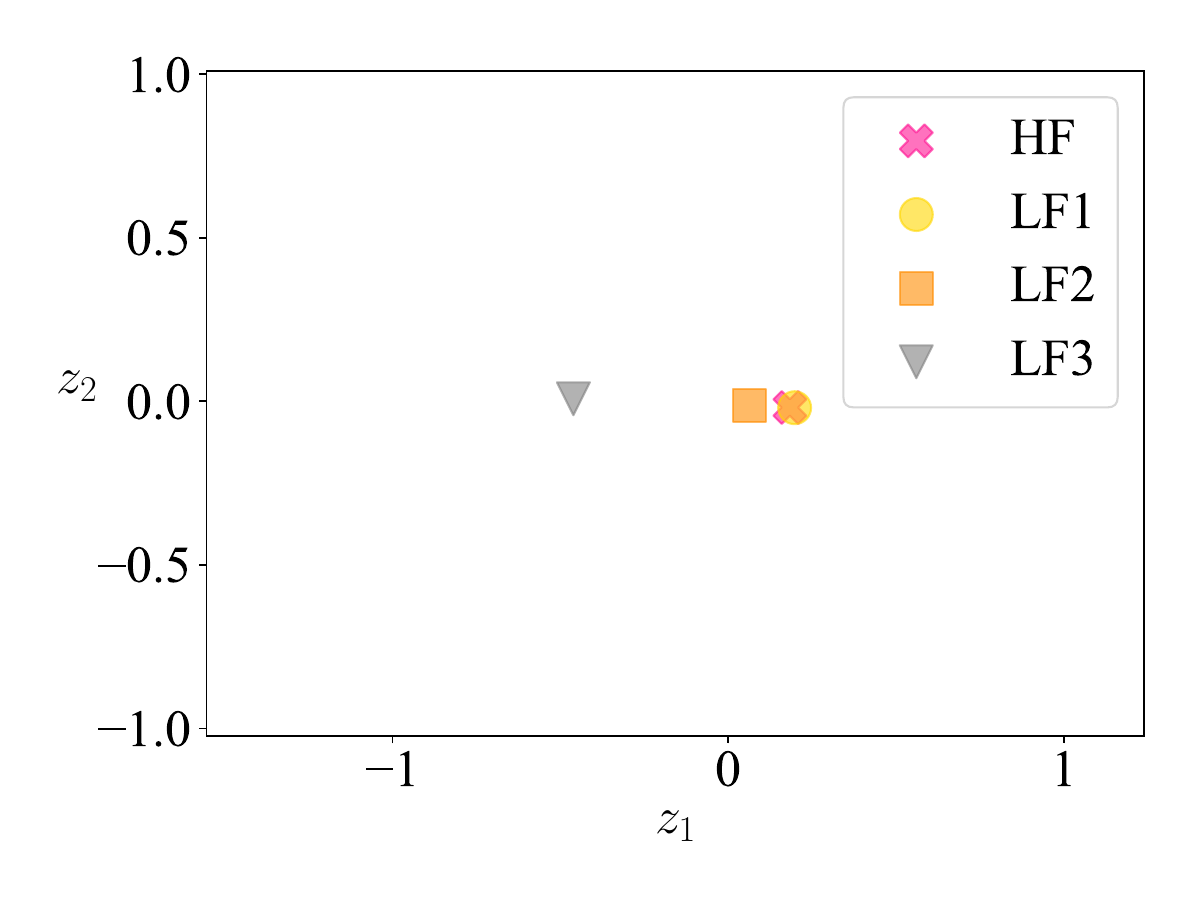}
    \vspace{-7mm}
    \caption{\textbf{Deterministic Embedding}}
    \label{fig: manifold_det_wing}
    \end{subfigure}
    \begin{subfigure}{0.49\textwidth}
    \includegraphics[width=\textwidth]{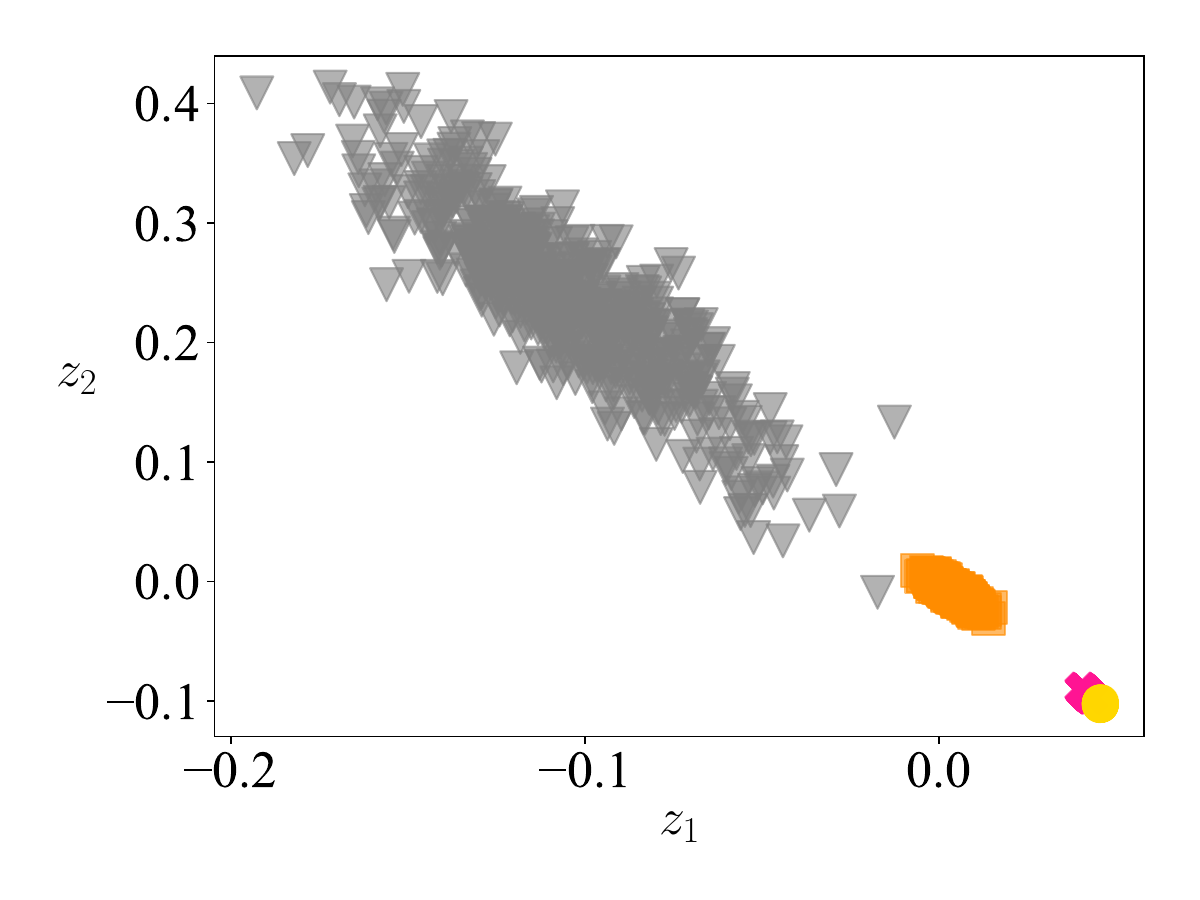}
    \vspace{-7mm}
    \caption{\textbf{Probabilistic Embedding}}
    \label{fig: manifold_pro_wing}
    \end{subfigure}
    \caption{\textbf{Probabilistic vs deterministic embedding for MF modeling:} While both embeddings estimate the same degree of similarity among sources, {\color{black}the probabilistic one characterizes more uncertainties.}}
    \label{fig: manifold_wing}
\end{figure}

\begin{figure}[!h]
  \centering
    \begin{subfigure}{0.49\textwidth}
    \includegraphics[width=\textwidth]{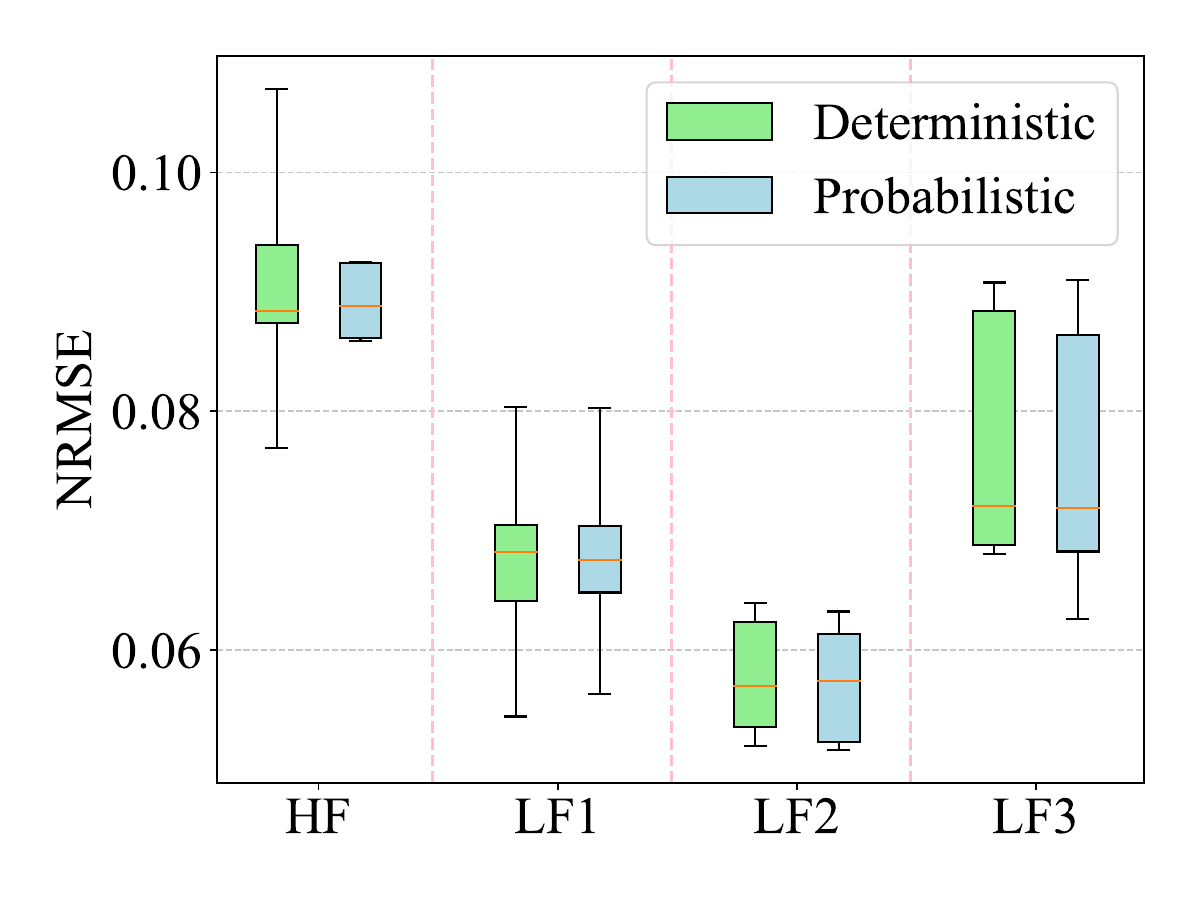}
    \vspace{-7mm}
    \caption{\textbf{NRMSE}}
    \label{fig: Wing_prob_det_NRMSE}
    \end{subfigure}
    \begin{subfigure}{0.49\textwidth}
    \includegraphics[width=\textwidth]{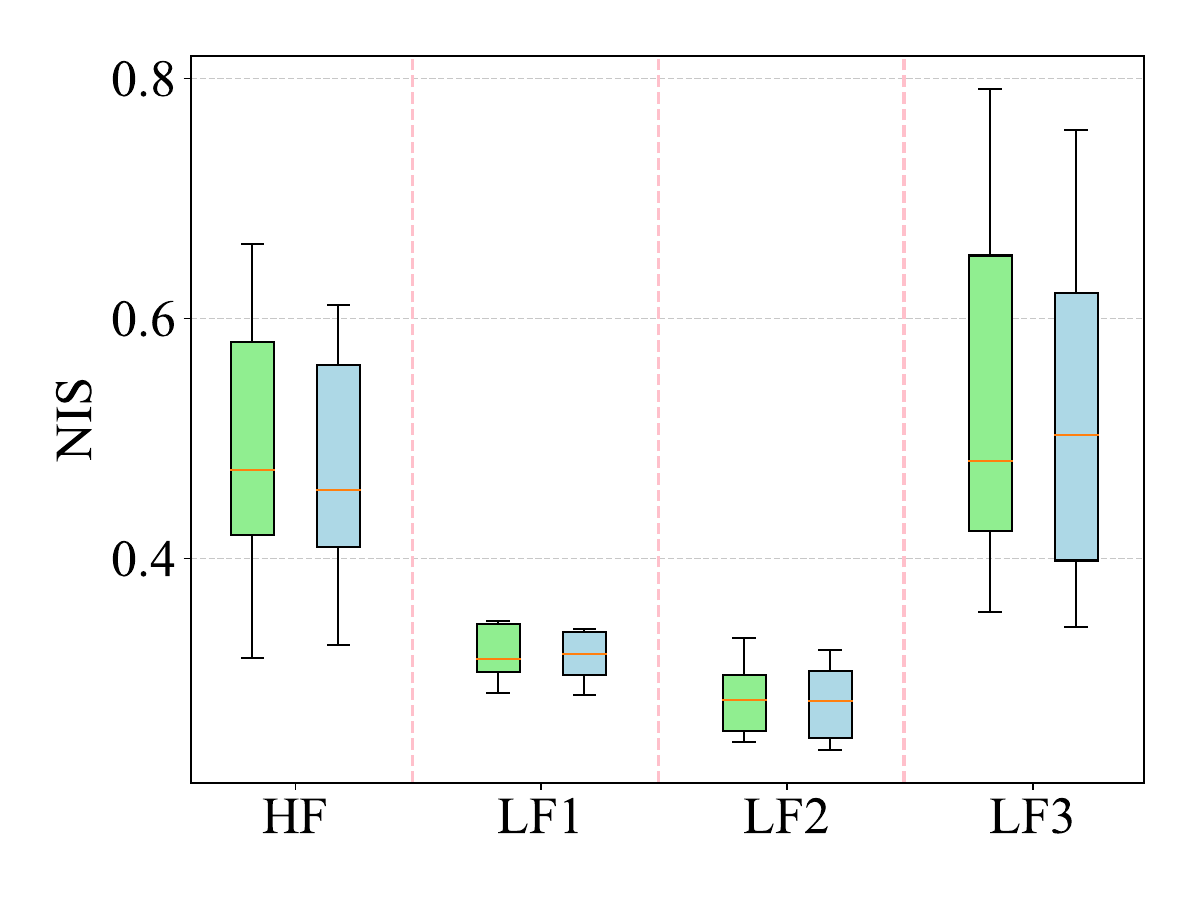}
    \vspace{-7mm}
    \caption{\textbf{NIS}}
    \label{fig: Wing_prob_det_NIS}
    \end{subfigure}
    \caption{\textbf{Pblackiction performance with probabilistic vs deterministic embedding:} We report NRMSE and NIS values calculated across $10$ repetitions. Both approaches are very accurate but the probabilistic one is more robust for HF emulation especially since HF data is very small. The similar NRMS and NIS values for LF1 and LF2 are expected because of their high correlation.}
    \label{fig: Wing_prob_det_results}
\end{figure}

\Cref{fig: manifold_wing} illustrates the fidelity embeddings learned through probabilistic and deterministic MF modeling. We only show the learnt embeddings in one of the repetitions since the relative distances across the $10$ repetitions are quite similar (albeit the exact locations are different). As explained in \Cref{sec: GP_Plus}, these embeddings indicate how similar or correlated different data sources are with respect to each other. Specifically, the latent distance between the points encoding the LF sources from the point that encodes the HF source are consistent with the NRMSE values reported in \Cref{table: analytic-formulation} where, e.g., the furthest encoded source (LF3) has the largest NRMSE ($5.57$). 
In addition, we observe in \Cref{fig: manifold_wing} that while the general trends are the same across the two embeddings (e.g., LF3 and LF1 are the furthest and closest to HF), probabilistic embedding more accurately quantifies model form uncertainties especially in the case of highly biased LF sources. 

\Cref{fig: Wing_prob_det_results} compares the pblackiction accuracy on unseen data with probabilistic and deterministic embeddings. As it can be observed, the probabilistic approach is slightly more robust in HF emulation but both approaches $(1)$ provide the same degree of accuracy and robustness for LF sources, and $(2)$ are less accurate in emulating HF and LF3 sources. The reason LF is learnt less accurately than other LF sources is its low correlation while the errors in emulating the HF source primarily stem from the lack of HF data.

\subsection{Inverse Parameter Estimation} \label{sec: Calibration_Functionality}
In this section, we compare the performance of \gp~in inverse parameter estimation against the method of KOH and the open-source software package \uql~\cite{marelli2014uqlab}. As schematically demonstrated in \Cref{fig: fusion-graph}\textbf{(a)}, KOH uses an additive bias and can calibrate a single LF source at a time. In our studies, we consider two versions of this method where its parameters (e.g., the kernel parameters of LF and HF sources) are estimated either jointly or via a modular scheme \cite{RN791, RN270, RN645} which first estimates the parameters of the LF source and then optimizes the rest of the parameters. 
The calibration module of \uql~relies on Bayesian inference and leverages MCMC for parameter estimation. Similar to previous sections, we repeat each of our studies $10$ times and report the average values.

As detailed in \Cref{sec: analytic_example}, \beam~is a $5$-dimensional bi-fidelity example where the objective is to infer a beam's Young’s modulus ($\zeta$) whose ground truth value is $30$ GPa. This example is directly taken from the documentation of \uql~where there is only one HF data point and the difference between the LF and HF sources is a zero-mean noise. To explore the effects of prior distributions on the results, we assign three different priors to $\zeta$ while comparing \gp~with \uql. While $200$ LF samples are used in \gp, \uql~leverages the analytic form of the LF model in MCMC.

The results presented in \Cref{table: cal_GP+ vs UQLab} indicate that both methods estimate similar posterior means for the calibration parameter if the prior is relatively precise. However, the two methods behave quite differently as the prior mean is shifted away from the ground truth value. Specifically, \uql~provides posterior means that are quite close to the prior means while \gp~is significantly less sensitive to imprecise priors. In the case of \gp~, we note that the deterministic approach does not provide uncertainties and the reported standard deviations for the probabilistic approach are very small. These tight posteriors are expected since the model has access to sufficient LF data and the bias between the sources is a Gaussian noise. 

\begin{table}[!h]
    \centering
    \caption{\textbf{Inverse estimation of Young's Modulus using \gp~and \uql:} The posterior mean and standard deviations provided by \uql~is very sensitive to the assigned prior while \gp~does not suffer from this issue. The reported uncertainties by \gp~are small since the bias between LF and HF sources is simple and the model has access to sufficient LF data.}
    \label{tab:youngs_modulus_estimates}
    \begin{tabular}{|c|c|c|c|}
        \hline
        Prior on $\zeta$ & Method & Estimated mean (GPa) & Estimated std (GPa) \\ 
        \hline
        \multirow{3}{*}{$\mathcal{N}(30, 5)$} & \uql & \textbf{30.0101} & 4.6832  \\
        \cline{2-4}
         & Probabilistic \gp & 29.2363 & 15.867 $10^{-6}$ \\  
        \cline{2-4}
         & Deterministic \gp & 29.2304 & - \\
        \hline
        \multirow{3}{*}{$\mathcal{N}(25, 5)$} & \uql & 26.1540 & 4.3772  \\
        \cline{2-4}
         & Probabilistic \gp & 29.0311 & 29.320 $10^{-6}$ \\ 
        \cline{2-4}
         & Deterministic \gp & \textbf{29.0449} & - \\
        \hline
        \multirow{3}{*}{$\mathcal{N}(20, 5)$} & \uql & 22.2348 & 4.3355 \\
        \cline{2-4}
         & Probabilistic \gp & \textbf{29.1462} & 10.3699 $10^{-5}$ \\ 
        \cline{2-4}
         & Deterministic \gp & 28.9418 & - \\
        \hline
    \end{tabular}
\label{table: cal_GP+ vs UQLab}
\end{table}
\begin{figure}[!h]
  \centering
    \begin{subfigure}{0.49\textwidth}
    \includegraphics[width=\textwidth]{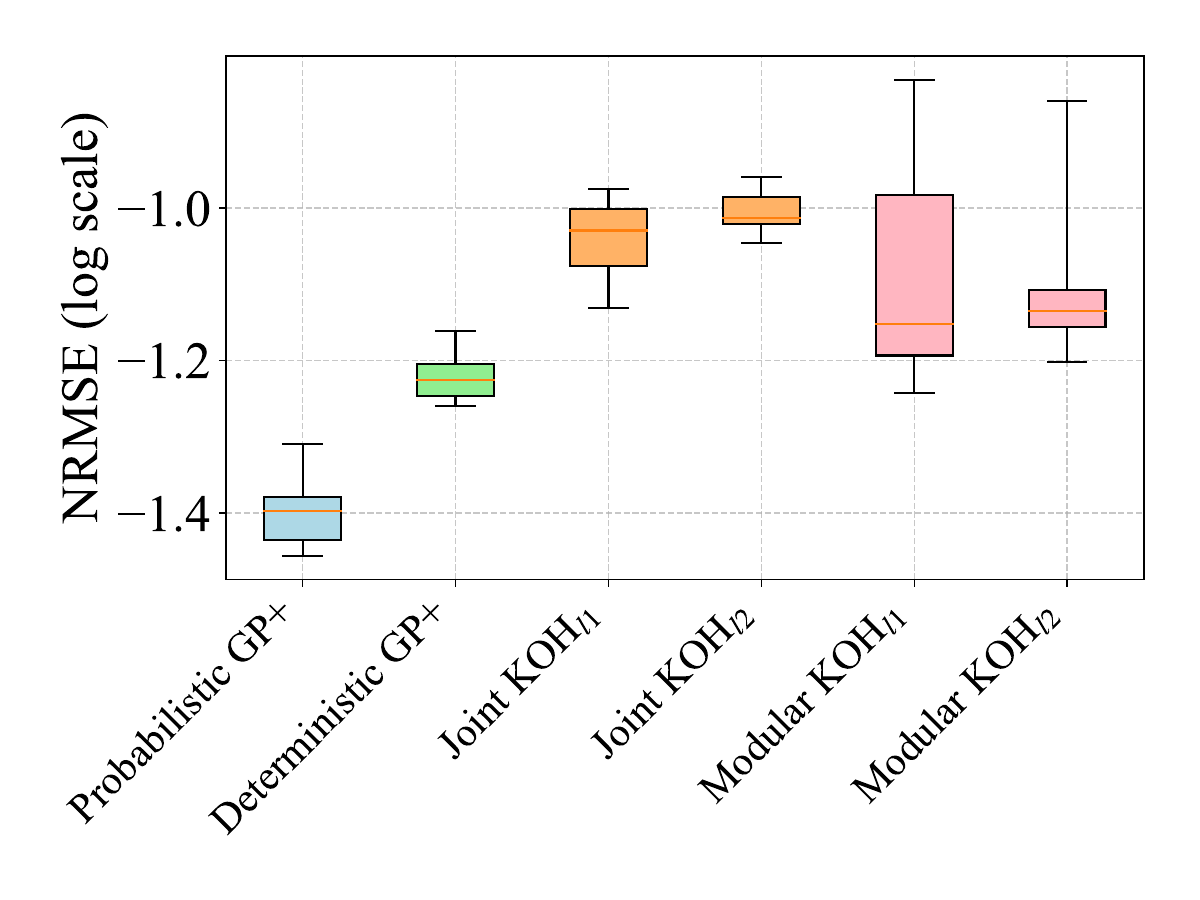}
    \vspace{-7mm}
    \caption{\textbf{}}
    \label{fig: Cal_Borehole_NRMSE}
    \end{subfigure}
    \begin{subfigure}{0.49\textwidth}
    \includegraphics[width=\textwidth]{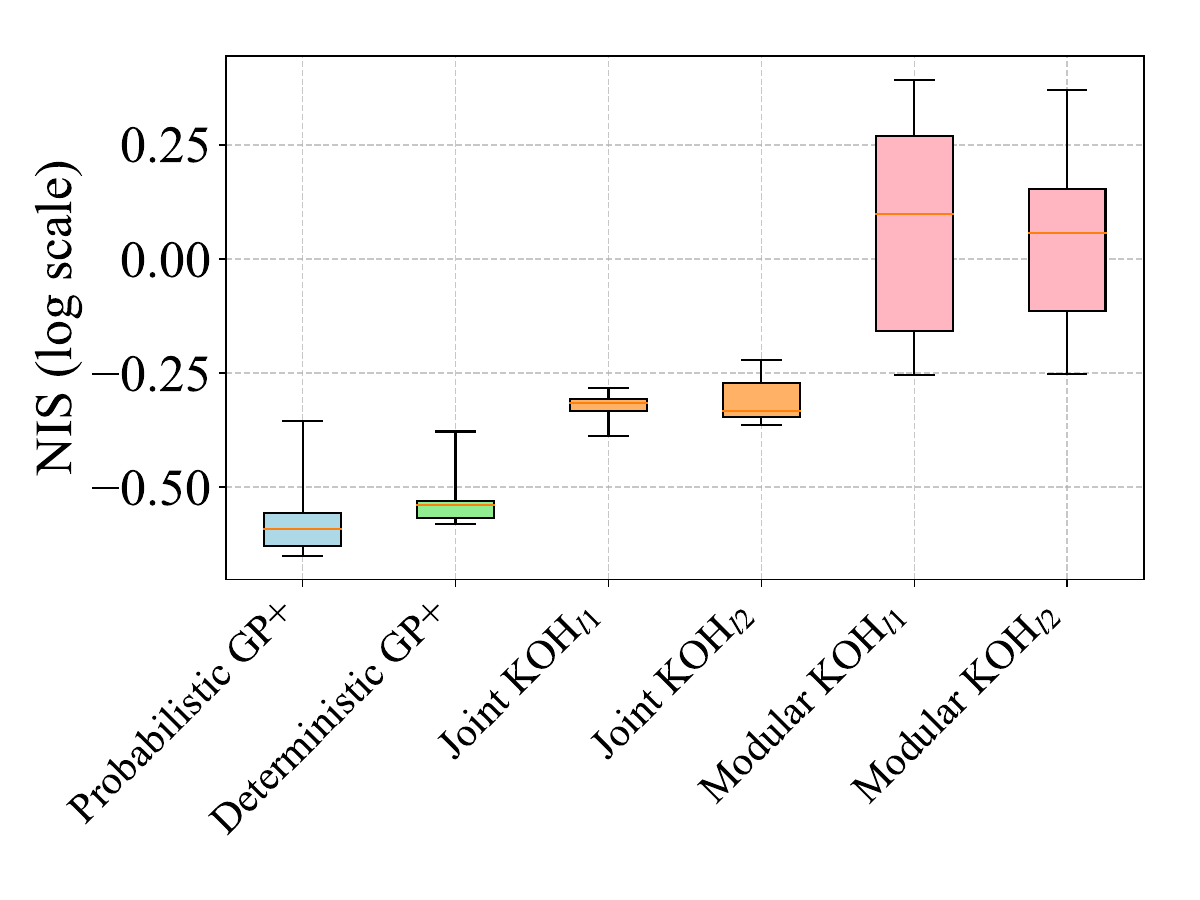}
    \vspace{-7mm}
    \caption{\textbf{}}
    \label{fig: cal_boreholre_NIS}
    \end{subfigure}
    \caption{\textbf{High-fidelity emulation performance in \borehole:} \gp~consistently outperforms KOH’s approach. The superior performance of both variations of \gp~are primarily attributed to using all the data jointly and dispensing with the assumption that model form errors are additive.}
    \label{fig: cal_boreholre_NRMSE_NIS}
\end{figure}

We now use the high-dimensional \borehole~problem presented in \Cref{table: analytic_calibration} with $1$ HF and $2$ LF sources that both have nonlinear model form errors (note that LF1 is more biased compablack to LF2, see the NRMSEs in \Cref{table: analytic_calibration}). We generate $20$ and $100$ samples from, respectively, the HF and each of the LF sources. We only corrupt the HF training samples with noise and use $1800$ noise-free HF samples for evaluation of models' performance.  
There are two calibration parameters in this example and their ground truth values are $\zeta_1 = 250$ and $\zeta_2 = 1500$. We compare the calibration results obtained by \gp~with those from KOH whose parameters are estimated either jointly or by a modular approach. Since KOH can only fuse two sources at a time, we consider different combinations of LF and HF sources in our experiments.

As shown in \Cref{fig: cal_boreholre_NRMSE_NIS}, both configurations of \gp~convincingly outperform KOH’s approach in terms of both NRMSE and NIS. 
We attribute this superior performance to three key factors: $(1)$ \gp's ability to utilize all data simultaneously while KOH's approach can only work with two sources at a time (expectedly, modular KOH with LF1 has the least accuracy), $(2)$ \gp's capability to capture nonlinear correlations whereas KOH is confined to learning additive model form errors, and $(3)$ \gp~leverages a more stable and regularized training procedure. We note that probabilistic \gp~outperforms its deterministic counterpart as it learns a posterior distribution for $\zeta$ rather than just a single point estimate. While probabilistic calibration improves the performance on average, it shows more variability across the $10$ repetitions primarily due to the fact that it has more parameters than the deterministic one.

\begin{figure}[!t]
  \centering
    \begin{subfigure}{0.49\textwidth}
    \includegraphics[width=\textwidth]{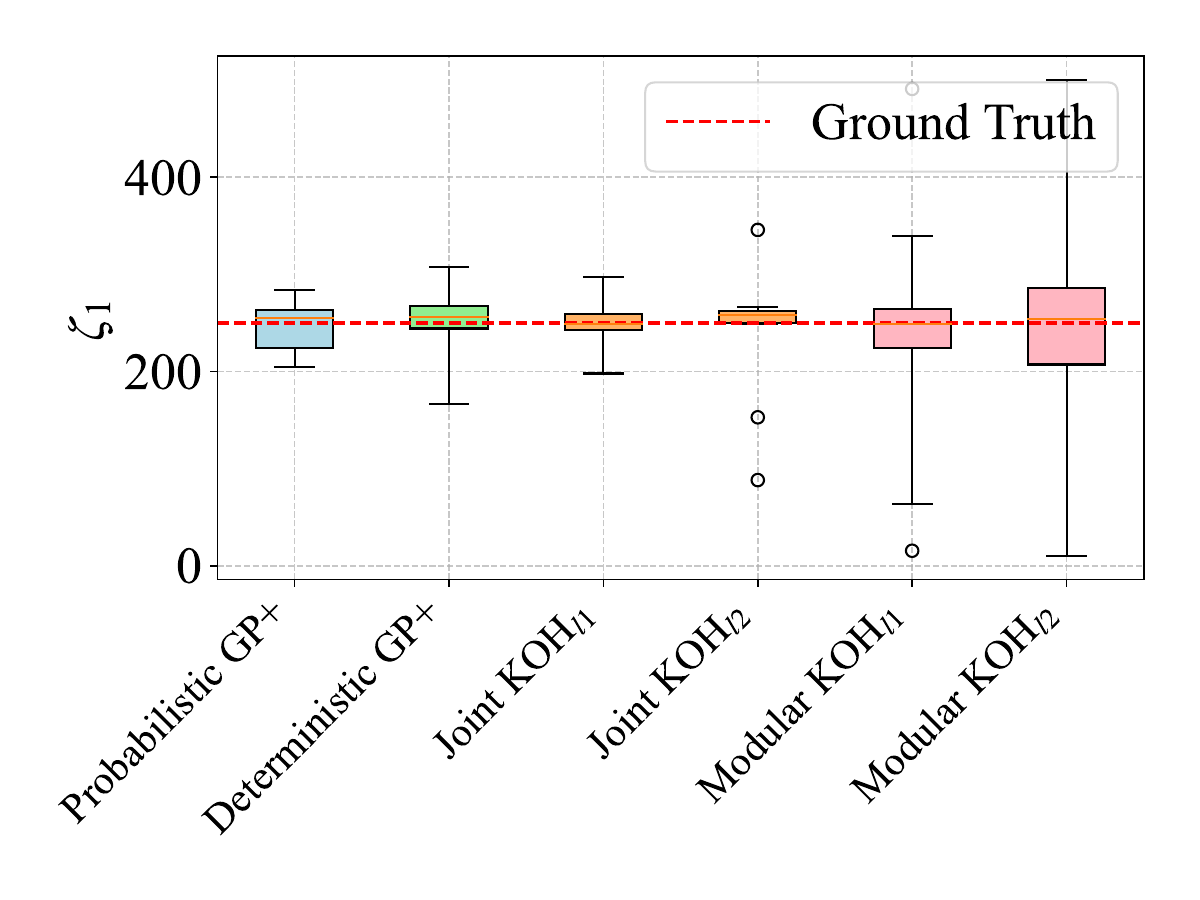}
    \vspace{-7mm}
    \caption{\textbf{}}
    \label{fig: cal_borehole_parameter_1}
    \end{subfigure}
    \begin{subfigure}{0.49\textwidth}
    \includegraphics[width=\textwidth]{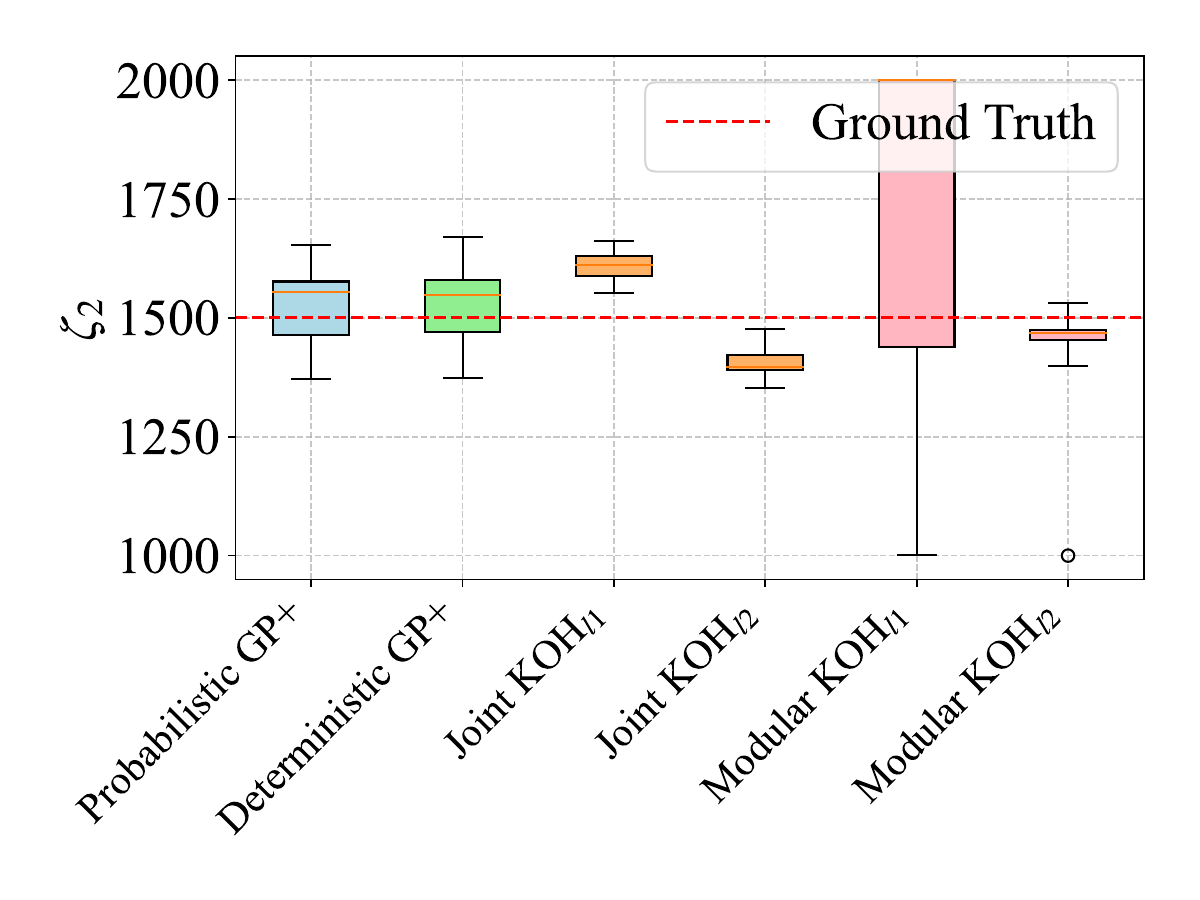}
    \vspace{-7mm}
    \caption{\textbf{}}
    \label{fig: cal_borehole_parameter_2}
    \end{subfigure}
    \caption{\textbf{Calibration performance in \borehole:} Ground truth values are the numbers used to generate the HF samples. While all methods estimate $\zetab$ quite well, probabilistic \gp~performs the best.}
    \label{fig: cal_borehole_parameters}
\end{figure}

The superiority of \gp~in emulation is coupled with a more accurate estimation of the calibration parameters as illustrated in \Cref{fig: cal_borehole_parameters}. 
While the estimated values are quite accurate across all methods, probabilistic \gp~provides slightly better results, especially compablack to modular KOH that fuses the data in a sequence of steps rather than jointly. We also observe that estimations for $\zeta_1$ in \Cref{fig: cal_borehole_parameter_1} are generally more accurate than those of $\zeta_2$ in \Cref{fig: cal_borehole_parameter_2}. This trend is primarily due to the fact that the underlying functions (i.e., HF and both LF sources) are more sensitive to $\zeta_1$ than $\zeta_2$ (see \Cref{table: SA_borehole} for sensitivity analysis of \borehole).

We now revisit a variation of the \wing~problem which has four calibration parameters whose ground truth values are $\zetab^{*T} = [40, 0.85, 0.17, 3]$ (these values correspond to the numbers used in the HF source). As detailed in \Cref{table: analytic_calibration}, there are four data sources in \wing~where LF1 and LF3 are the most and least accurate LF sources, respectively. We take a small number of samples from each of the four sources and corrupt all the data with noise. Similar to the previous study, we compare probabilistic and deterministic calibration capabilities of \gp~against two versions of KOH. Throughout, we repeat the simulations $10$ times and use $2500$ noise-free HF samples for testing the emulation performance. 

\begin{figure}[!h]
  \centering
    \begin{subfigure}{0.49\textwidth}
    \includegraphics[width=\textwidth]{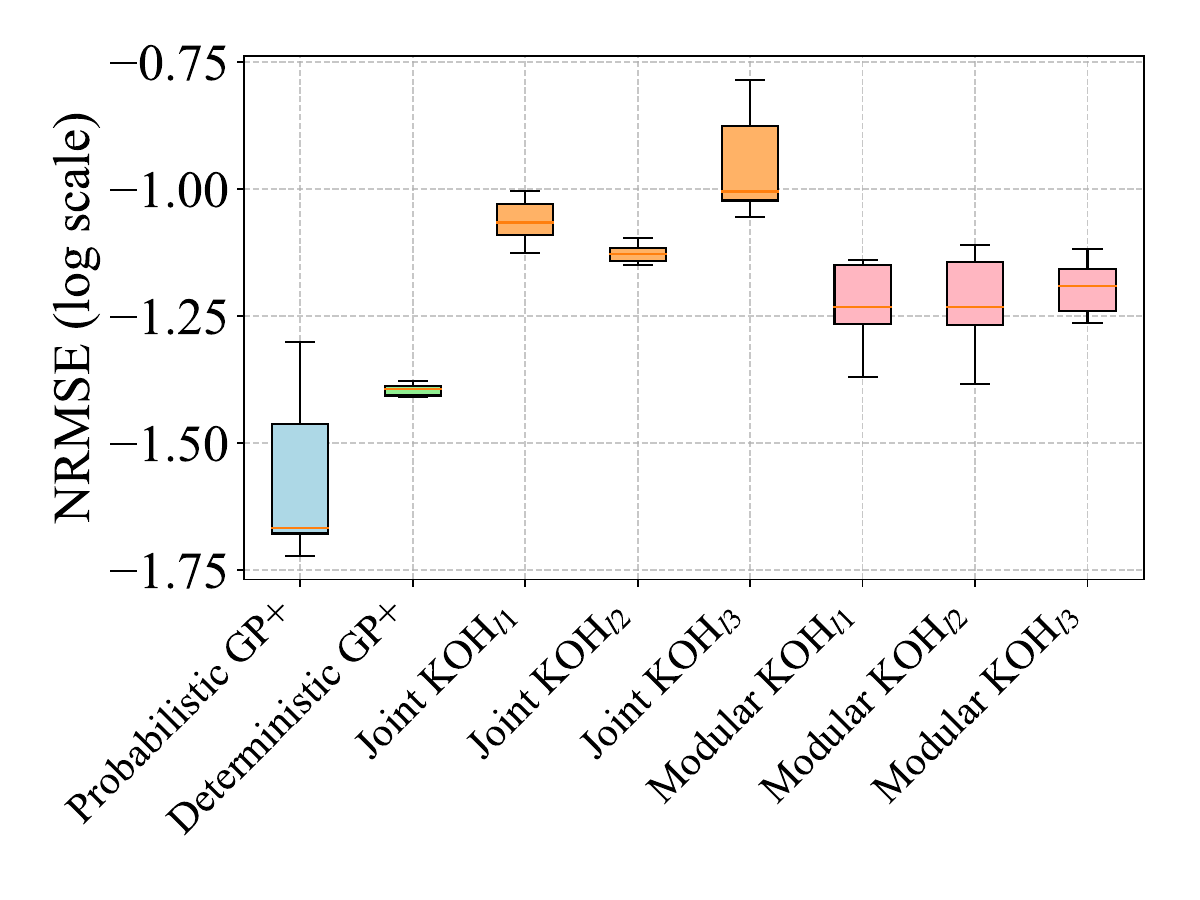}
    \vspace{-7mm}
    \caption{\textbf{}}
    \label{fig: fidelity_borehole}
    \end{subfigure}
    \begin{subfigure}{0.49\textwidth}
    \includegraphics[width=\textwidth]{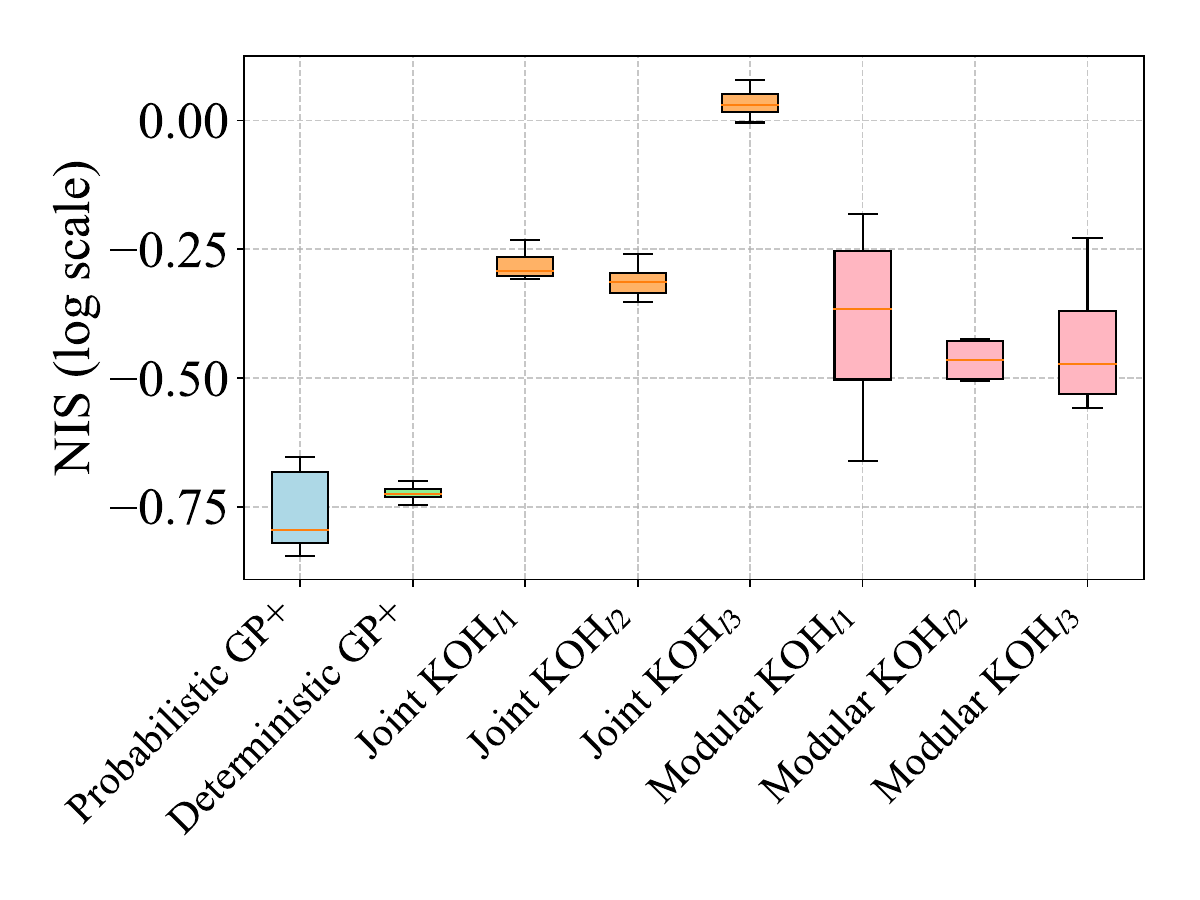}
    \vspace{-7mm}
    \caption{\textbf{}}
    \label{fig: fidelity_alloy}
    \end{subfigure}
    \caption{\textbf{High-fidelity emulation performance in \wing:} \gp~strategies outperform KOH’s approach which can only fuse two sources at a time and uses an additive formulation for the model form errors.}
    \label{fig: calibration_NRMSE_NIS}
\end{figure}

\begin{figure}[!h]
  \centering
    \begin{subfigure}{0.49\textwidth}
    \includegraphics[width=\textwidth]{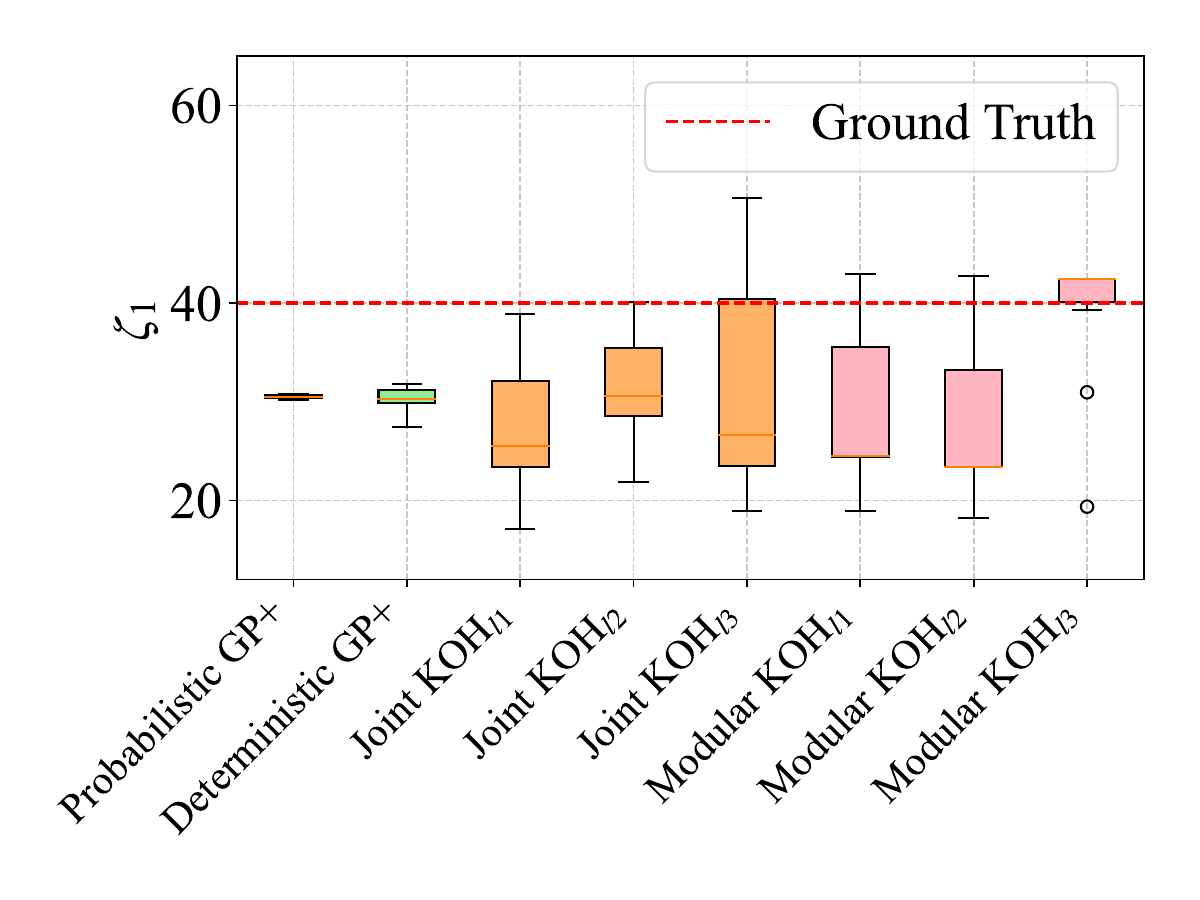}
    \vspace{-7mm}
    \caption{\textbf{}}
    \label{fig: cal_wing_parameter_1}
    \end{subfigure}
    \begin{subfigure}{0.49\textwidth}
    \includegraphics[width=\textwidth]{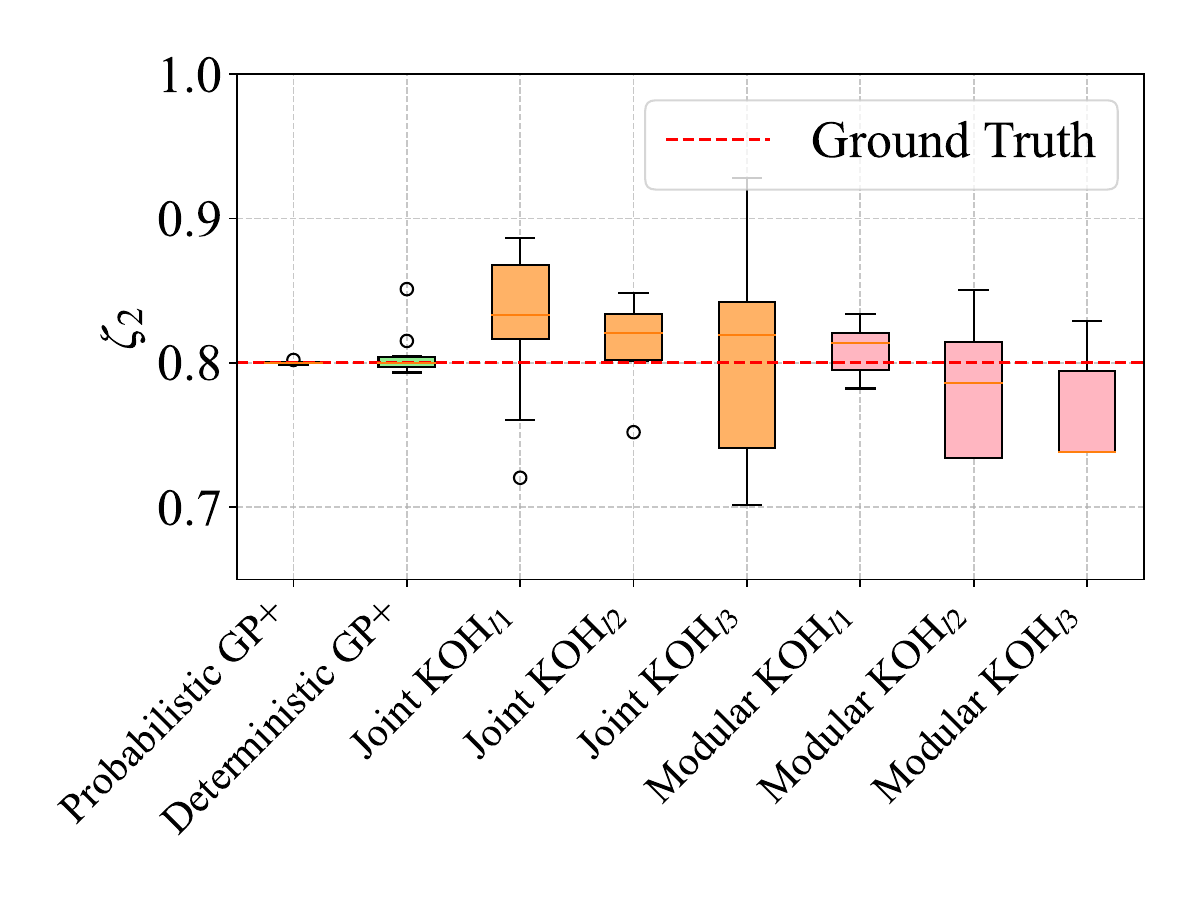}
    \vspace{-7mm}
    \caption{\textbf{}}
    \label{fig: cal_wing_parameter_2}
    \end{subfigure}
    \begin{subfigure}{0.49\textwidth}
    \includegraphics[width=\textwidth]{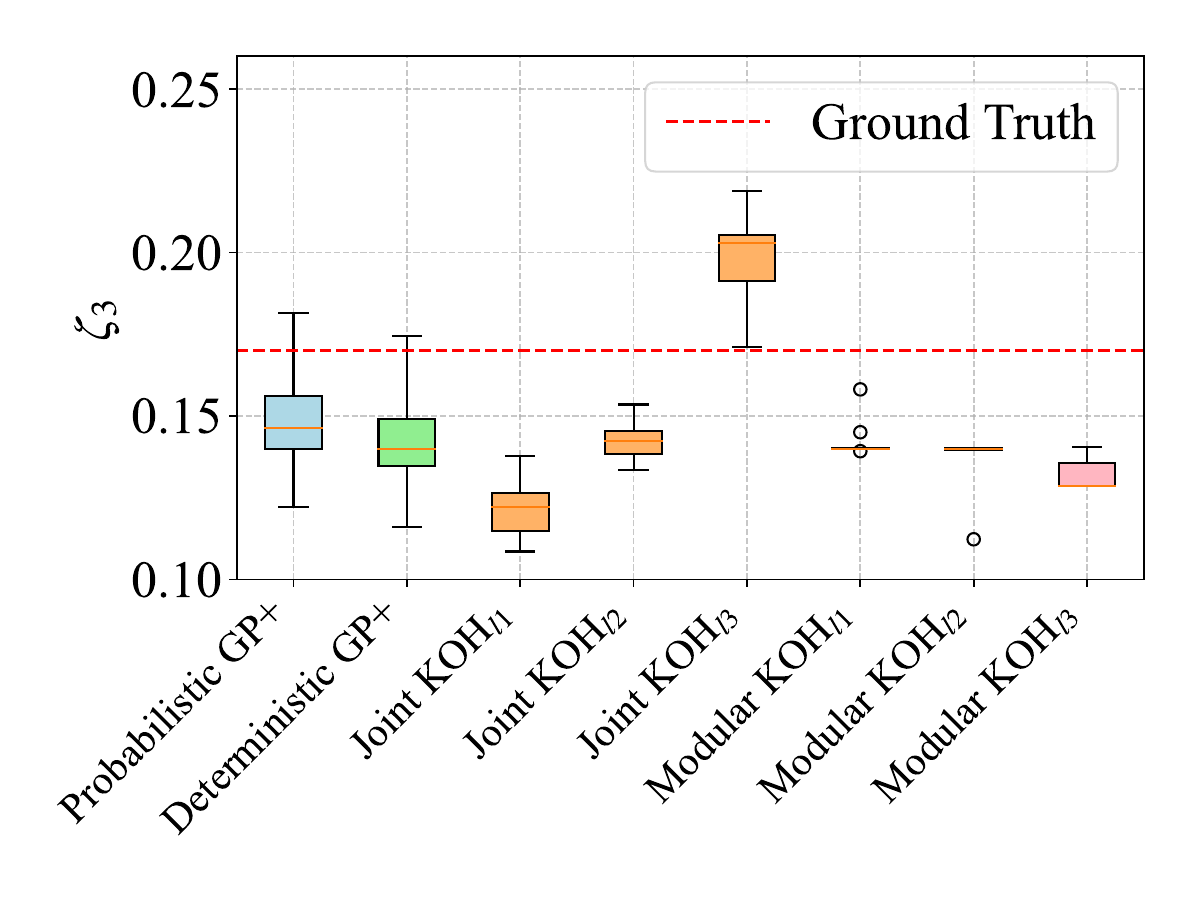}
    \vspace{-7mm}
    \caption{\textbf{}}
    \label{fig: cal_wing_parameter_3}
    \end{subfigure}
    \begin{subfigure}{0.49\textwidth}
    \includegraphics[width=\textwidth]{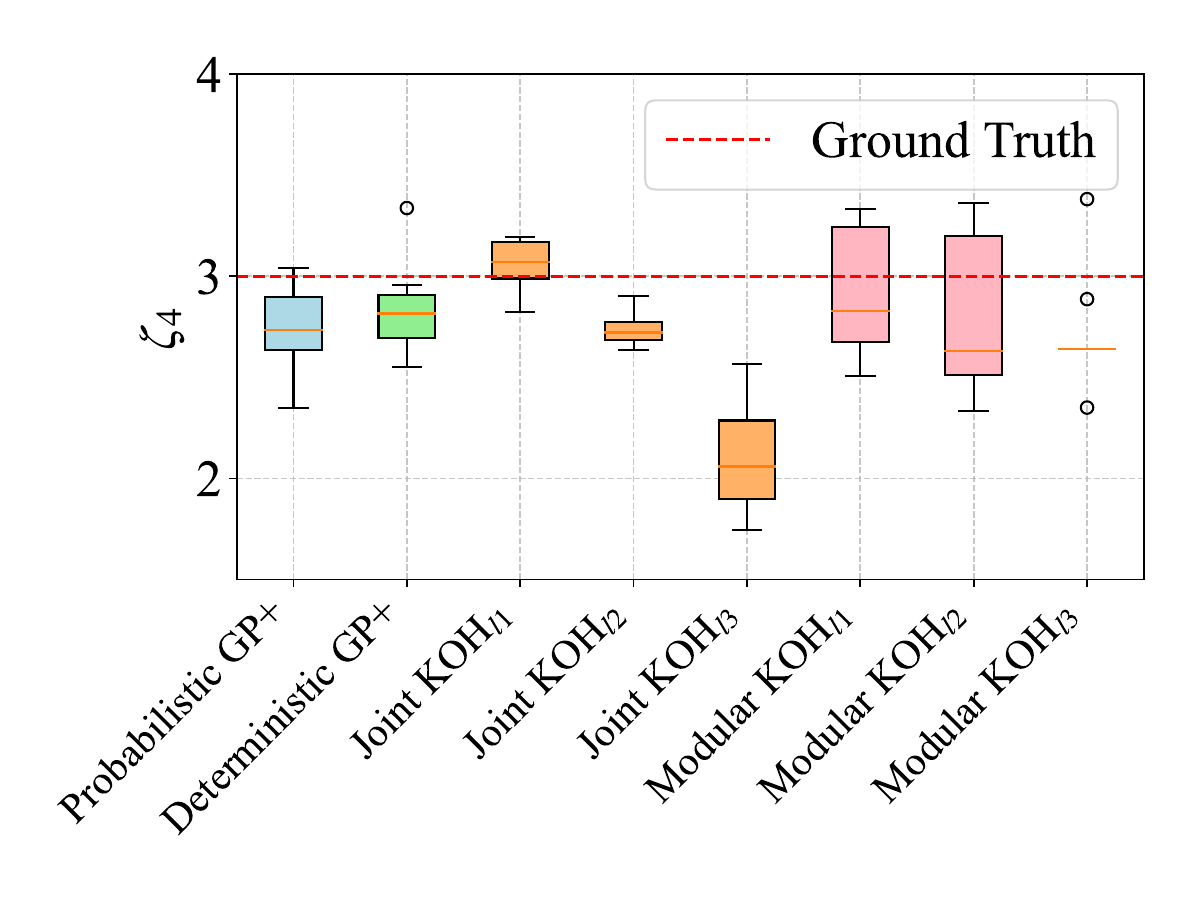}
    \vspace{-7mm}
    \caption{\textbf{}}
    \label{fig: cal_wing_parameter_4}
    \end{subfigure}
    \caption{\textbf{Calibration performance in \wing:} Ground truth values are the numbers used to generate the HF samples. While all methods estimate $\zetab$ quite well, probabilistic \gp~performs the best, especially in the case of $\zeta_1$ and $\zeta_2$ which have low sensitivity index as enumerated in \Cref{table: SA_wing}.}
    \label{fig: calibration_parameters}
\end{figure}

\Cref{fig: calibration_NRMSE_NIS} compares the performance of \gp~in pblackicting unseen HF data against both the modular and joint variations of KOH's method. As it can be observed, the probabilistic \gp~achieves the lowest NRMSE as well as NIS and is closely followed by its deterministic counterpart. As demonstrated in \Cref{fig: calibration_parameters} we observe a similar trend in estimating the calibration parameters where both versions of \gp~not only provide estimates that are closer to the ground truth values, but also show more robustness to variations in the training data. {\color{black}Comparing the estimates across the four calibration parameters, we notice that all the models are less accurate in the case of $\zeta_1$. We attribute this behavior to identifiability issues and the fact that the underlying functions are almost insensitive to $\zeta_1$ (see the sensitivity analysis in \Cref{table: SA_wing}).}

\subsection{Bayesian Optimization} \label{sec: BO}
BO is a global optimization method that is increasingly used in optimizing black-box and expensive-to-evaluate objective functions. The two main ingblackients of BO are an acquisition function (AF) \cite{turner2021bayesian,frazier2015bayesian,couckuyt2022bayesian,nguyen2023tutorial,brochu2010tutorial, foumani2023multi} and an emulator which iteratively interact while searching for the global optimum. In BO, GPs are dominantly used for emulation \cite{kopsiaftis2019gaussian,binois2022survey,sui2018stagewise,morita2022applying,mcintire2016sparse,rana2017high} since they are easy and fast to train, can effectively learn from small data, and naturally provide pblackiction uncertainties which are needed in the AF. 

\begin{figure}[!t]
    \centering
        \includegraphics[width=1\linewidth]{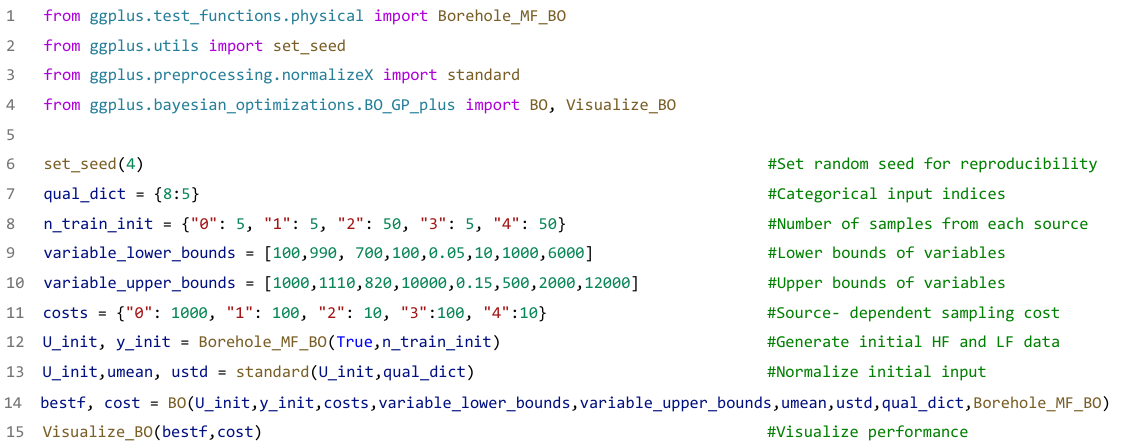}
    \vspace{-.7cm}
    \caption{\textbf{Multi-fidelity Bayesian optimization in \gp:} With just a few lines of code, we solve the \borehole~problem where the goal is to optimize the HF source (denoted by $\texttt{"0"})$ while leveraging four LF sources.}
    \label{fig: BO_screenshot}
\end{figure}

Given the widespread use of GPs in BO, we equip \gp~with a few unique functionalities that streamline optimization of black-box and expensive-to-evaluate objective functions. As schematically demonstrated in \Cref{fig: BO_screenshot}, \gp~enables MFBO with just a few lines of codes; primarily with the \texttt{BO} function which has a few features that distinguish it from other BO packages such as BoTorch. {\color{black}First}, it leverages the emulator described in \Cref{subsec: det-mfmodeling,subsec: prob-mfmodeling,sec: gp_mixed_basis_functions} which provides more accuracy than competing GP modeling packages. {\color{black}Second}, it has the option to tailor the emulation process to BO by integrating MAP with scoring rules. As detailed in \Cref{sec: BO_Appendix}, this integration improves the accuracy of GPs' pblackiction intervals and, in turn, improves the exploration aspect of BO in the context of MF problems \cite{foumani2023effects}. Lastly, the \texttt{BO} function in \gp~has a specialized AF that quantifies the information value of HF and LF data such that they are used primarily for exploitation and exploration, respectively \cite{RN1845}.
At iteration $k$ during MFBO, this AF quantifies the value of a sample as:
\begin{equation} 
    \begin{split}
    \gamma(\boldsymbol{u} ; j)= \begin{cases}\gamma_{LF}(\boldsymbol{u} ; j) / O_j=\frac{\tau_j \phi(\frac{y_j^*-\mu_j}{\tau_j})}{O_j}~& j=[1, \cdots, d s] \hspace{2mm} \& \hspace{2mm} j \neq l \\ \nonumber \\
    \gamma_{HF}(\boldsymbol{u} ; l) / O_l= \frac{\mu_l-y_l^*}{O_l} & j=l \\\nonumber \end{cases} 
    \end{split}
    \label{eq: composite_AF}
\end{equation}
\noindent where $\gamma_{LF}(\cdot)$ and $\gamma_{HF}(\cdot)$ are the AFs of the low- and high-fidelity data sources, $O_j$ is the cost of querying source $j$, $y_j^*$ is the best function value obtained so far from source $j$ (it is assumed that $j=l$ denotes the HF source which we aim to optimize), and $\phi(\cdot)$ denotes the probability density function (PDF) of the standard normal variable. $\tau_j=\tau_j(\boldsymbol{u})$ and $\mu_j=\mu_j(\boldsymbol{u})$ are the standard deviation and mean, respectively, of point $\boldsymbol{u}$ from source $j$.
To proceed to iteration $k+1$, the AF in \Cref{eq: composite_AF} and the emulator are used to solve an auxiliary optimization problem that determines the next point to sample and its corresponding data source:
\begin{equation} 
    \begin{split}
    [\boldsymbol{u}^{(k+1)}, j^{(k+1)}]=\underset{\boldsymbol{u}, j}{\operatorname{argmax}}~\gamma_{\mathrm{MFBO}}(\boldsymbol{u} ; j)
    \end{split}
    \label{eq: auxiliary-opt}
\end{equation}

The \texttt{BO} function in \gp~uses two simple convergence criteria to stop the optimization process: overall data collection costs and the maximum number of iterations without improvement. The former is a rather generic metric but it can result in a considerably high number of iterations in the context of MF problems if an LF source is extremely inexpensive to query. The second metric avoids this issue by putting an upper bound on the maximum number of iterations. These convergence criteria can be easily modified in \gp. 

To demonstrate the effectiveness of \gp~in MFBO we evaluate it against two alternatives: $(1)$ an SF method (denoted by \MEI) that uses the same emulator as \gp~but expected improvement as its AF, $(2)$ \BOT~with STMF-GP and knowledge gradient (KG) as the emulator and AF, respectively (see \Cref{sec: BoTorch} for more details). \BOT~is not applicable to the engineering examples studied below since it cannot handle categorical variables and reports \fbest~values that optimize the learned posterior (rather than the directly sampled data). In the following simulations we denote our methods via \LMGP.

We utilize an analytic (\borehole) and two engineering (\hoip~and \HEA) examples for comparison. We use all $4$ LF sources of \borehole~whose formulation, initialization, and source-dependent sampling costs are provided in \Cref{sec: analytic_example}.
\hoip~has $2$ LF sources and the sampling costs are $40-10-1$ where $40$ is associated with the HF source. We initialize \LMGP~with $15-20-15$ samples and note that the relative accuracy of the LF sources is unknown a priori. 
\HEA~is a $5$-dimensional bi-fidelity problem where we start the optimization with $5-20$ HF-LF samples with sampling costs of $50-10$, respectively (see \Cref{sec: engineering_example} for more details on \hoip~and \HEA). 
In the two engineering problems, \gp~excludes the best compound from the HF dataset and then builds the initial data by randomly sampling from the MF datasets. In all the examples, the maximum number of iterations without improvement is $50$ and a maximum budget $10000$ and $40000$ are used for the engineering and analytic examples, respectively. We report the results for $10$ random initializations. 

\begin{figure}[!t] 
  \centering
    \begin{subfigure}{0.3\textwidth}
    \includegraphics[width=\textwidth]{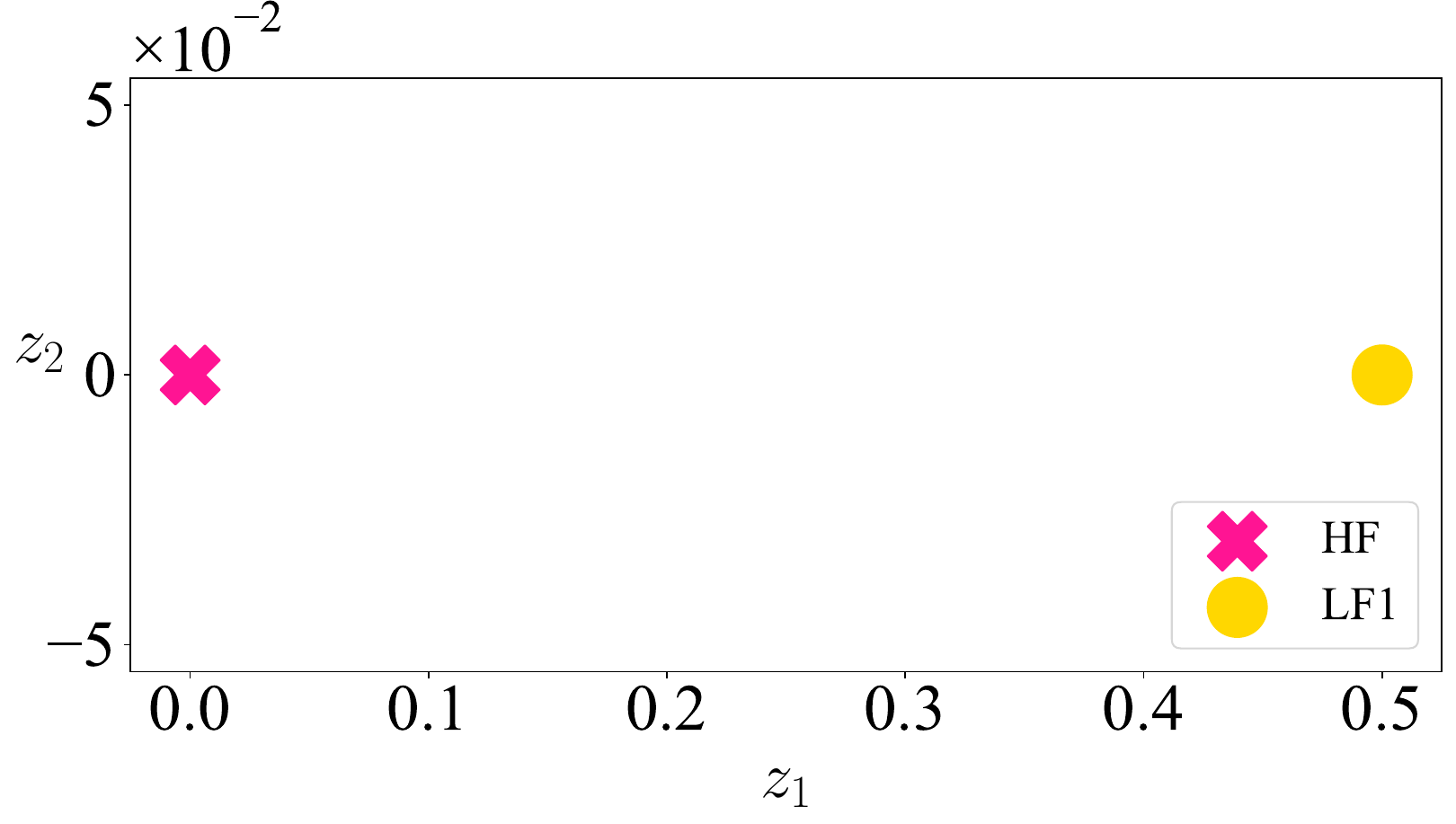}
    \vspace{-7mm}
    \caption{\textbf{\HEA}}
    \label{fig: fidelity_alloy}
    \end{subfigure}
    \begin{subfigure}{0.31\textwidth}
    \includegraphics[width=\textwidth]{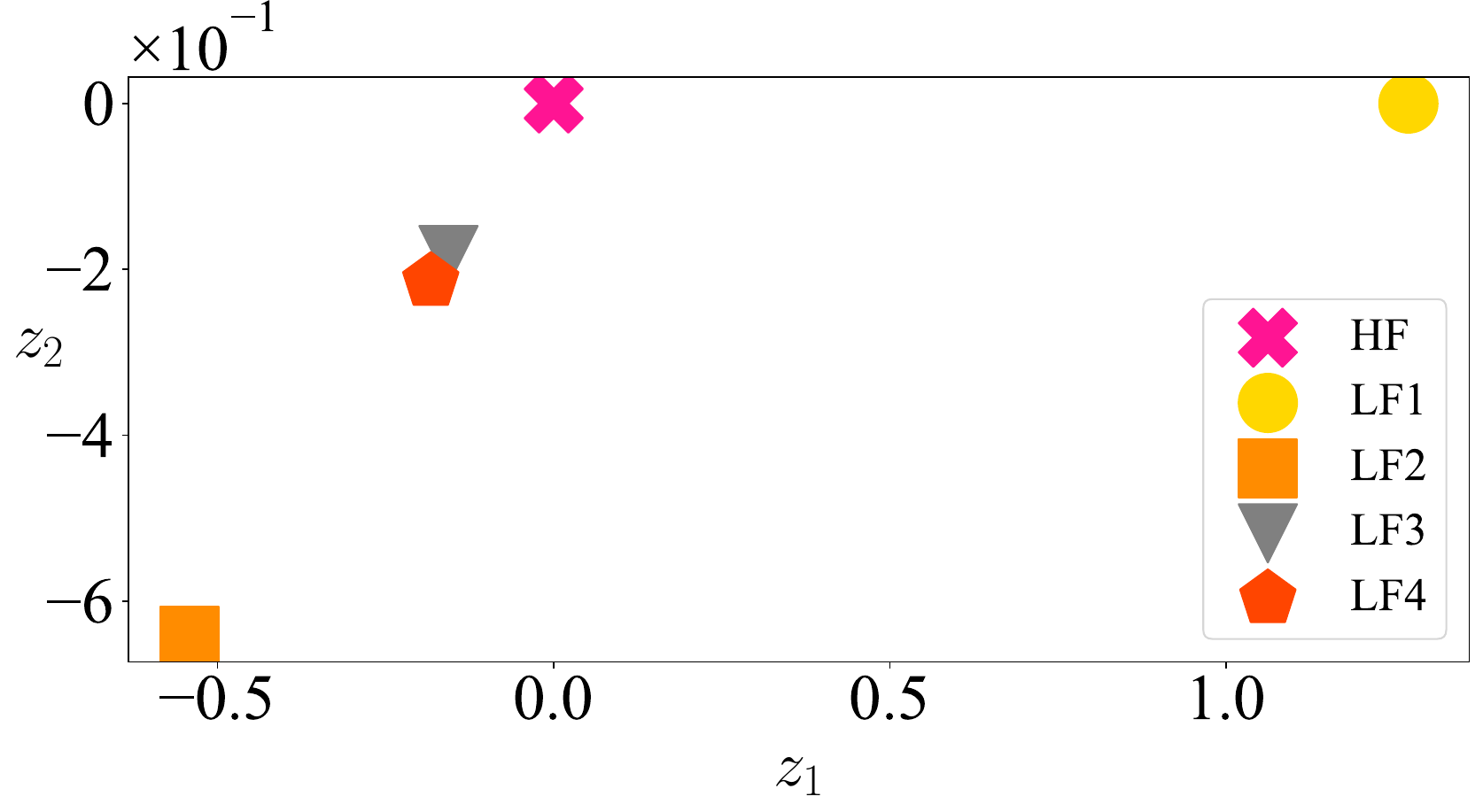}
    \vspace{-7mm}
    \caption{\textbf{\borehole}}
    \label{fig: fidelity_borehole}
    \end{subfigure}
    \begin{subfigure}{0.3\textwidth}
    \includegraphics[width=\textwidth]{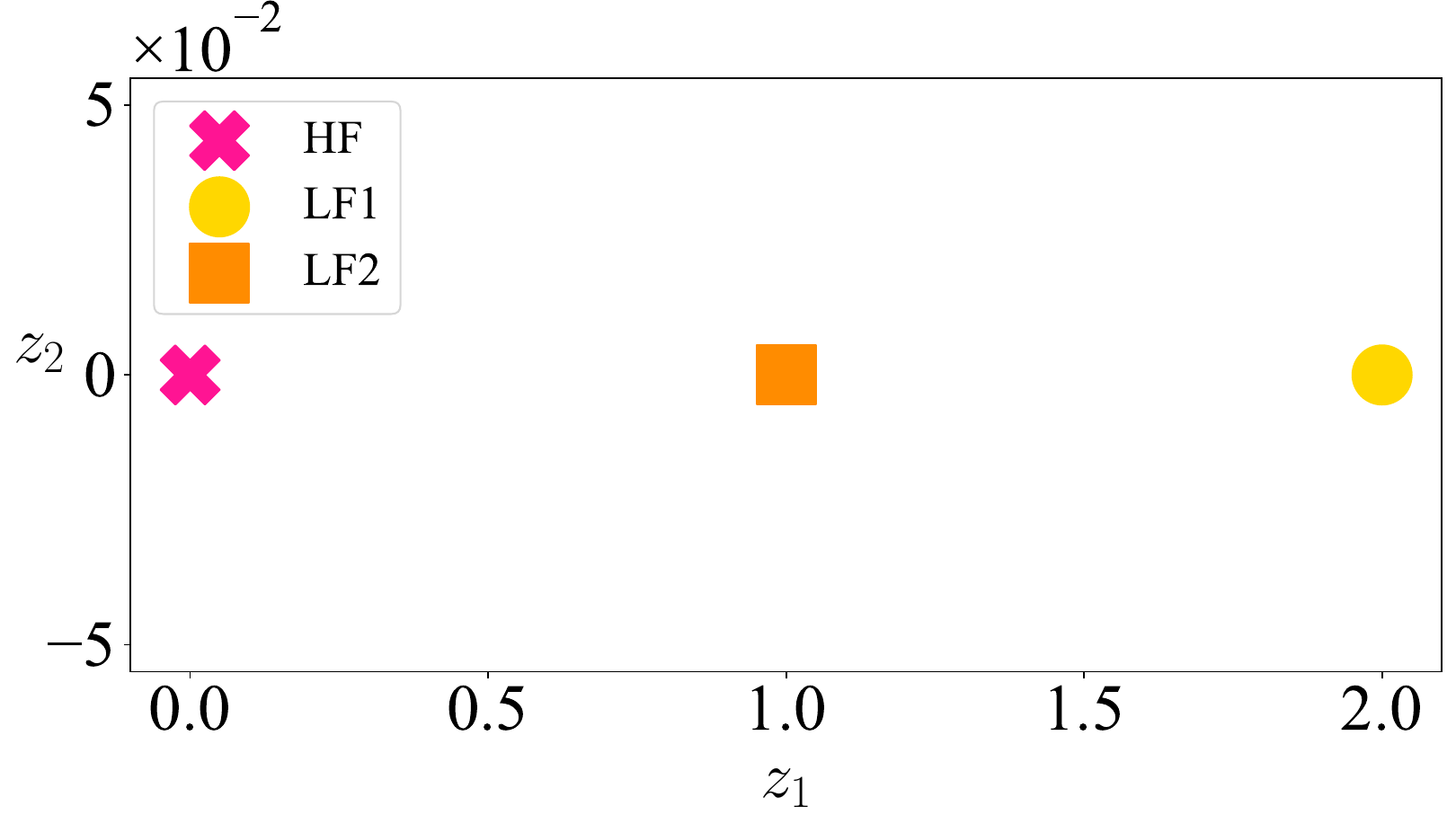}
    \vspace{-7mm}
    \caption{\textbf{\hoip}}
    \label{fig: fidelity_HOIP}
    \end{subfigure}
    \caption{{\color{black} \textbf{Fidelity manifolds in BO problems:} The figures display the fidelity embeddings learned by \gp~based on the initial data in each example. These embeddings quantify the global correlation among the LF sources and the HF source. Based on these plots, while the LF sources in \borehole~and \HEA~ are sufficiently correlated with the HF source, both LF sources of \hoip~reveal low correlation with the HF source. This can cause premature convergence of \gp~due to its second stop condition that limits the number of iterations.}}
  \label{fig: BO_plots_fidelity}
\end{figure}

\begin{figure}[h] 
  \centering
    \begin{subfigure}{0.3\textwidth}
    \includegraphics[width=\textwidth]{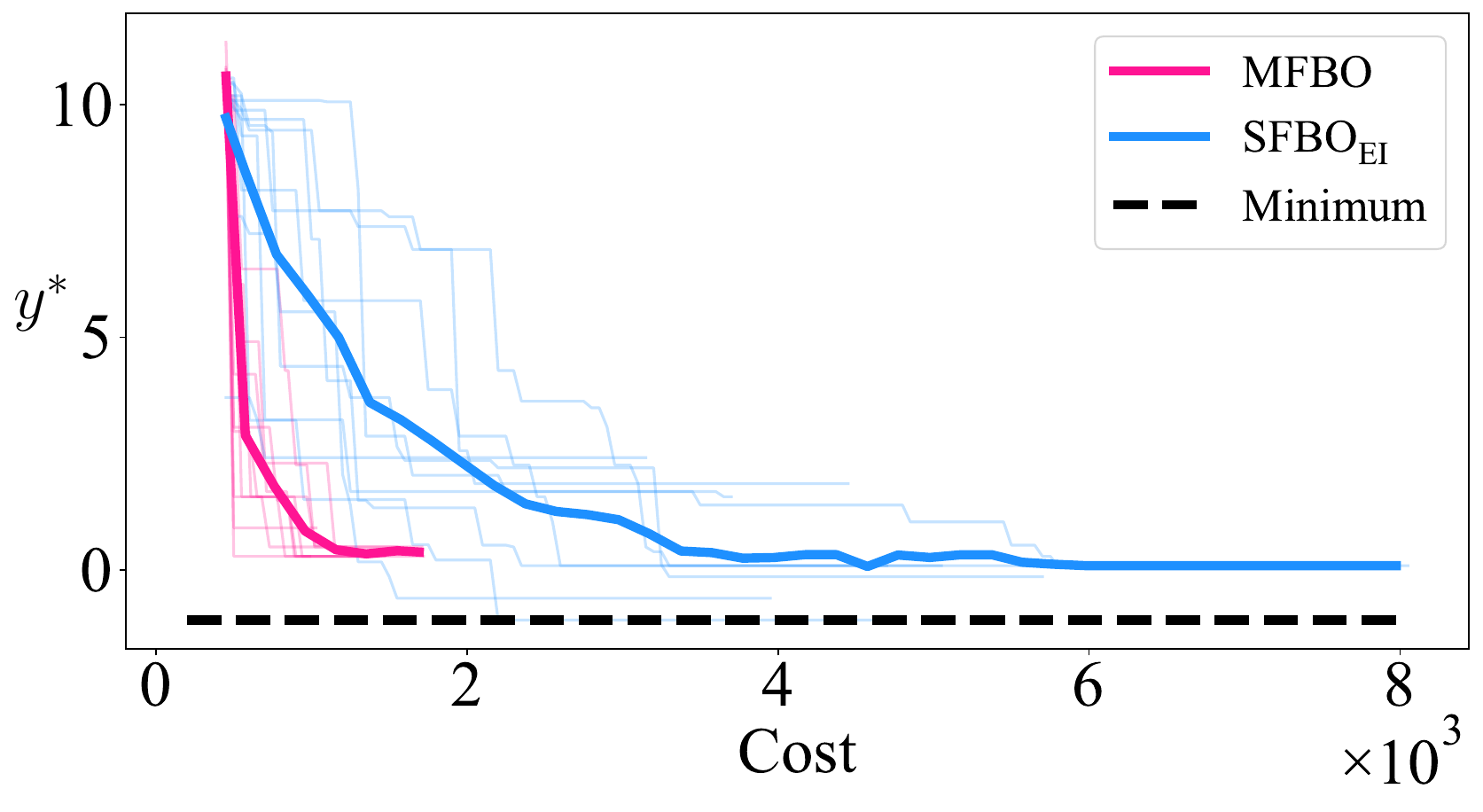}
    \vspace{-7mm}
    \caption{\textbf{\HEA}}
    \label{fig: convergence_alloy}
    \end{subfigure}
    \begin{subfigure}{0.3\textwidth}
    \includegraphics[width=\textwidth]{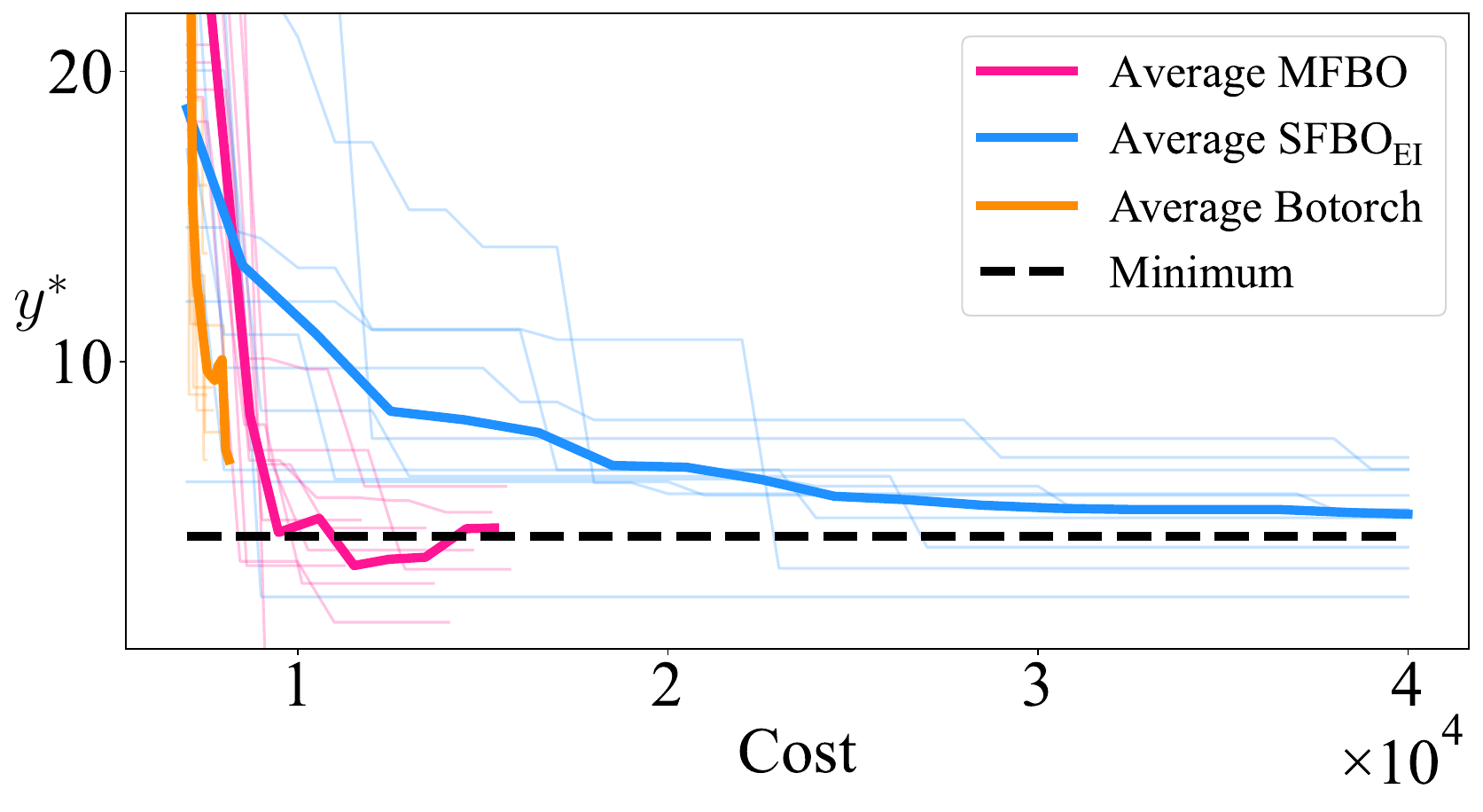}
    \vspace{-7mm}
    \caption{\textbf{\borehole}}
    \label{fig: convergence_borehole}
    \end{subfigure}
    \begin{subfigure}{0.3\textwidth}
    \includegraphics[width=\textwidth]{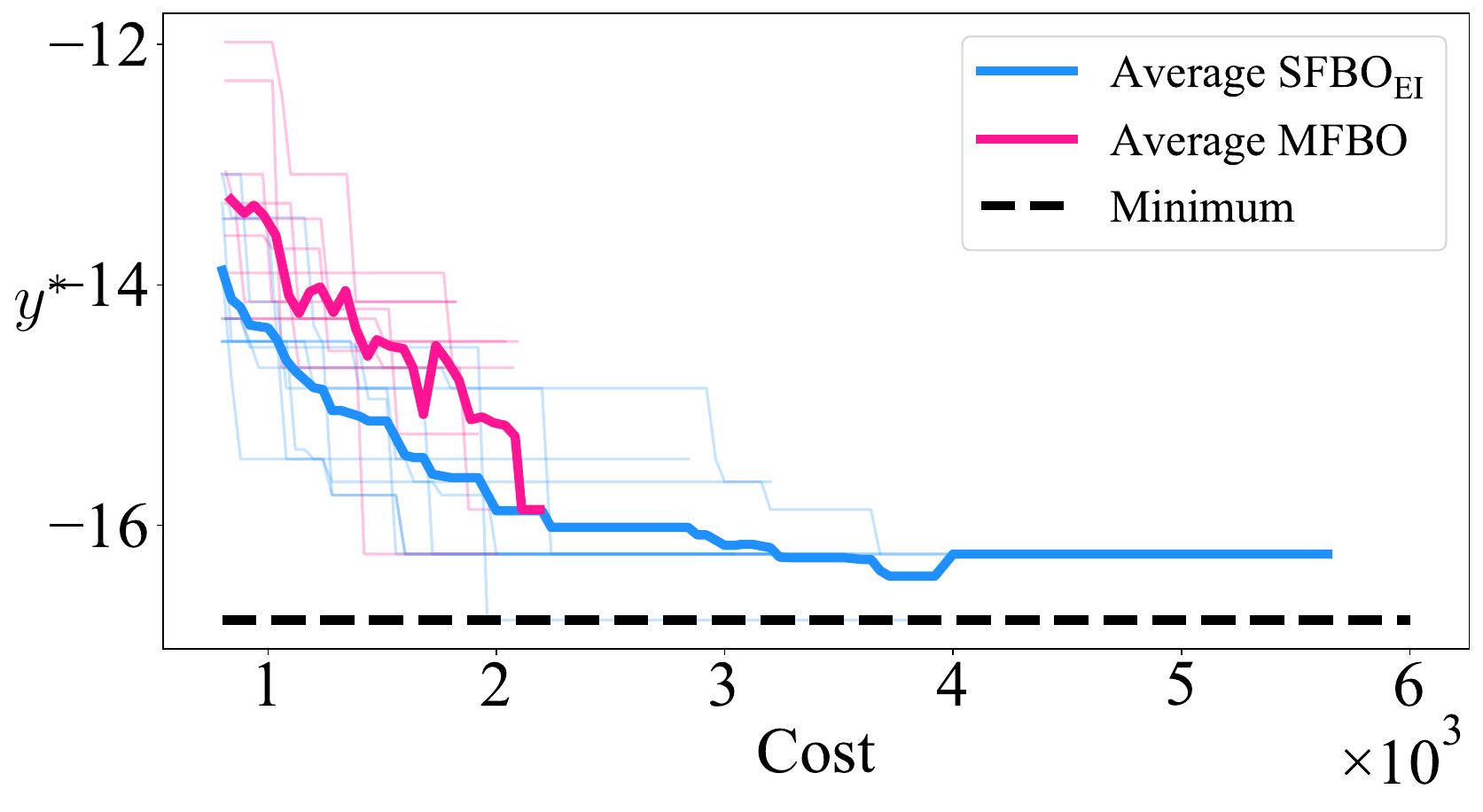}
    \vspace{-7mm}
    \caption{\textbf{\hoip}}
    \label{fig: Convergence_HOIP}
    \end{subfigure}
    \caption{\color{black} \textbf{Convergence histories in BO problems:} The figures show the convergence history of each method versus their accumulated cost. The thin curves illustrate each repetition and the thick curves indicate the average behavior across the $10$ repetitions. Based on these plots, \gp~significantly outperforms \MEI~when LF sources are sufficiently correlated with the HF source (see also \Cref{fig: fidelity_alloy}). However, in scenarios without any correlated sources (\hoip, see \Cref{fig: fidelity_HOIP}), \gp~may converge prematurely due to its second stop condition that limits the number of iterations.}
  \label{fig: BO_plots_convergence}
\end{figure}

The results of our comparison studies are summarized in \Cref{fig: BO_plots_fidelity,fig: BO_plots_convergence} where \Cref{fig: BO_plots_fidelity} illustrates the fidelity embeddings learned for each example by \gp. These embeddings are learnt based on the initial MF data and provide a quantitative metric for assessing the relative accuracy of each LF source with respect to the HF source. Based on these embeddings, while all the fidelity sources are globally correlated in \HEA, two of the LF sources in \hoip~and \borehole~are highly biased and have a limited potential to improve optimization. 
The effects of these correlated/uncorrelated sources on the optimization are illustrated in \Cref{fig: BO_plots_convergence} where the convergence histories are provided by tracking the best HF estimate found by each method (i.e., $y_l^*$ in \Cref{eq: composite_AF}) as a function of the accumulated sampling cost. 
Specifically, as shown in \Cref{fig: convergence_alloy}, the inexpensive correlated LF sources of \HEA~significantly improve the BO where \LMGP~finds the same compound as \MEI~but at a much lower cost.
We note that both \LMGP~and \MEI~converge before finding the smallest HF value primarily because both the HF and LF data are inherently noisy.


As shown in \Cref{fig: convergence_borehole}, the superior performance of \LMGP~is more evident in the \borehole~example which has two informative LF sources (LF3 and LF4 based on \Cref{fig: fidelity_borehole}). In this case, \LMGP~is able to effectively leverage the LF sources in exploring the input space and occasionally samples from the HF source even though the sampling cost associated with it is very high. This infrequent sampling blackuces the overall cost but is necessary for converging to the true minimum. Unlike \LMGP, \BOT~fails in this example since $(1)$ its MF emulator is inaccurate, and $(2)$ its AF incorrectly quantifies the information value to the extent that it cannot find an HF candidate whose value warrants its high sampling cost. As expected, \MEI~converges to a value that is quite close to the minimum but at a much higher cost than \LMGP.

\begin{figure}[!t]
    \centering
    \includegraphics[width=0.8\linewidth]{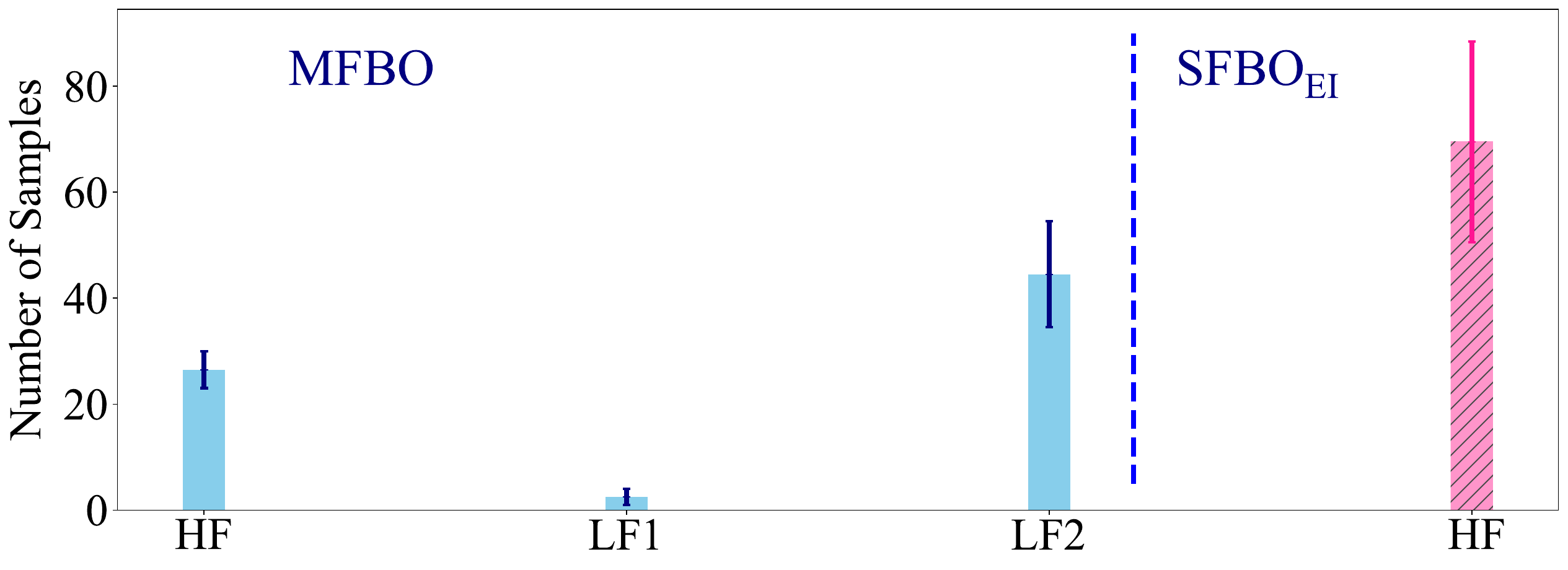}
    \caption{\textbf{Sampling history in \hoip:} We report the number of samples taken from each source in \LMGP~and \MEI~across the $10$ repetitions. While \LMGP~converges with $25$ HF samples on average, \MEI~requires more than $70$ HF samples.}
    \label{fig: HOIP_Sampling}
\end{figure}

\hoip~is an example where the interpretable diagnostic tools of \gp~prove useful. Specifically, the fidelity embedding in \Cref{fig: fidelity_HOIP} indicates that both LF sources are highly biased compablack to the HF one and hence optimizing the latter may not benefit from sampling from the LF sources especially if the cost ratios are unbalanced. By investigating the optimization histories in \Cref{fig: Convergence_HOIP} we realize that \LMGP~is unable to provide the same improvements as in the other two examples. 
Specifically, \LMGP~finds an optimum that is quite close to the one found by \MEI~($-15.87$ vs $-16.2$) but it does so at a much lower cost ($2000$ vs $6000$) since it primarily samples from the LF sources, see \Cref{fig: HOIP_Sampling}. The performance of \LMGP~can be improved in such applications with highly biased LF sources by initializing the optimization process with more HF data.

    \section {Conclusions and Future Directions} \label{sec: conclusion}
In this paper, we introduce \gp~which is an open-source Python library that systematically integrates nonlinear manifold learning techniques with GPs. As demonstrated with the examples in \Cref{sec: functionalities}, this integration provides a unified platform for studying a broad range of problems ranging from MF emulation to probabilistic calibration and BO. 

In this paper, we primarily focused on applications where data (especially high-fidelity data) is scarce and hence probabilistic modeling has an edge over a deterministic one. We achieve this probabilistic nature in \gp~via variational approaches and plan to extend it in our future works based on MCMC. We anticipate this extension to more accurately quantify uncertainties but with a noticeable increase in computational costs. {\color{black} Additionally, all the problems analyzed in this paper were single-output. Currently, \gp~manages multi-output problems by concatenating the input space with categorical features to distinguish among the outputs. We aim to add more options for handling multi-output problems in future versions of \gp.}
Another interesting future direction is integrating our contributions with techniques that extend GPs to solve partial differential equations, classification tasks, or big data problems. 

A unique advantage of \gp~over other GP modeling libraries is the interpretability of its learnt latent spaces. One example is the fidelity embedding that \gp~constructs for MF data which provides a visualizable metric that quantifies the relative similarity of the data sources. Based on the formulations in \Cref{subsec: det-mfmodeling,subsec: prob-mfmodeling}, this fidelity embedding quantifies the global correlations among the data sources since it is not a function of the input features $\xb$ or $\tb$. We plan to extend this approach to learn local correlations but note that this extension will rely on larger training datasets since the requiblack embedding functions will have more parameters. 
{\color{black}With a somewhat similar approach, we believe the calibration scheme introduced in \Cref{subsec: prob-calibration} can be generalized to cases where not all the unknown parameters are shablack across the computer models. This generalization is tightly connected to the so-called non-identifiability issues which are both worthy of in-depth future investigations.}

\section*{Acknowledgments}
We appreciate the support from NASA’s Space Technology Research Grants Program (grant number $80NSSC21K1809$), Office of the Naval Research (grant number $N000142312485$), National Science Foundation (grant number $2238038$), and the UC National Laboratory Fees Research Program of the University of California (Grant Number $L22CR4520$).

    \pagebreak

\begin{appendices}

\setcounter{equation}{0}
\renewcommand{\theequation}{\thesection-\arabic{equation}}

\section{Details on Data and Benchmark Problems} \label{sec: appendix-Data}
In this section, we provide details on the engineering (\Cref{sec: engineering_example}) and analytic (\Cref{sec: analytic_example}) examples that are used in \Cref{sec: functionalities} of the paper.

\subsection{Engineering Examples}  \label{sec: engineering_example}

Hybrid organic–inorganic perovskite (\hoip) is a material design problem where the goal is to identify the composition with the smallest inter-molecular binding energy \cite{egger2016hybrid}. The dataset used here is generated via three distinct sources that simulate the band gap property of \hoip~as a function of composition based on the density functional theory (DFT). These compositions are characterized via three categorical variables that have $10$, $3$, and $16$ levels (i.e., there are $480$ unique compositions in total). The major differences between the datasets generated by the three sources are their fidelity (or accuracy) and size. Specifically, dataset generated by Source $1$ is the most accurate among the three and contains $480$ samples while the datasets generated by Source $2$ and Source $3$ contain $179$ and $240$ samples, respectively, and have lower levels of accuracy compablack to Source $1$. In \Cref{sec: emulation_functionality} we exclusively use the HF data but use all three sources with the corresponding costs of $40$, $10$, and $1$ in \Cref{sec: BO}.

High entropy alloy (\HEA) is another alloy design problem with the goal of finding the compound with the lowest thermal expansion coefficient \cite{rao2022machine}. This is a $5$-dimensional example where the features show the percentage of each element ($Fe, Ni, Co, Cr, V, Cu$) in the alloy. It has one high- and one low-fidelity data sources whose corresponding costs are $50$ and $10$, respectively. Both datasets have $700$ samples and are single-response.

The \DNS~dataset \cite{deng2023data} aims to enhance the speed of multiscale damage simulations for cast aluminum alloys. The acceleration process involves replacing the direct numerical simulations (DNS) at the microscale with blackuced-order models (ROMs). The ROMs used in \DNS~have three distinct cluster counts: $(800, 1600, 3200)$, where a greater cluster count offers outcomes closer to DNS at a higher computational cost. \DNS~is a $6$-dimensional problem where the features are all quantitative and characterize pore volume fraction, pore count, pore aspect ratio, average distance between neighboring pores, evolutionary rate parameter, and the critical effective plastic strain, with the latter two determining the material's damage behavior under stress. The number of samples from the highest to the lowest fidelities are $70, 110, 170, \text{and}~250$, respectively.

\nta~is the problem of designing a nanolaminate ternary alloy that is used in applications such as high-temperature structural materials \cite{cover2009comprehensive}. These alloys have compositions of the form $M_2AX$ where $M$ is an early transition metal, $A$ is a main group element, and $X$ is either carbon or nitrogen. \nta~is $3-$dimensional with just categorical features which have $10$, $12$, and $2$ levels. The dataset has $224$ samples and the response is the bulk modulus of the alloy.

\subsection{Analytic Examples} \label{sec: analytic_example}
The mathematical formulations are provided in \Cref{table: analytic-formulation} which also includes details on initializations and source-dependent sampling costs used in \Cref{sec: BO} for BO. \Cref{table: analytic-formulation} also shows the error of each LF source with respect to the corresponding HF source based on NRMSE in \Cref{eq: nrmse}. That is:
\begin{equation} 
    NRMSE = \sqrt{\frac{(\boldsymbol{y}_l - \boldsymbol{y}_h)^T(\boldsymbol{y}_l - \boldsymbol{y}_h)}{10000 \times var(\boldsymbol{y}_h)}}
    \label{eq: nrmse-appendix}
\end{equation}
\noindent where $\boldsymbol{y}_l$ and $\boldsymbol{y}_h$ are vectors of size $10000 \times 1$ that store random samples taken from the low and high-fidelity sources, respectively. 

\begin{table}[!b]
    \setlength{\extrarowheight}{0pt}
    \setlength{\tabcolsep}{5pt} 
    {\renewcommand{\arraystretch}{2}
    \centering
    \begin{tabular}{
        |c
        |c
        |c
        |c
        |c
        |c
        |c|
    }
        \hline
        \textbf{Name} &
        \textbf{Source ID} &
        \textbf{Formulation} &
        \textbf{n}  &
        \textbf{NRMSE} &
        \textbf{Cost} &
        \textbf{Noise} \\
        \hline
        \multirow{5}{*}{\textbf{\borehole}} &
        HF &
        $\frac{2 \pi T_{u}(H_u-H_l)}{\ln (\frac{r}{r_w})(1+\frac{2 L T_u}{\ln (\frac{r}{r w}) r_w^2 k_w}+\frac{T_{u}}{T_l})}$ &
        $5$ &
        $-$ &
        $1000$&
        $2$    \\
        \hhline{~|------|}
        & LF1 &
        $\frac{2 \pi  T_{\mathrm{u}} (H_{\mathrm{u}}-0.8  H_l)} {\ln (\frac{r}{r_w})(1+\frac{1 L T_{\mathrm{u}}}{\ln (\frac{r}{r_w}) r_w^2 k_w}+\frac{T_{u}}{T_l})}$ &
        $5$ &
        $4.40$ &
        $100$ &
        $-$ \\
        \hhline{~|------|}
        & LF2 &
        $\frac{2 \pi T_u(H_u- H_l)}{\ln (\frac{r}{r_w})(1+\frac{8 L T_u}{\ln (\frac{r}{r w}) r_w^2 k_w}+0.75 \frac{T_{u}}{T_l})}$ &
        $50$ &
        $1.54$ &
        $10$ &
        $-$ \\
        \hhline{~|------|}
        & LF3 &
        $\frac{2 \pi T_{\mathrm{u}}(1.09 H_u-H_l)}{\ln (\frac{4 r}{r_w})(1+\frac{3 L T_{u}}{\ln (\frac{r}{r_w}) r_w^2 k_w}+\frac{T_u}{T_l})}$ &
        $5$ &
        $1.30$ &
        $100$ &
        $-$ \\
        \hhline{~|------|}
        & LF4 &
        \makecell{%
          $\frac{2 \pi T_{\mathrm{u}}(1.05 H_u-H_l)}{\ln (\frac{2 r}{r_w})}$ \\ [0.5ex]
          $\times \left(1+\frac{3 L T_{i u}}{\ln (\frac{r}{\tau w}) r_w^2 k_{\mathrm{W}}}+ \frac{T_{\mathrm{u}}}{T_l}\right)$
        } &
        $50$ &
        $1.3$ &
        $10$ &
        $-$\\
        \hline
        \multirow{4}{*}{\textbf{\wing}} &
        HF &
        \makecell{%
          $0.036 s_w^{0.758} w_{f w}^{0.0035}(\frac{A}{\cos ^2(\Lambda)})^{0.6} \times q^{0.006}$ \\ [0.5ex]
          $ \times \lambda^{0.04}(\frac{100 t_c}{\cos (\Lambda)})^{-0.3}(N_z W_{d g})^{0.49}+s_w w_p$
        } &
        $10$ &
        $-$ &
        $-$ &
        $1$ \\
        \hhline{~|------|}
        & LF1 &
        \makecell{%
          $0.036 s_w^{0.758} w_{f w}^{0.0035}(\frac{A}{\cos ^2(\Lambda)})^{0.6} \times q^{0.006}$ \\ [0.5ex]
          $ \times \lambda^{0.04}(\frac{100 t_c}{\cos (\Lambda)})^{-0.3}(N_z W_{d g})^{0.49}+w_p$
        } &
        $20$ &
        $0.19$ &
        $-$ &
        $1$\\
        \hhline{~|------|}
        & LF2 &
        \makecell{%
          $0.036 s_w^{0.8} w_{f w}^{0.0035}(\frac{A}{\cos ^2(\Lambda)})^{0.6} \times q^{0.006}$ \\ [0.5ex]
          $ \times \lambda^{0.04}(\frac{100 t_c}{\cos (\Lambda)})^{-0.3}(N_z W_{d g})^{0.49}+w_p$
        } &
        $20$ &
        $1.14$ &
        $-$ &
        $1$ \\
        \hhline{~|------|}
        & LF3 &
        \makecell{%
          $0.036 s_w^{0.9} w_{f w}^{0.0035}(\frac{A}{\cos ^2(\Lambda)})^{0.6} \times q^{0.006}$ \\ [0.5ex]
          $ \times \lambda^{0.04}(\frac{100 t_c}{\cos (\Lambda)})^{-0.3}(N_z W_{d g})^{0.49}$
        } &
        $20$ &
        $5.75$ &
        $-$ &
        $1$\\
        \hline
        \multirow{2}{*}{\textbf{\sine}} &
        HF &
        \makecell{%
          $2 \sin{(x)}$ 
        } &
        $4$ &
        $-$ &
        $-$ &
        $1$ \\     
        \hhline{~|------|}
        & LF1 &
        \makecell{%
          $2 \sin{(x)}+ 0.3x^2 -0.7x+1$
        } &
        $20$ &
        $0.11$ &
        $-$ &
        $1$\\
        \hline
    \end{tabular} 
    }
    \caption{\textbf{Analytic Examples}: The examples have a diverse degree of dimensionality, number of sources, and complexity. $n$ denotes the number of initial samples used in BO and the NRMSE of an LF source is calculated by comparing its output to that of the HF source at $10000$ random points, see \Cref{eq: nrmse-appendix}.}
    \label{table: analytic-formulation}
\end{table}

\borehole~is an $8$-dimensional single-response example whose input space only has quantitative features. Only the Hf source is noisy in this case and there are $4$ LF sources two of which are highly biased based on the NRMSEs. Note that the relative accuracy levels of the LF sources are calculated based on large data in \Cref{table: analytic-formulation} but interestingly these are consistent with the latent distances learnt by \gp~based on small data, see \Cref{fig: fidelity_borehole}.

\boreholeM~refers to the \borehole~example whose first and sixth features are converted into categorical variables with $5$ distinct levels each. To achieve this, we first sample $5$ values within the upper and lower bounds of these features. Subsequently, we randomly assign these sampled values to their corresponding variables and calculate the outputs using the formulation outlined in \Cref{table: analytic-formulation} for \borehole. Then, the categorical conversion is done by sorting the sampled values for the first and sixth features and substituting each feature's sampled point with its index in the sorted list.

\begin{figure}[!h]
    \centering
    \begin{subfigure}{1\textwidth}
        \centering
        \includegraphics[width=1\linewidth]{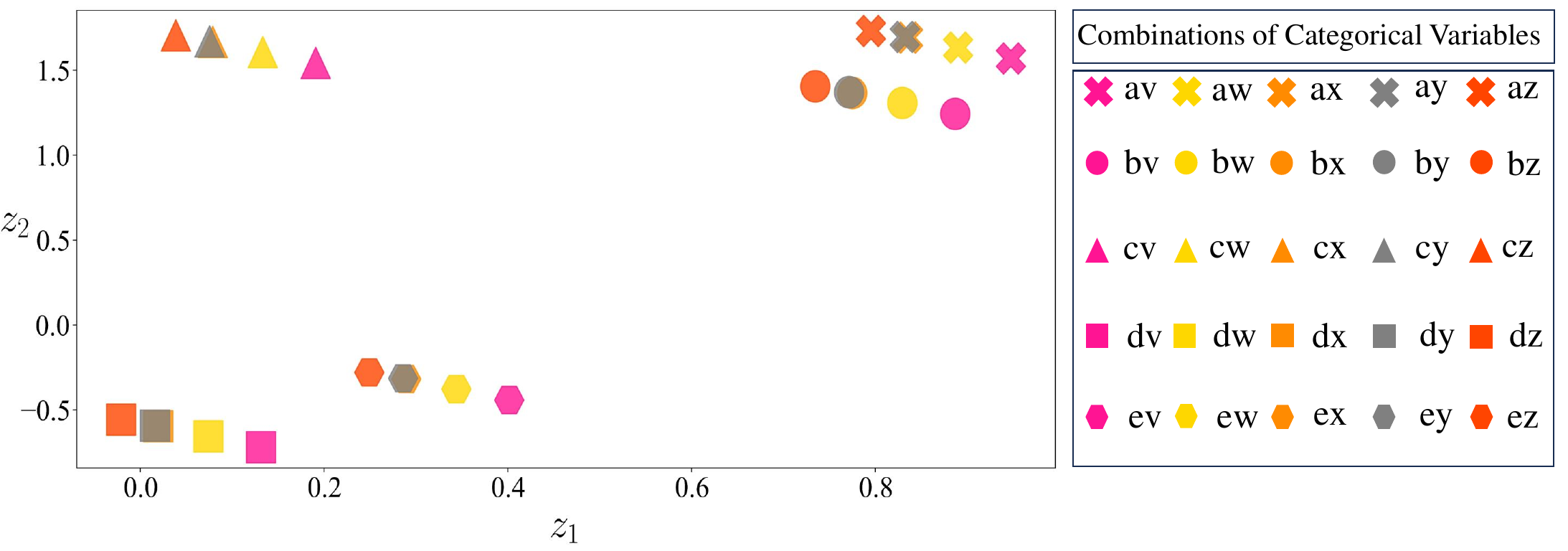}
        \caption{Colors correspond to the unique combinations of levels of the two categorical variables.}
        \label{fig: encoded_cat_borehole}
    \end{subfigure}%
    \newline
    \begin{subfigure}{1\textwidth}
        \centering
        \includegraphics[width=1\linewidth]{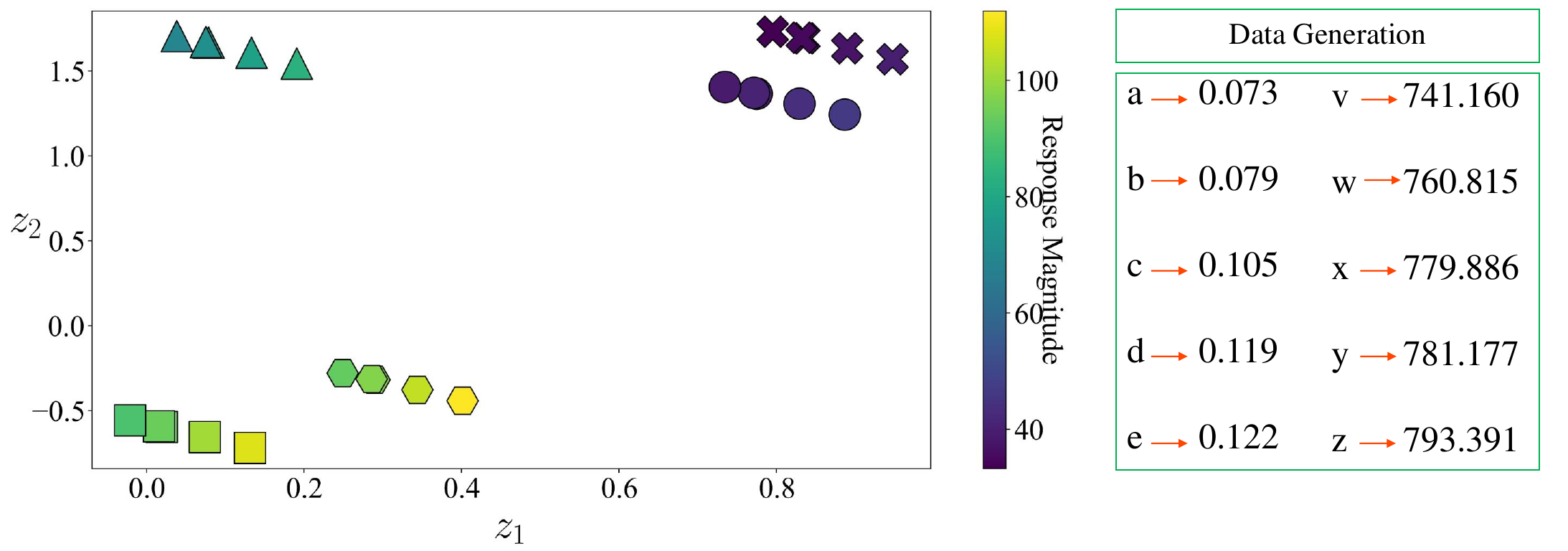}
        \caption{Colors are based on the response magnitude.}
        \label{fig: color_coded_borehole}
    \end{subfigure}%
    \caption{\textbf{Latent positions learned via \gp~for \boreholeM:} We use the numerical values shown in \textbf{(b)} to generate $400$ training samples via \borehole. These values are then replaced with the corresponding categorical features before \gp~is used for emulation. Upon training, we visualize the embedding learnt by \gp~while color-coding it based on either the variable combinations or response magnitude.}
    \label{fig: borehole_cat_encoded}
\end{figure}

To demonstrate the interpretability of embeddings learnt by \gp, we now use \boreholeM~ to generate a training dataset of size $400$ and then fit an emulator to it. Upon training, we visualize the latent points learnt for the two categorical variables and color-code them based on either the combination or response magnitude, see \Cref{fig: encoded_cat_borehole,fig: color_coded_borehole}, respectively. As it can be observed in these figures, the learnt embeddings preserve the underlying numerical relations even though \gp~does not have access to the numerical values used in data generation. For instance, all the variability in the latent space is in two directions with is in line with the fact that all combinations of the categorical data can be quantified via the two underlying numerical features. We also observe that close-by (distant) latent points have similar (different) response values which increases as we move from the top-right corner to the bottom left corner of the latent space, see \Cref{fig: color_coded_borehole}. These embeddings also indicate variable importance. For instance, based on \Cref{fig: encoded_cat_borehole} we observe that changing the levels of the first categorical variable from \texttt{"a"} to \texttt{"e"} results in larger latent movements than changing the levels of the second categorical variable. Such a behavior indicates the the underlying function is more sensitive to the former variable and we validate this argument in \Cref{sec: Sobole} by conducting sensitivity analyses. 


\wing~is a $10$-dimensional single-response example with one HF and $3$ LF sources. Based on the NRMSE values shown in \Cref{table: analytic-formulation}, the source ID, true fidelity level, and sampling costs follow the same trend (unlike \borehole). For instance, the first LF source is the most accurate and most expensive among all the LF sources. Additionally, a Gaussian noise with a standard deviation of $1$ is added to all the fidelity sources.

\sine~is a $1$-dimensional bi-fidelity problem where there are high correlations between the two sources as indicated by the NRMSE value in \Cref{table: analytic-formulation}. We use a standard Gaussian noise to corrupt the data from both sources.

\subsection{Calibration Examples}
\wing, \borehole~and \beam~are used in \Cref{sec: Calibration_Functionality} to illustrate the performance of \gp~in calibration. Their formulation and details are provided in \Cref{table: analytic_calibration}. 
As explained in \Cref{sec: Calibration_Functionality}, there are $2$ and $4$ calibration parameters in \borehole~and \wing, respectively. Also, \beam~is a bi-fidelity $5$-dimensional problem \cite{marelli2014uqlab} whose features are a constant distributed load applied to the beam ($p=12000N/m$), width ($b=0.15m$), height ($h=0.3m$), and the length of the beam ($L=5~m$). The last feature is the unknown material Young’s modulus ($\zeta$) which we aim to estimate. The analytic formulation and more details regarding \beam~is also presented in \Cref{table: analytic_calibration}. 

\begin{table}[!ht]
    \setlength{\extrarowheight}{0pt}
    \setlength{\tabcolsep}{5pt} 
    {\renewcommand{\arraystretch}{2}
    \centering
    \begin{tabular}{
        |c
        |c
        |c
        |c
        |c
        |c|
    }
        \hline
        \textbf{Name} &
        \textbf{Source ID} &
        \textbf{Formulation} &
        \textbf{n}  &
        \textbf{NRMSE} &
        \textbf{Noise} \\
        \hline
        \multirow{3}{*}{\textbf{\borehole}} &
        HF &
        $\frac{2 \pi T_{u}(H_u-H_l)}{\ln (\frac{r}{r_w})(1+\frac{2\times 1500 T_u}{\ln (\frac{r}{r w}) r_w^2 k_w}+\frac{T_{u}}{250})}$ &
        $20$ &
        $-$ &
        $2$    \\
        \hhline{~|-----|}
        & LF1 &
        $\frac{2 \pi  T_{\mathrm{u}} (H_{\mathrm{u}}-0.8 H_l)} {\ln (\frac{r}{r_w})(1+\frac{1 \zeta_1 T_{\mathrm{u}}}{\ln (\frac{r}{r_w}) r_w^2 k_w}+\frac{T_{u}}{\zeta_2})}$ &
        $100$ &
        $4.40$ &
        $-$ \\
        \hhline{~|-----|}
        & LF2 &
        $\frac{2 \pi T_u(H_u- H_l)}{\ln (\frac{r}{r_w})(1+\frac{8 \zeta_1 T_u}{\ln (\frac{r}{r w}) r_w^2 k_w}+0.75 \frac{T_{u}}{\zeta_2})}$ &
        $100$ &
        $1.54$ &
        $-$ \\
        \hline
        \multirow{4}{*}{\textbf{\wing}} &
        HF &
        \makecell{%
          $0.036 s_w^{0.758} w_{f w}^{0.0035}(\frac{A}{\cos ^2(\Lambda)})^{0.6} \times {40}^{0.006}$ \\ [0.5ex]
          $ \times {0.85}^{0.04}(\frac{100 \times 0.17}{\cos (\Lambda)})^{-0.3}(3 W_{d g})^{0.49}+s_w w_p$
        } &
        $25$ &
        $-$ &
        $1$ \\
        \hhline{~|-----|}
        & LF1 &
        \makecell{%
          $0.036 s_w^{0.758} w_{f w}^{0.0035}(\frac{A}{\cos ^2(\Lambda)})^{0.6} \times {\zeta_1}^{0.006}$ \\ [0.5ex]
          $ \times {\zeta_2}^{0.04}(\frac{100 \zeta_3}{\cos (\Lambda)})^{-0.3}(\zeta_4 W_{d g})^{0.49}+w_p$
        } &
        $40$ &
        $0.19$ &
        $1$\\
        \hhline{~|-----|}
        & LF2 &
        \makecell{%
          $0.036 s_w^{0.8} w_{f w}^{0.0035}(\frac{A}{\cos ^2(\Lambda)})^{0.6} \times {\zeta_1}^{0.006}$ \\ [0.5ex]
          $ \times {\zeta_2}^{0.04}(\frac{100 \zeta_3}{\cos (\Lambda)})^{-0.3}(\zeta_4 W_{d g})^{0.49}+w_p$
        } &
        $50$ &
        $1.14$ &
        $1$ \\
        \hhline{~|-----|}
        & LF3 &
        \makecell{%
          $0.036 s_w^{0.9} w_{f w}^{0.0035}(\frac{A}{\cos ^2(\Lambda)})^{0.6} \times {\zeta_1}^{0.006}$ \\ [0.5ex]
          $ \times {\zeta_2}^{0.04}(\frac{100 \zeta_3}{\cos (\Lambda)})^{-0.3}(\zeta_4 W_{d g})^{0.49}$
        } &
        $60$ &
        $5.75$ &
        $1$\\
        \hline
        \multirow{2}{*}{\makecell{\textbf{\texttt{Beam}} \\ \textbf{\texttt{Deflection}}}}&
        HF &
        \makecell{%
          $\frac{5}{32}\frac{pL^4}{3 \times{10^{10}} bh^3}$
        } &
        $1$ &
        $-$ &
        $0.05$ \\
        \hhline{~|-----|}
        & LF1 &
        \makecell{%
          $\frac{5}{32}\frac{pL^4}{\zeta_1 bh^3}$
        } &
        $200$ &
        $-$ &
        $0$\\
        \hline
    \end{tabular} 
    }
    \caption{\textbf{Analytic examples used for calibration}: $n$ denotes the number of samples used in \Cref{sec: Calibration_Functionality} and the NRMSE of an LF source is calculated by comparing its output to that of the HF source at $10000$ random points, see \Cref{eq: nrmse-appendix}.}
    \label{table: analytic_calibration}
\end{table}

\section{Sensitivity Analysis} \label{sec: Sobole}
Sobol sensitivity analysis is a global variance-based method used for quantifying each input’s main and total contribution to the output variance \cite{saltelli2010variance}. While main-order Sobol indices (SIs) reveal the individual contributions of input variables, total-order indices capture both the individual and interaction effects of inputs on the output. We highlight that while Sobol indices are typically applied to quantitative features, we extend the idea to qualitative features in \gp~by sampling random quantitative values and mapping them to the unique levels of a categorical variable. This functionality is accessible with the $\texttt{model.Sobol()}$ command in \gp.

The $\omegab$ parameter in GP-based emulation (see \Cref{sec: gp-background} for details) plays a similar role to the Sobol indices in that it reveals the sensitivity of the emulator to the quantitative features where a smaller $\omega_i$ value indicates that the output is less sensitive to the $i^{th}$ feature. Since $\omegab$ are not defined for categorical inputs, we propose to measure their sensitivities based on the average distance among the learnt encoded points. Specifically, we encode the $l_i$ categories of variable $t_i$ to $l_i$ latent points whose average inter-distances is then used to measure the sensitivity of the output to $t_i$. We denote this metric by $S_{cat}$ and highlight that we calculate it by endowing each categorical variable with its own latent space (this is in contrast to the examples in \Cref{sec: emulation_functionality,sec: BO} where we encode all the combinations of all categorical variables into a single latent space). We adopt this approach primarily for ease of implementation and increasing interpretability.

To demonstrate the interpretability of a GP's parameters, we compare them against main and total SIs in \borehole, \boreholeM, and \wing~ problems. To this end, we train three GPs via \gp~to emulate these functions (we use sufficient samples to ensure the trained GPs accurately learn the underlying functions) and then compare the estimated parameters of these GPs to main and total SIs. The results are enumerated in \Cref{table: SA_borehole,table: SA_borehole_mixed,table: SA_wing} and indicate that there is a good agreement between the two different metrics, that is, important features that have large SIs also have large $10^{\omega_i}$ or $S_{cat}$. For instance, $r_w, H_u, H_l, \text{and} L$ are the most sensitive features of \borehole~and \boreholeM~as indicated by both Sobol and \gp.
We note that reported metrics in \Cref{table: SA_borehole} and \Cref{table: SA_borehole_mixed} are slightly different since the latter is affected by categorization of two of its inputs. 


\begin{table}[!h]
    \centering
    \begin{tabular}{c|cccccccc}
    \setlength{\tabcolsep}{1pt}
        & \multicolumn{3}{c}{} & Features &\\ \hline
        \multicolumn{1}{c|}{Metric} &\multicolumn{1}{c}{$r_w$}& \multicolumn{1}{c}{$r$} & \multicolumn{1}{c}{$T_u$} & \multicolumn{1}{c}{$H_u$}  & \multicolumn{1}{c}{$T_l$} & \multicolumn{1}{c}{$H_l$}&\multicolumn{1}{c}{$L$} & \multicolumn{1}{c}{$k_w$}   \\ \hline
        \multicolumn{1}{c|}{Main SI}  & $0.830$       & $1.57e{-7}$      & $2.34e{-8}$ & $0.042$   & $2.09e{-6}$  & $0.041$ & $0.039$ & $0.009$      \\
        \multicolumn{1}{c|}{Total SI}  & $0.868$       & $3.65e{-6}$      & $4.07e{-8}$ & $0.054$   & $1.25e{-5}$  & $0.054$ & $0.051$ & $0.013$       \\
        \multicolumn{1}{c|}{$10^{\omega_{i}}$ } & $0.125$ & $1.00e{-4} $      & $2.45e{-7}$      & $0.009$ & $2.95e{-5}$ & $0.013$ & $0.029$   & $0.003$        \\
        \hline    
    \end{tabular}
    \caption{\textbf{Sensitivity analysis of \borehole:} Sensitivity analysis of \borehole~using Sobol indices and \gp~emulator. Based on \Cref{eq: all-kernels-GP_Plus}. In \Cref{sec: Calibration_Functionality}, $\zetab = \brackets{T_l, L}^T$ are treated as calibration parameters.}
    \label{table: SA_borehole}
\end{table}

\begin{table}[!h]
    \centering
    \begin{tabular}{c|cccccccc}
    \setlength{\tabcolsep}{1pt}
        & \multicolumn{3}{c}{} & Features &\\ \hline
        \multicolumn{1}{c|}{Metric} &\multicolumn{1}{c}{$r_w$}& \multicolumn{1}{c}{$r$} & \multicolumn{1}{c}{$T_u$} & \multicolumn{1}{c}{$H_u$}  & \multicolumn{1}{c}{$T_l$} & \multicolumn{1}{c}{$H_l$}&\multicolumn{1}{c}{$L$} & \multicolumn{1}{c}{$k_w$}   \\ \hline
        \multicolumn{1}{c|}{Main SI}   &$\cellcolor{columncolor}0.541$       & $\cellcolor{columncolor}6.67e{-6}$      & $\cellcolor{columncolor}1.68e{-7}$ & $\cellcolor{columncolor}0.153$   & $\cellcolor{columncolor}1.92e{-5}$  & $\cellcolor{columncolor}0.091$ & $\cellcolor{columncolor}0.153$ & $\cellcolor{columncolor}0.037$    \\
        \multicolumn{1}{c|}{Total SI} & $\cellcolor{columncolor}0.562$       & $\cellcolor{columncolor}3.98e{-6}$      & $\cellcolor{columncolor}6.95e{-8}$ & $\cellcolor{columncolor}0.163$   & $\cellcolor{columncolor}2.83e{-5}$  & $\cellcolor{columncolor}0.096$ & $\cellcolor{columncolor}0.164$ & $\cellcolor{columncolor}0.037$ \\
        \multicolumn{1}{c|}{$10^{\omega_{i}}$} & $\cellcolor{columncolor}-$ & $\cellcolor{columncolor}4.00e{-4} $      & $\cellcolor{columncolor}5.06e{-6}$      & $\cellcolor{columncolor}0.013$ & $\cellcolor{columncolor}3.00e{-4}$ & $\cellcolor{columncolor}-$  & $\cellcolor{columncolor}0.029$ & $\cellcolor{columncolor}0.004$           \\
        \multicolumn{1}{c|}{$S_{cat}$} & $\cellcolor{columncolor}0.115$ & $\cellcolor{columncolor}-$      & $\cellcolor{columncolor}-$      & $\cellcolor{columncolor}-$ & $\cellcolor{columncolor}-$ & $\cellcolor{columncolor}0.019$  & $\cellcolor{columncolor}-$ & $\cellcolor{columncolor}-$           \\
        \hline   
    \end{tabular}
       \caption{\textbf{Sensitivity analysis of \boreholeM:} Sensitivity analysis of \boreholeM~using Sobol indices and \gp~emulator. The $\boldsymbol{\omega}$ parameters in \gp~are only learned for quantitative features.}
    \label{table: SA_borehole_mixed}
\end{table}

\begin{table}[!h]
    \centering
    \begin{tabular}{c|cccccccccc}
    \setlength{\tabcolsep}{1pt}
        & \multicolumn{4}{c}{} & Features &\\ \hline
        \multicolumn{1}{c|}{Metric} &\multicolumn{1}{c}{$S_w$}& \multicolumn{1}{c}{$w_{fw}$} & \multicolumn{1}{c}{$A$} & \multicolumn{1}{c}{$\Lambda$}  & \multicolumn{1}{c}{$q$} & \multicolumn{1}{c}{$\lambda$} & \multicolumn{1}{c}{$t_c$}  & \multicolumn{1}{c}{$N_z$} & \multicolumn{1}{c}{$w_{dg}$} & \multicolumn{1}{c}{$w_p$} \\ \hline
        \multicolumn{1}{c|}{Main SI}  & $0.125$       & $2.3e{-6}$      & $0.220$ & $4.8e{-4}$   & $8.4e{-5}$  & $1.8e{-3}$ & $0.142$ & $0.412$ & $0.085$ & $3.3e{-3}$     \\
        \multicolumn{1}{c|}{Total SI}  & $0.128$       & $2.4e{-6}$      & $0.225$ & $5.1e{-4}$   & $8.7e{-5}$  & $1.8e{-3}$ & $0.146$ & $0.420$  & $0.087$ &$3.3e{-3}$    \\
        \multicolumn{1}{c|}{$10^{\omega_i}$} & $0.005$ & $8.7e{-7}$      & $0.010$      & $0.006$ & $2.0e{-4}$ & $5.0e{-4}$  & $0.027$ & $0.023$ & $0.005$  & $2.0e{-4}$         \\
        \hline    
    \end{tabular}
    \caption{\textbf{Sensitivity analysis of \wing:} Sensitivity analysis of \wing~using Sobol indices and \gp emulator. In \Cref{sec: Calibration_Functionality}, $\zetab = \brackets{q, \lambda, t_c, N_z}^T$ are treated as calibration parameters.}
    \label{table: SA_wing}
\end{table}


\section{Emulation and Optimization Options}
\gp~offers a wide range of options that streamline its adoption for a wide range of applications that involve emulation, MF modeling, identification of model form errors, inverse parameter estimation, and BO.
In \Cref{table: model_options,table: optim_options} we provide a comprehensive list of options related to model initialization and training, respectively. We have chosen the default values of these options based on the most common uses of GPs while striking a balance between accuracy and cost. 

\begin{table}[h]
    \footnotesize
    \centering
    \begin{tabular}{lll}
        \hline
        \textbf{Option Name} & \textbf{Description} & \textbf{Type/Default Setting} \\ \hline
        
        \texttt{train\_x} & Input & Tensor \\ \hline
        \texttt{train\_y} & Output (response) & Tensor \\ \hline
        \texttt{dtype} & Data type of the model and data & \texttt{torch.float} \\ \hline
        \texttt{device} & Device to build the model (CPU or CUDA) & \texttt{"cpu"} \\ \hline
        \texttt{qual\_dict} & Column index and number of levels of categorical variables & \{\} \\ \hline
        \texttt{multiple\_noise} & Modeling separate noise for each data source & \texttt{False} \\ \hline
        \texttt{lb\_noise} & Lower bound for noise & $1e-8$ \\ \hline
        \texttt{fix\_noise} & Flag for estimating noise via the nugget parameter & \texttt{False} \\ \hline
        \texttt{fix\_noise\_val} & Fixed noise value if \texttt{fix\_noise = True} & $1e-5$ \\ \hline
        \texttt{quant\_correlation\_class} & Kernel of numerical variables & \texttt{"Rough\_RBF"} \\ \hline
        \texttt{fixed\_length\_scale} & Flag to fix the length scale & \texttt{False} \\ \hline
        \texttt{fixed\_length\_scale\_val} & Fixed length scale value & \texttt{torch.tensor([1.0])} \\ \hline
        \texttt{encoding\_type} & Type of $f_{\boldsymbol{\pi}}(\boldsymbol{t})$ & \texttt{"one-hot"} \\ \hline
        \texttt{embedding\_dim} & Dimension of embedding (manifold) to be learnt & $2$ \\ \hline
        \texttt{separate\_embedding} & Which categorical features be learned separately & [~] \\ \hline
        \texttt{embedding\_type} & Type of embedding & \texttt{"deterministic"} \\ \hline
        \texttt{NN\_layers\_embedding} & Network structure of $f_h(\boldsymbol{\pi_{t}},\boldsymbol{\theta}_h)$ & [~] \\ \hline
        \texttt{m\_gp} & Type of mean function for GP & \texttt{"single"} \\ \hline
        \texttt{m\_gp\_ref} & Mean function for reference source (ID = 0) & \texttt{"zero"} \\ \hline
        \texttt{NN\_layers\_m\_gp} & Structure of neural network for mean function & $[4,4]$ \\ \hline
        \texttt{calibration\_type} & Deterministic or probabilistic calibration & \texttt{"deterministic"} \\ \hline
        \texttt{calibration\_id} & Index of the parameter to be calibrated & [~] \\ \hline
        \texttt{mean\_prior\_cal} & Mean prior for calibration parameter& $0$ \\ \hline
        \texttt{std\_prior\_cal} & Standard deviation prior for calibration parameter & $1$ \\ \hline
        \texttt{interval\_score} & Interval scoring during optimization & \texttt{False} \\ \hline
        {\color{black}\texttt{num\_pass\_train}} & {\color{black}Number of training passes; deterministic/probabilistic} & {\color{black}$1/20$} \\ \hline
        {\color{black}\texttt{num\_pass\_pblack}} & {\color{black}Number of pblackiction passes; deterministic/probabilistic} & {\color{black}$1/30$} \\ \hline
    \end{tabular}
    \caption{\textbf{Model options:} We provide a range of options that streamline the adoption of \gp~in many applications with just a few lines of code.}
    \label{table: model_options}
\end{table}

\begin{table}[!t]
    \footnotesize
    \centering
    \begin{tabular}{lll}
        \hline
        \textbf{Option Name} & \textbf{Description} & \textbf{Type/Default Setting } \\ \hline
        \texttt{add\_prior} & Flag for using MAP instead of MLE & \texttt{True} \\ \hline
        \texttt{jac} & Flag for using Jacobian & \texttt{True} \\ \hline
        \texttt{num\_restarts} & Number of optimization restarts & $32$ \\ \hline
        \texttt{method} & Optimization method & \texttt{"L-BFGS-B"} \\ \hline
        \texttt{options} & Optional parameters & $\{\}$ \\ \hline
        \texttt{n\_jobs} & Number of cores (uses all cores if $\texttt{n\_jobs}=-1$) & $-1$ \\ \hline
        \texttt{constraint} & Flag for adding constraints & \texttt{False} \\ \hline
        \texttt{bounds} & Flag for adding bounds on parameters & \texttt{False} \\ \hline
        \texttt{regularization\_parameter} & Regularization coefficients & $[0,0]$ \\ \hline
    \end{tabular}
    \caption{\textbf{Options for model training:} These options control the optimization process and their default values are selected to strike a balance between accuracy and cost.}
    \label{table: optim_options}
\end{table}

\section{Mixed Single-Task GP (MST-GP)} \label{sec: MST_GP}
As mentioned in \Cref{sec: emulation_functionality}, BoTorch employs MST-GP to model problems with categorical variables. MST-GP defines two distinct correlation functions for numerical and categorical features. The final correlation function is the combination of these two:

\begin{equation}
    \begin{aligned}
     r(\left[ \begin{array}{l} \xb \\ \tb \end{array} \right], \left[ \begin{array}{l} \xb^{\prime} \\ \tb^{\prime} \end{array} \right]; \boldsymbol{\Omega})=  r(\xb, \xb^{\prime}; \omegab_{1})+ r(\tb, \tb^{\prime}; \omegab_{2}) + r(\xb, \xb^{\prime}; \omegab_{3}) \times r(\tb, \tb^{\prime}; \omegab_{4})
    \end{aligned}
    \label{eq: mst_gp_corr}
\end{equation}
\noindent where $\{\omegab_{1}, \omegab_{2}, \omegab_{3}, \omegab_{4}\} \in \Omega$ are the distinct length scale parameters for each correlation function. Specifically, $\omegab_{1}$ and $\omegab_{3}$ are length scale parameters associated with quantitative features while $\omegab_{2}$ and $\omegab_{4}$ scale the categorical features ($\omegab_{1}$ and $\omegab_{3}$ are $dx$-dimensional while $\omegab_{2}$ and $\omegab_{4}$ are of dimension $dt$). $r(\xb, \xb^{\prime}; \omegab_j)$ is the Mat\`ern correlation function with $v=2.5$ while $r(\tb, \tb^{\prime}; \omegab_j)$ is formulated as:

\begin{equation}
    \begin{aligned}
     r\left(\tb, \tb^{\prime}; \omegab_j \right) = 
            \exp \left\{(-\sum_{i=1}^{dt} \frac{(t_i-t_i^{\prime})}{\omega_{ji}})/dt \right \} 
    \end{aligned}
    \label{eq: mst_gp_cat}
\end{equation}
\noindent where $\omega_{ji}$ is the length scale parameter estimated for the $i^{(th)}$ feature through correlation function $j$. Also, $(t_i-t_i^{\prime})$ is the Hamming distance which is $0$ when the two categorical variables are the same and $1$ otherwise.

\section{Single-Task Multi-Fidelity GP (STMF-GP)} \label{sec: STMF_GP}

The STMF-GP modifies the correlation function of GP for MF hierarchical MF modeling. Specifically, it adopts an additive covariance function that relies on introducing two user-defined quantitative features \cite{RN1392, RN1270}. The first feature, denoted by $x_a$, is restricted to the $[0, 1]$ range and assigns a fidelity value to a source based on the user's belief (larger values correspond to higher fidelities). This assigned fidelity value directly affects the correlation and cost function. The second feature, denoted by $x_b$, is the iteration fidelity parameter and benefits MF BO specifically in the context of hyperparameter tuning of large machine learning models.  These two features are used in three user-defined functions defined as follows. $e_1(\cdot)$ and $e_3(\cdot)$ are bias kernels that aim to take the discrepancies among the sources into account: 
\begin{equation} 
    \begin{split}
        e_1(x_a, x^\prime_{a})=(1-x_{a})(1-x^\prime_{a})(1+x_{a} x^\prime_{a})^p
    \end{split}
    \label{eq: e1}
\end{equation}
\begin{equation} 
    \begin{split}
        e_3(x_b, x^\prime_{b})=(1-x_{b})(1-x^\prime_{b})(1+x_{b} x^\prime_{b})^p
    \end{split}
    \label{eq: e3}
\end{equation}
\noindent where $p$ is the degree of polynomial (which needs to be estimated) and has a Gamma prior. $e_2(\cdot)$ is the interaction term with four deterministic terms and one polynomial kernel:
\begin{equation} 
    \begin{split}
        e_2([x_a, x_b]^T,[x_a^{\prime}, x_b^{\prime}]^T)=(1-x_{b})(1-x^\prime_{b})(1-x_{a})(1-x^\prime_{a})(1+[x_a, x_b]^T[x_a^{\prime}, x_b^{\prime}]^T)^p
    \end{split}
    \label{eq: e2}
\end{equation}

Finally, the modified covariance function is~\cite{RN1790}:


\begin{equation}
\begin{split}
    \operatorname{cov}(\boldsymbol{x}, \boldsymbol{x}^\prime) = &
        c(\boldsymbol{x}, \boldsymbol{x}^\prime; \thetab_{0}, {\sigma_{0}}^2) + e_1(x_{a}, x^\prime_{a}) c(\boldsymbol{x}, \boldsymbol{x}^\prime;{\thetab_{1}}, {\sigma_{1}}^2) \\
    & + e_2([x_a,x_b]^T, [ x^\prime_{a}, x^\prime_{b}]^T) c(\boldsymbol{x}, \boldsymbol{x}^\prime;{\thetab_{2}}, {\sigma_{2}}^2)  + e_3(x_{b}, x^\prime_{b}) c(\boldsymbol{x},\boldsymbol{x}^\prime;{\thetab_{3}}, {\sigma_{3}}^2)
\end{split}
\label{eq:Singletaskgp-Cov}
\end{equation}
\noindent where $c(\boldsymbol{x}, \boldsymbol{x}^\prime;{\thetab_{i}}, {\sigma_{i}}^2)$ is the Matern kernel that characterize the spatial correlations across the numerical inputs (the parameters of these kernels are endowed with Gamma priors in BoTorch).

In \Cref{sec: mfmodeling_functionality} we use STMF-GPs as one of the baselines for evaluating the performance of \gp~in MF emulation. Therein, we assing two sets of values to the fidelity indices of STMF-GP to quantify their effect on the results. These two sets of values are enumerated in
\Cref{table: mfst_gp_fidelities}.
\begin{table}[!h]
    \centering
    \begin{tabular}{c|ccc}
        \hline 
        & \multicolumn{1}{c}{} & Fidelity Parameters ($\boldsymbol{x}_a$) &\\ \hline
        \multicolumn{1}{c|}{Model} & \multicolumn{1}{c}{Sinusoidal} & \multicolumn{1}{c}{Wing-weight} & \multicolumn{1}{c}{DNS-ROM} \\ \hline
        \multicolumn{1}{c|}{$STMF-GP_{1}$}   & $[1,0.25]$    &  $[1,0.96,0.83,0.49]$       & $[1,0.96,0.83,0.49]$                             \\ 
        \multicolumn{1}{c|}{$STMF-GP_{2}$ }   & $[1,0.5]$    &  $[1,0.75,0.5,0.25]$      & $[1,0.6,0.4,0.2]$\\
        \hline
        
    \end{tabular}
    \caption{\textbf{Fidelity indices of STMF-GP:} We use these fidelity indices in \Cref{sec: mfmodeling_functionality} to demonstrate their effect of MF emulation. Even though the two sets of numbers are close, the performance of the corresponding emulators are quite different.}
    \label{table: mfst_gp_fidelities}
\end{table}

        

\section{Neural Network Architectures for Multi-fidelity Modeling} \label{sec: appendix_MF_details}
In \Cref{sec: mfmodeling_functionality}, FFNNs are employed both as MF emulators and as basis functions in \gp. We design the architecture of these models as detailed in \Cref{table: nn_structure}. 
\begin{table}[!h]
\centering
\begin{tabular}{c|c|ccc}
\hline
\multirow{2}{*}{Method} & \multirow{2}{*}{Option} & \multicolumn{3}{c}{Problems}        \\ \cline{3-5} 
                        &                         & \sine     & \wing       & \DNS          \\ \hline
\multirow{2}{*}{\gp}     & Small FFNN as $m(\ub; \betab)$                   & $[1,2]$   & $[4,4]$      & $[4,4]$        \\ \cline{2-2}
                        & Medium FFNN as $m(\ub; \betab)$                    & $[2,2,2]$ & $[8,8,8]$    & $[8,8,8]$      \\ \hline
\multirow{3}{*}{FFNN}   & Small                   & $[4,4]$   & $[8,8,8]$    & $[8,8,8]$      \\ \cline{2-2}
                        & Medium                  & $[16,16]$ & $[4,16,32]$  & $[32,32,32]$   \\ \cline{2-2}
                        & Large                   & $[16,32]$  & $[4,16,128]$ & $[128,128,32]$ \\ \hline
\end{tabular}
\caption{\textbf{Network architectures of feed-forward neural networks:} We design different architectures for our MF emulation studies in \Cref{sec: mfmodeling_functionality}}
    \label{table: nn_structure}
\end{table}

\section{Bayesian Optimization (BO)} \label{sec: BO_Appendix}
In this section, we first explain how \gp~handles highly biased sources in MFBO and then provide a few options available in \gp~for SFBO and MFBO.

\subsection{BO Improvements} \label{sec: BO_Appendix_improvement}

As mentioned in \Cref{sec: BO}, the accuracy of the emulator in quantifying uncertainties significantly affects the performance of BO.
This impact is more noticeable in MF problems where biased LF sources can misguide the optimization process. To address this challenge, we employ interval scores ($IS$) to penalize the objective function. We choose $IS$ since it is robust to outliers, rewards narrow pblackiction intervals, and is flexible in the choice of desiblack coverage levels \cite{bracher2021evaluating, mitchell2017proper}. 
$IS$ is a special case of quantile pblackiction that penalizes the emulator for each observation that is not inside the $(1-v) \times 100 \%$ pblackiction interval and is calculated as:

\begin{equation} 
    \begin{split}
        IS= \frac{1}{n} \sum_{i=1}^n(\mathcal{U}^{(i)}-\mathcal{L}^{(i)})+\frac{2}{v}(\mathcal{L}^{(i)}-y^{(i)}) \mathbbm{1}~\{y^{(i)}<\mathcal{L}^{(i)}\} +\frac{2}{v}(y^{(i)})-\mathcal{U}^{(i)}) \mathbbm{1}\{y^{(i)}>\mathcal{U}^{(i)}\} 
    \end{split}
    \label{eq: IS}
\end{equation}
\noindent where $y^{(i)}=y(\boldsymbol{u}^{(i)})$ is the response of the $i^{th}$ training sample. $\mathcal{U}^{(i)}$, $\mathcal{L}^{(i)}$, $v$ and $\mathbbm{1}$ are defined in \Cref{sec: functionalities}. 

Using the $IS$ in \Cref{eq: IS} we now formulate the new objective function for emulation within BO where $IS$ is used as a penalty term. Since the effectiveness of this penalization mechanism depends on the value of the posterior, we introduce an adaptive coefficient whose magnitude depends on the posterior value. With this penalty term, the modified objective function for the GP emulator is:
\begin{equation} 
    \begin{split}
     {[\widehat{\betab}, \widehat{\sigma}^2, \widehat{\boldsymbol{\thetab}}, \widehat{\boldsymbol{\delta}}]=}  \underset{\betab, \sigma^2, \boldsymbol{\thetab}, \boldsymbol{\delta}}{\operatorname{argmin}}  ~L_{MAP}  +\epsilon|L_{MAP}| \times IS
    \end{split}
    \label{eq: IS_based_objective}
\end{equation}
\noindent where $|\cdot|$ denotes the absolute function and $\epsilon$ is a user-defined scaling parameter which is set to $0.08$ by default in \gp. We refer the reader to \cite{foumani2023effects} for more in-depth information. 

\subsection{BO Options} \label{sec: BO_appendix_table}
The BO options of \gp~are summarized in \Cref{table: BO_options}. 
We note that \gp~is able to handle both analytic functions and datasets for BO. This versatility is achieved by specifying the analytic function for the former (see \Cref{fig: BO_screenshot} for an example) and utilizing datasets for the latter through the \texttt{data\_func} option.


\begin{table}[h]
    \footnotesize
    \centering
    \begin{tabular}{lll}
        \hline
        \textbf{Option Name} & \textbf{Description} & \textbf{Type/Default Setting } \\ \hline
        \texttt{U\_init} & Initial input & Tensor \\ \hline
        \texttt{y\_init} & Initial output (response)& Tensor \\ \hline
        \texttt{costs} & Source-dependent sampling costs & $\{\}$ \\ \hline
        \texttt{l\_bound} & Lower bound of the  variables & [~] \\ \hline
        \texttt{u\_bound} & Upper bound of the  variables & [~] \\ \hline
        \texttt{U\_mean} & Mean of the initial inputs & [~] \\ \hline
        \texttt{U\_std} & Standard deviation of the initial inputs & [~] \\ \hline
        \texttt{qual\_dict} & Column index and number of levels of categorical variables & $\{\}$ \\ \hline
        \texttt{data\_func} & Data (function / dataset) & - \\ \hline
        \texttt{n\_train\_init} & Number of initial data & $\{\}$ \\ \hline
        \texttt{maximize\_flag} & Flag for maximization  & \texttt{False} \\ \hline
        \texttt{one\_iter} & Flag for suggesting only one new sample & \texttt{False} \\ \hline
        \texttt{max\_cost} & Maximum budget for optimization & $40000$ \\ \hline
        \texttt{MF} & Flag for doing MFBO  & \texttt{True} \\ \hline
        \texttt{AF\_hf} & AF of HF source  & \texttt{AF\_HF} \\ \hline
        \texttt{AF\_lf} & AF of LF source  & \texttt{AF\_LF} \\ \hline
        \texttt{IS} & Flag for penalizing the objective function with scoring rules  & \texttt{True} \\ \hline
    \end{tabular}
    \caption{\textbf{Options provided by \gp~for Bayesian optimization:} \gp~accommodates both single- and multi-fidelity BO. The user can provide both analytic functions and datasets where the former is typically used in comparison studies while the latter is used in real world applications.}
    \label{table: BO_options}
\end{table}

\section{BoTorch} \label{sec: BoTorch}
\BOT~is an MF cost-aware BO package that employs STMF-GP (explained in \Cref{sec: STMF_GP}) as the emulator and leverages the knowledge gradient (KG) as the AF. KG is a look-ahead AF that chooses the next sampling point ($\boldsymbol{x}^{(k+1)}$) based on the effect of the yet-to-be-seen observation (i.e., $y^{k+1}$ which follows a normal distribution) on the optimum value pblackicted by the emulator. This AF quantifies the expected utility of $\boldsymbol{x}$ at iteration $k+1$ as:
\begin{equation} 
    \begin{split}
    \gamma_{KG}(\boldsymbol{x}) = 
    \mathbb{E}_{p(y \mid \boldsymbol{x}, \mathcal{D}^k)}[\max{\mu^{(k+1)}}] -
    \max{\mu^{(k)}}
    \label{eq: KG-AF-untract}
    \end{split}
\end{equation}
\noindent where $\mathcal{D}^{k}=\{(\boldsymbol{x}^{(i)}, y^{(i)})\}_{i=1}^k$ is the training data in iteration $k$ and $\max{\mu^{(k)}} =\max{\mu^{(k)}}(\boldsymbol{x})$ denotes the maximum mean pblackiction of the emulator trained on $\mathcal{D}^k$. The expectation operation in \Cref{eq: KG-AF-untract} appears due to the fact that $y^{(k+1)}$ is not observed yet and $\alpha_{KG}(\boldsymbol{x})$ is relying on the pblackictive distribution provided by the emulator that is trained on $\mathcal{D}^k$. This expectation cannot be calculated analytically and hence a Monte Carlo estimate is used in practice:
\begin{equation} 
    \begin{split}
        \gamma_{KG}(\boldsymbol{x}) \approx 
        \frac{1}{M} \sum_{m=1}^M \max{{\mu^{(k+1)}}^m} -
        \max{\mu^{(k)}}
        \end{split}
    \label{eq: KG-AF}
\end{equation}
\noindent where $\max{{\mu^{(k+1)}}^m}=\max{{\mu^{(k+1)}}^m(\boldsymbol{x})}$ is calculated by first drawing a sample at $\boldsymbol{x}$ from the emulator that is trained on $\mathcal{D}^k$ and then retraining the emulator on $\mathcal{D}^{k} \cup (\boldsymbol{x}, y^m)$ where $y^m$ is response of the drawn sample. In practice, a small value must be chosen for $M$ since maximizing $\gamma_{KG}(\boldsymbol{x})$ over the input space at each iteration of BO is very expensive. Refer to \cite{frazier2008knowledge, balandat2020botorch} for more information on KG and its implementation.

{\color{black}\section{Computational Costs} \label{sec: Computational Costs of Emulation}} 

\begin{figure}[!h] 
    \centering
        \centering
        \includegraphics[width=1\linewidth]{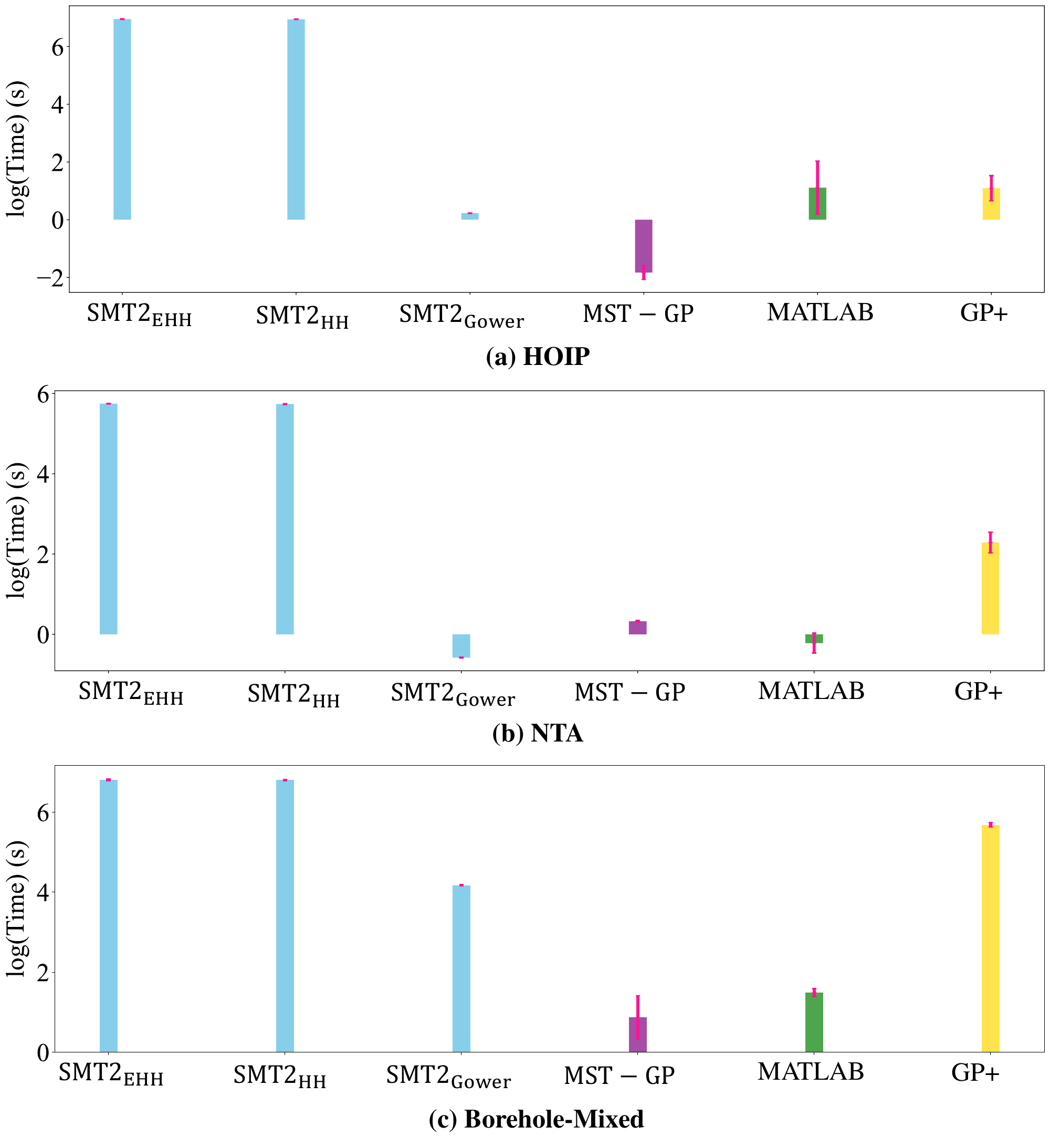}
        \vspace{-8mm}
    \caption{{\color{black}\textbf{Training cost in examples with categorical inputs:} Each bar in the graph represents the computational costs associated with different methods for various examples. Since EHH, HH, and Gower are from SMT2, we use the same color to represent them. In the context of mixed-input problems, the manifold learning process of \gp~leads to a slight increase in its computational cost. However, given the significantly higher accuracy it achieves, this difference in time is justified.}}
    \label{fig: time_comapre}
\end{figure}

\begin{figure}[!h] 
    \centering
        \includegraphics[width=1\linewidth]{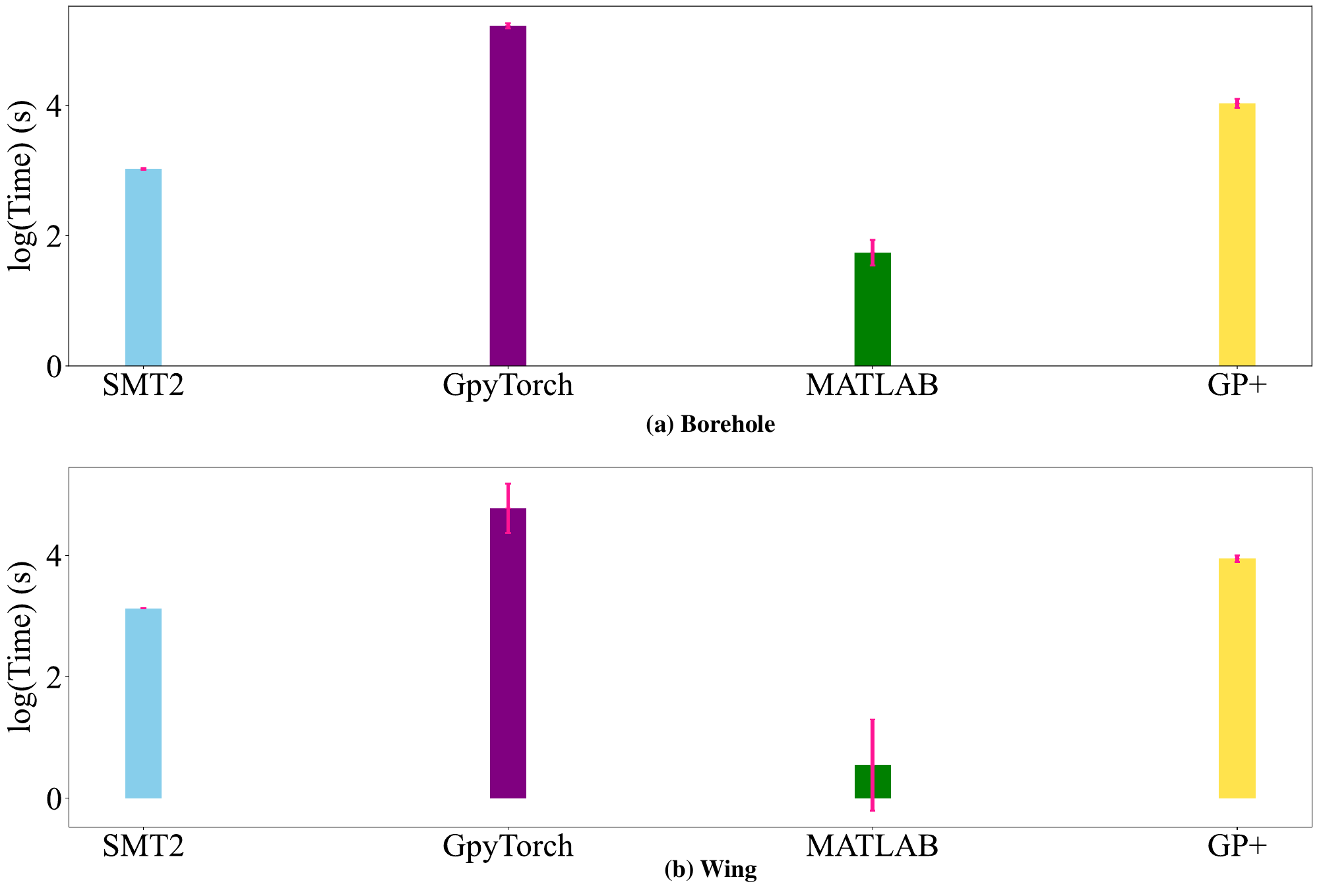}
        \vspace{-8mm}
    \caption{{\color{black}\textbf{Training cost in examples with all-numerical inputs:} Each bar in the graph represents the computational costs associated with different methods for various examples. In the context of numerical problems, all the benchmarks have the same formulation and they just differ in their optimization. This difference causes slightly different computational costs.}}
    \label{fig: time_comapre-numericals}
\end{figure}

{\color{black}In this section, we compare the computational costs of various baselines discussed in \Cref{sec: emulation_functionality} for the examples outlined in \Cref{table: results_emulation}. 
The results are summarized in \Cref{fig: time_comapre,fig: time_comapre-numericals}. \Cref{fig: time_comapre} is for problems whose input space has categorical variables while \Cref{fig: time_comapre-numericals} is for problems that only have numerical inputs. We observe in \Cref{fig: time_comapre} that \gp~is slightly more expensive than Gower, MST-GP, and MATLAB. This is because, unlike packages that convert categorical features to numerical ones in a naive manner (MATLAB, Gower, and MST-GP), \gp~nonlinearly learns the underlying characteristics of the categorical inputs. Additionally, we find that the larger number of parameters used in $\text{SMT2}_{\text{EHH}}$ and $\text{SMT2}_{\text{HH}}$ not only fails to improve accuracy, but also leads to significantly higher computational costs for SMT2.

In numerical examples (Wing and Borehole, see \Cref{fig: time_comapre-numericals}), the formulations of all methods are the same and they only differ in the parameter estimation and optimization. 
Specifically, for MATLAB, we employ default settings that utilize MLE for parameter estimation and the BFGS algorithm for optimization. The lower computational costs observed in MATLAB are attributed to its optimization process and streamlined implementation. SMT2 exhibits slightly lower computational costs compablack to \gp~and GPytorch which is due to its use of the profiling method. Furthermore, while \gp~and GPytorch are quite similar, the minor difference in their computational costs stems from the different optimizers they employ (\gp~uses L-BFGS from SciPy by default, while GPytorch uses Adam\footnote{Adaptive Moment Estimation} from PyTorch). }



\end{appendices}

    \pagebreak
    \printbibliography
\end{document}